# Towards Understanding Ensemble, Knowledge Distillation and Self-Distillation in Deep Learning

Zeyuan Allen-Zhu
zeyuan@csail.mit.edu
Microsoft Research Redmond

Yuanzhi Li
yuanzhil@andrew.cmu.edu
Carnegie Mellon University

December 17, 2020

(version 2)*

**Abstract**

We formally study how *ensemble* of deep learning models can improve test accuracy, and how the superior performance of ensemble can be distilled into a single model using *knowledge distillation*. We consider the challenging case where the ensemble is simply an average of the outputs of a few independently trained neural networks with the *same* architecture, trained using the *same* algorithm on the *same* data set, and they only differ by the random seeds used in the initialization.

We empirically show that ensemble/knowledge distillation in *deep learning* works very differently from traditional learning theory, especially differently from ensemble of random feature mappings or the neural-tangent-kernel feature mappings, and is potentially out of the scope of existing theorems. Thus, to properly understand ensemble and knowledge distillation in deep learning, we develop a theory showing that when data has a structure we refer to as "multi-view", then ensemble of independently trained neural networks can provably improve test accuracy, and such superior test accuracy can also be provably distilled into a single model by training a single model to match the output of the ensemble instead of the true label. Our result sheds light on how ensemble works in deep learning in a way that is completely different from traditional theorems, and how the "dark knowledge" is hidden in the outputs of the ensemble— that can be used in knowledge distillation— comparing to the true data labels. In the end, we prove that self-distillation can also be viewed as implicitly combining ensemble and knowledge distillation to improve test accuracy.

---

*V1.5 appears on arXiv on this date, and V2 polishes writing.

# 1 Introduction

Ensemble [25, 41, 50, 66, 69, 69, 70, 72, 90] (a.k.a. model averaging) is one of the oldest and most powerful techniques in practice to improve the performance of deep learning models. By simply averaging the output of a few independently trained neural networks over the **same** training data set, it can significantly boost the prediction accuracy over the test set comparing to individual models. Moreover, it is also known that such performance boost holds *even when* all the neural networks have the *same* architecture and are trained using the *same* training algorithm over the *same* training data set. The only difference is the randomness used to initialize these neural networks and/or the randomness during training (e.g. if stochastic gradient descent is used). For example, on the standard CIFAR-100 data set, averaging the output of ten independently trained ResNet-34 can easily offer a 5% improvement in terms of test accuracy. Later, it is also discovered in the seminal work of Hinton, Vinyals, and Dean [42] that such superior test-time performance of the ensemble can be transferred into a single model (of the same size as the individual models) using a technique called *knowledge distillation*: that is, simply train a single model to match the output of the ensemble (such as "90% cat + 10% car", also known as *soft labels*) as opposite to the true data labels, over the same training data.

On the theory side, there are lots of works focusing on understanding the superior performance of ensemble from a principled perspective [11, 13, 28, 30–33, 36, 43, 46–48, 51, 71, 71, 74, 75]. However, most of these works only apply to: (1). Boosting: where the coefficients associated with the combinations of the single models are actually trained, instead of simply taking average; (2). Bootstraping/Bagging: the training data are different for each single model; (3). Ensemble of models of different types and architectures; or (4). Ensemble of random features or decision trees.

To the best of our knowledge, *none* of these theoretical works apply to the particular type of ensemble that is widely used in deep learning: simply take a *uniform* average of the output of the learners, which are neural networks with the *same* architecture and are trained by stochastic gradient descent (SGD) over the *same* training set. Nor can they explain why the superior performance of ensemble can be distilled into a single model [20, 22, 29, 34, 42, 60, 91]. In fact, as we shall argue in our paper, some traditional views of ensemble, such as ensemble can enlarge the feature space in random feature mappings, *cannot* explain knowledge distillation in deep learning.

Motivated by the gap between theory and practice in *ensemble and knowledge distillation in deep learning*, we formally study the following question for *multi-class* classification:

> **Our theoretical questions:**
> How does ensemble improve the test-time performance in deep learning when we simply (unweighted) average over a few independently trained neural networks? – Especially when all the neural networks have the same architecture, are trained over the same data set using the same standard training algorithm (i.e. stochastic gradient descent with the same learning rate and sample regularization) and only differ by the random seeds, and even when all single models already have 100% *training accuracy*? How can such superior test-time performance of ensemble be later "distilled" into a single neural network of the same architecture, simply by training the single model to match the output of the ensemble over the same training data set?

## 1.1 Our Theoretical Results at a High Level

To the best of our knowledge, this paper makes a first step towards answering these questions in deep learning. On the *theory side*, we prove that over certain multi-class classification tasks with a special structure that we shall refer to as **multi-view**, with a training set $\mathcal{Z}$ consisting of $N$ i.i.d.

samples from some unknown distribution $\mathcal{D}$, for certain two-layer convolutional network $f$ with (smoothed-)ReLU activation as learner:

- (Single model has bad test accuracy): there is a value $\mu > 0$ such that when a single model $f$ is trained over $\mathcal{Z}$ using the cross-entropy loss, via gradient descent (GD) starting from random Gaussian initialization, the model can reach zero training error *efficiently*. However, w.h.p. the prediction (classification) error of $f$ over $\mathcal{D}$ is between $0.49\mu$ and $0.51\mu$.

- (Ensemble provably improves test accuracy): let $f_1, f_2, \cdots, f_L$ be $L = \widetilde{\Omega}(1)$ independently trained single models as above, then w.h.p. $G = \frac{1}{L}\sum_\ell f_\ell$ has prediction error $\leq 0.01\mu$ over $\mathcal{D}$.

- (Ensemble can be distilled into a single model): if we further train (using GD from random initialization) another single model $f_0$ (same architecture as each $f_\ell$) to match the output of $G = \frac{1}{L}\sum_\ell f_\ell$ merely over the same training data set $\mathcal{Z}$, then $f_0$ can be trained *efficiently* and w.h.p. $f_0$ will have prediction error $\leq 0.01\mu$ over $\mathcal{D}$ as well.

- (*Self-distillation* [35, 62, 88] also improves test accuracy): if we further train (using GD from random initialization) another single model $f'$ (same architecture as $f_1$) to match the output of *the single model* $f_1$ merely over the same training data set $\mathcal{Z}$, then $f'$ can be trained *efficiently* and w.h.p. has prediction error at most $\leq 0.26\mu$ over $\mathcal{D}$. The main idea is that self-distillation is performing "*implicit ensemble + knowledge distillation*", as we shall argue in Section 4.2.

Thus, on the theory side, we make a step towards understanding ensemble and knowledge distillation in deep learning both computationally (training efficiency) and statistically (generalization error).

## 1.2 Our Empirical Results at a High Level

On the *empirical side*, to further justify our approach studying ensemble and knowledge distillation indeep learning, we show:

- **Ensemble (i.e. model averaging) in deep learning works *very differently* from ensemble in random feature mappings** — in particular, different from the neural tangent kernel (NTK) approach [3, 5, 6, 8, 9, 19, 24, 26, 27, 39, 40, 44, 54, 58, 85, 92].

  Let us do a thought experiment. If ensemble works, can we obtain the same test performance of ensemble by training the sum of over $L$ neural networks $G = \sum_{\ell \in [L]} f_\ell$ directly? (As opposite to training each $f_\ell$ independently and then average them.) Actually, this "direct training" approach in deep learning is unable to improve the test accuracy even comparing to single models, not to say the ensemble model. See Figure 1.

  In contrast, when each $f_\ell$ is a linear function over random feature mappings (e.g., the NTK feature mappings given by the random initialization of the network), although the ensemble of these random feature models *does improve* test accuracy, training directly over $F = \sum_{\ell \in [L]} f_\ell$ gives even superior test accuracy comparing to the ensemble. See Figure 1.

- **Knowledge distillation works for "ensemble of neural networks" but *does not work* for "ensemble of random feature mappings" on standard data sets**.

  When $f_\ell$ is a linear function over random feature mappings (e.g., the NTK feature mappings), the superior test performance of ensemble *cannot be distilled* into a single model. In contrast, in deep learning, such superior performance can be distilled into a single model using [42]. The situation is similar for *self-distillation*, where it hardly works on improving test performance for neural kernel methods, but works quite well for real neural networks. See Figure 1. Together with the first point, experiments suggest that to understand the benefit of ensemble and knowledge distillation in deep learning, it is perhaps **inevitable to study deep learning as**

| finite-width neural kernel models | CIFAR10 test accuracy | | | | | CIFAR100 test accuracy | | | | |
|---|---|---|---|---|---|---|---|---|---|---|
| | single model (best of 10) | ensemble (over 10) | train $\sum_\ell f_\ell$ (over 10) | knowledge distillation | self-distill | single model (best of 10) | ensemble (over 10) | train $\sum_\ell f_\ell$ (over 10) | knowledge distillation | self-distill |
| **SimpleCNN-10-3**-NTK | 64.36% | 67.38% | 69.37% | 64.63% | 65.24% | | | | | |
| **ResNet10-2**-NTK | 69.15% | 73.29% | 74.71% | 68.82% | 66.09% | out of memory | | | | |
| **ResNet16-2**-NTK | 68.32% | 73.79% | 74.62% (over 7)$^\diamond$ | 66.12% | 70.61% | $^\diamond$due to memory restriction, trained $\sum_\ell f_\ell$ over fewer than 10 models. | | | | |
| **ResNet16-5**-NTK | 74.21% | 78.46% | out of memory | 70.23% | 75.66% | | | | | |
| **ResNet10-10**-NTK | 76.66% | 80.39% | out of memory | 77.25% | 74.46% | | | | | |
| **SimpleCNN10-6**-NTK' | 59.92% | 63.43% | 65.69% | 59.12% | 57.81% | 18.99% | 26.54% | 28.28% | 18.27% | 18.40% |
| **ResNet10-4**-NTK' | 66.68% | 70.54% | 72.86% | 66.01% | 62.91% | 31.90% | 38.32% | 41.47% | 31.38% | 27.64% |
| **SimpleCNN-10-6**-GP | 30.48% | 35.33% | 40.08% | 29.43% | 29.10% | 9.82% | 11.82% | 12.22% | 8.95% | 9.33% |
| **ResNet-10-4**-GP | 42.17% | 48.60% | 53.17% | 39.45% | 41.63% | 18.89% | 22.92% | 25.88% | 16.91% | 16.59% |

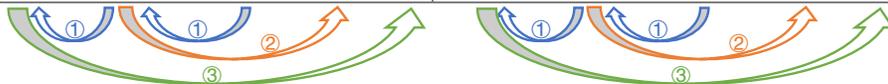

**Message ①**: for neural kernel methods, ensemble helps on improving test accuracies, but ensemble is _not better than_ training the sum of the individuals directly. In other words, the benefit of using ensemble here merely comes from the richer set of prescribed features.

**Message ②**: for neural kernel methods, the superior test performance of ensemble _cannot be distilled_ into a single model.

**Message ③**: for neural kernel methods, self-distillation is generally _no better than_ a single model's test performance.

| neural networks | single model (over 10) ⑦ | ensemble (over 10) | train $\sum_\ell f_\ell$ (over 10) | knowledge distillation | self-distill | single model (over 10) ⑦ | ensemble (over 10) | train $\sum_\ell f_\ell$ (over 10) | knowledge distillation | self-distill |
|---|---|---|---|---|---|---|---|---|---|---|
| **ResNet-28-2** | 95.22±0.14% | 96.33% | 95.02% | 96.16% | 95.78% | 76.38±0.23% | 81.13% | 73.18% | 79.03% | 78.12% |
| **ResNet-34** | 93.65±0.19% | 94.97% | 93.12% | 94.59% | 94.21% | 71.66±0.43% | 76.85% | 68.88% | 73.74% | 73.14% |
| **ResNet-34-2** | 95.45±0.14% | 96.55% | 95.00% | 96.08% | 95.86% | 77.01±0.35% | 81.48% | 72.99% | 79.23% | 79.07% |
| **ResNet-16-10** | 96.08±0.16% | 96.80% | 95.88% (over 6)$^\diamond$ | 96.81% | 96.62% | 80.03±0.17% | 83.18% | 80.53% (over 6)$^\diamond$ | 82.67% | 82.25% |
| **ResNet-22-10** | 96.44±0.09% | 97.12% | 96.41% (over 5)$^\diamond$ | 97.09% | 97.05% | 81.17±0.23% | 84.33% | 81.59% (over 5)$^\diamond$ | 83.71% | 83.26% |
| **ResNet-28-10** | 96.70±0.21% | 97.20% | 96.46% (over 4)$^\diamond$ | 97.22% | 97.13% | 81.51±0.16% | 84.69% | 81.83% (over 4)$^\diamond$ | 83.81% | 83.56% |

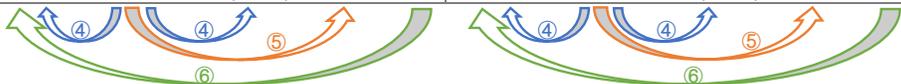

**Message ④**: for neural nets, ensemble helps on improving test accuracies, **and** this accuracy gain _cannot_ be matched by training the sum of the individuals directly. In other words, the benefit of using ensemble comes from somewhere other than enlarging the model.

**Message ⑤**: for neural nets, the superior test performance of ensemble _can be distilled_ into single model by a large extent.

**Message ⑥**: for neural nets, self-distillation _clearly improves_ the test performance of single models.

**Message ⑦**: for neural nets, the superior performance of ensemble _does not_ come from the variance of test accuracies in single models.

Figure 1: Comparing the performances of (1) training 10 independent single models $f_1, \ldots, f_{10}$, (2) their ensemble, (3) training $f_1 + \cdots + f_{10}$ directly, (4) knowledge distillation of the ensemble into a single model, and (5) training a single model using self-distillation.
(NTK' = the original finite-width neural network first-order approximation [6], NTK = the more popular variant where for each output label one learns a different linear function over the NTK features (e.g. [8]), and GP = training only the last layer of a finite-width random neural network [23]. All the neural networks in these experiments are trained to ∼ 100% training accuracy, and the single model performances match the state-of-the-art for these models on CIFAR-10/100. For experiment details, see Appendix A.1.)

**a feature learning process**, instead of feature selection process (e.g. NTK or other neural kernel methods) where only the linear combinations of prescribed features are trained.

- Some prior works attribute the benefit of ensemble to **reducing the variance** of individual solutions [15, 61, 63, 81, 82] due to label noise or non-convex landscape of the training objective. We observe that this **_may not_ be the cause for "ensemble in deep learning" to work.**

  – For standard deep learning datasets (e.g. CIFAR-10/100), individual neural networks (e.g. ResNets) trained by SGD typically have already converged to global optimas with 100% training accuracy and nearly zero training loss (no failure in training).

  – For standard deep learning datasets (e.g. CIFAR-10/100), ensemble helps even when there is essentially no label noise. In contrast, in our synthetic experiment Figure 2, ensemble does not help on Gaussian-like data even when there is label noise.

  – For standard neural networks (e.g. ResNets) trained on standard data set (e.g. CIFAR-10/100), when all the individual models are well-trained with the same learning rate/weight decay and only differ by their random seeds, there is *almost no variance* in test accuracy for individual models (e.g. $0.1 \sim 0.4\%$ std on CIFAR-100, see Figure 1). Hence with high probability, all individual models are learned *almost equally well* (no failure in generalization), yet ensemble still offers a huge benefit in test performance.

|  |  | no label noise | | | | with 10% label noise | | | |
|  |  | uniform sampling | | rejection sampling | | uniform sampling | | rejection sampling | |
|  |  | gaussian input | mixture of gaussian | gaussian input | mixture of gaussian | gaussian input | mixture of gaussian | gaussian input | mixture of gaussian |
| --- | --- | --- | --- | --- | --- | --- | --- | --- | --- |
| data without margin | linear | 80.3% (79.6%) | 80.7% (80.1%) | 78.9% (78.6%) | 80.7% (80.7%) | 74.3% (74.1%) | 73.6% (74.0%) | 72.9% (72.2%) | 74.2% (73.7%) |
|  | fc2 | 67.7% (65.1%) | 67.7% (64.9%) | 66.3% (64.5%) | 67.6% (66.9%) | 64.3% (63.2%) | 70.1% (66.7%) | 64.6% (63.5%) | 66.2% (63.3%) |
|  | fc3 | 68.9% (69.0%) | 64.0% (64.4%) | 76.8% (76.6%) | 73.2% (73.1%) | 66.5% (66.4%) | 63.0% (62.4%) | 72.5% (72.0%) | 78.1% (78.6%) |
|  | res3 | 69.1% (68.0%) | 70.7% (71.2%) | 69.3% (69.0%) | 69.9% (69.4%) | 68.7% (65.9%) | 66.9% (63.8%) | 68.1% (68.1%) | 68.8% (69.5%) |
|  | conv2 | 65.4% (65.7%) | 67.0% (66.8%) | 68.3% (68.2%) | 68.3% (68.2%) | 67.1% (66.2%) | 65.1% (65.5%) | 65.8% (66.0%) | 67.5% (67.9%) |
|  | conv3 | 68.7% (68.5%) | 70.7% (71.2%) | 77.8% (77.1%) | 80.3% (80.3%) | 67.5% (68.2%) | 67.7% (67.5%) | 73.6% (73.4%) | 71.8% (72.1%) |
|  | resconv3 | 78.3% (78.3%) | 79.6% (79.3%) | 83.8% (82.6%) | 82.1% (81.8%) | 74.1% (73.9%) | 73.4% (73.1%) | 78.7% (78.5%) | 79.2% (78.5%) |
| data with margin | linear | 79.0% (78.0%) | 79.0% (77.2%) | 78.4% (77.3%) | 80.0% (80.0%) | 82.1% (81.7%) | 80.7% (80.0%) | 81.6% (80.2%) | 84.1% (82.4%) |
|  | fc2 | 80.6% (79.0%) | 80.4% (78.4%) | 78.4% (76.6%) | 78.4% (77.0%) | 77.4% (75.3%) | 73.7% (73.9%) | 74.7% (72.2%) | 75.7% (71.4%) |
|  | fc3 | 76.0% (75.8%) | 80.4% (80.1%) | 76.9% (77.0%) | 73.3% (72.9%) | 70.7% (70.8%) | 73.9% (74.6%) | 70.4% (70.5%) | 67.5% (66.4%) |
|  | res3 | 80.7% (80.8%) | 84.7% (83.9%) | 84.6% (84.0%) | 84.4% (83.7%) | 75.5% (74.0%) | 76.9% (76.4%) | 76.8% (74.8%) | 76.4% (74.5%) |
|  | conv2 | 70.6% (70.3%) | 74.5% (73.6%) | 67.8% (67.8%) | 69.6% (69.4%) | 68.8% (67.5%) | 73.3% (71.7%) | 67.0% (66.6%) | 67.6% (67.2%) |
|  | conv3 | 76.2% (76.2%) | 75.3% (76.1%) | 79.6% (79.1%) | 84.3% (83.6%) | 72.2% (72.1%) | 81.2% (81.3%) | 73.0% (72.3%) | 74.4% (75.1%) |
|  | resconv3 | 92.1% (91.7%) | 92.1% (92.3%) | 93.9% (93.8%) | 95.5% (95.1%) | 85.2% (85.1%) | 83.5% (84.4%) | 87.3% (86.8%) | 85.6% (85.9%) |

Figure 2: When data is Gaussian-like, and when the target label is generated by some fully-connected(fc) / residual(res) / convolutional(conv) network, *ensemble does not* seem to improve test accuracy. " xx % (yy %)" means $xx\%$ accuracy for single model and $yy\%$ for ensemble. More experiments in Appendix A.4.

– For neural networks trained on our Gaussian-like data, there is relatively *higher variance* (e.g. $0.5 \sim 1.0\%$ std in test accuracies, see Figure 11 in the appendix), yet ensemble offers no benefit at all.

– For individual neural networks trained using knowledge distillation with different random seeds, ensemble does not improve their test accuracy by much (see Figure 3) — despite that the knowledge distillation objective is "as non-convex as" the original training objective and only the training labels are changed from hard to soft labels.

• **Special structure in data** (such as the "multi-view" structure we shall introduce) **is arguably necessary for ensemble to work**. Over certain data sets with no multi-view structure, ensemble *does not improve the test-time performance in deep learning* — despite having a non-convex training objective and different random seeds are used. See Figure 2. In contrast, real-life data sets such as CIFAR-10/100 do have the multi-view structure, moreover, standard neural networks such as ResNet do utilize this structure during the training process in the same way as we show in our theory. See Figure 6.

• For neural networks, **knowledge distillation has learned most of the features from the ensemble, and the use of *hard labels* to train individual models is a key for why ensemble works in deep learning**.

Specifically, as in Figure 3, if one evaluates an ensemble over models that are independently at random trained from knowledge distillation (i.e., using soft labels), its performance does not exceed the ensemble over the original single models. This means, models trained via knowledge distillation have learned most of the features from the ensemble, and has less variety comparing to the original models. We shall see this is consistent with our theoretical result.

|  | CIFAR10 test accuracy | | | | CIFAR100 test accuracy | | | |
|---|---|---|---|---|---|---|---|---|
|  | single model (over 10) | ensemble (over 10) | 10 runs of knowledge distill | ensemble over knowledge distill | single model (over 10) | ensemble (over 10) | 10 runs of knowledge distill | ensemble over knowledge distill |
| ResNet-28-2  | 95.22±*0.14%* | 96.33% | 95.89±0.07% | 96.21% | 76.38±*0.23%* | 81.13% | 78.94±0.21% | 80.35% |
| ResNet-34    | 93.65±*0.19%* | 94.97% | 94.37±0.13% | 94.88% | 71.66±*0.43%* | 76.85% | 73.57±0.34% | 75.60% |
| ResNet-34-2  | 95.45±*0.14%* | 96.55% | 96.00±0.12% | 96.42% | 77.01±*0.35%* | 81.48% | 79.43±0.23% | 81.56% |
| ResNet-16-10 | 96.08±*0.16%* | 96.80% | 96.73±0.07% | 96.76% | 80.03±*0.17%* | 83.18% | 82.51±0.14% | 83.36% |
| ResNet-22-10 | 96.44±*0.09%* | 97.12% | 97.01±0.09% | 97.09% | 81.17±*0.23%* | 84.33% | 83.54±0.19% | 84.27% |
| ResNet-28-10 | 96.70±*0.21%* | 97.20% | 97.06±0.08% | 97.24% | 81.51±*0.16%* | 84.69% | 83.75±0.16% | 84.87% |

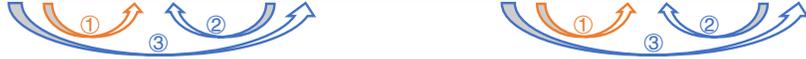

**Message ①**: an ensemble over *single models* (independently trained) can be distilled into a single model with moderate accuracy loss.
**Message ②**: an ensemble over models *after knowledge distillation does not improve accuracy by much* – in fact, not exceeding the ensemble accuracy of the original single models ③ – despite the training objective is still non-convex and different random seeds are used. This means, knowledge distillation models (i.e. simply matching the soft labels) *have learned most of the features* from the ensemble, and have less variety comparing to the original single models. This also means that "(huge) non-convexity" in neural networks and SGD with "different random seeds" *even together do not guarantee ensemble advantage unconditionally*; the structure of the data (and hard labels) is extremely important for ensemble to work as we mainly focus on in this paper.

Figure 3: Single models (+ their ensemble) vs. Knowledge distillations (+ their ensemble). Details in Appendix A.2.

## 2 Our Methodology and Intuition

### 2.1 Ensemble in Deep Learning v.s. Ensemble of Feature Mappings

The recent advance in deep learning theory shows that under certain circumstances, deep learning can be treated as a linear function over random feature mappings [3, 5, 6, 8, 9, 19, 24, 26, 27, 39, 40, 44, 54, 58, 85, 92]. In particular, the theory shows that when $f : \mathbb{R}^{D+d} \to \mathbb{R}$ is a neural network with inputs $x \in \mathbb{R}^d$ and weights $W \in \mathbb{R}^D$, in some cases, $f(W, x)$ can be approximated by:

$$f(W, x) \approx f(W_0, x) + \langle W - W_0, \nabla_W f(W_0, x) \rangle$$

where $W_0$ is the random initialization of the neural network, and $\Phi_{W_0}(x) := \nabla_W f(W_0, x)$ is the neural tangent kernel (NTK) feature mapping. This is known as the NTK approach. If this approximation holds, then training a neural network can be approximated by learning a linear function over $\Phi_{W_0}(x)$, which is very theory-friendly.

When training a linear function over such NTK features, traditional theorems [1, 14, 17, 80] suggest that the ensemble of $L$ independently trained models can indeed improve test-time performance, as it enlarges the feature space from $\Phi_{W_0}(x)$ to $\{\Phi_{W_0^{(i)}}(x)\}_{i \in [L]}$ for $L$ many independently sampled $W_0^{(i)}$. In this case, ensemble and knowledge distillation can be viewed as a feature selection process [7, 18, 65, 67, 73] to improve generalization. Motivate by this line of research, we ask:

*Can we understand ensemble and knowledge distillation in deep learning as feature selections? (in particular, using the NTK approach?)*

In this work, we provide empirical evidence towards **rejecting this point of view**, see Figure 1. In these experiments, we confirm that ensemble[1] of linear functions over NTK feature mappings with different random seeds: $f_i(x) = \langle W^{(i)}, \nabla_W f(W_0^{(i)}, x) \rangle$ does improve test accuracy, however, such improvement might be mainly due to a larger set of features in the ensemble model $\{\Phi_{W_0^{(i)}}(x)\}_{i \in [L]}$, whose combinations contain functions that generalize better.

To see this, we observe that an even superior performance (than the ensemble model) can simply be obtained by directly training $F(x) = f_1 + f_2 + \cdots + f_L$ from random initialization. *In contrast,* if $g_i(x)$'s are multi-layer neural networks with different random seeds, then training directly the sum of all the models $G(x) = \sum_{i \in [L]} g_i(x)$ does not give test accuracy gain anymore comparing to individual $g_i$'s, as now all the $g_i$'s *are capable of learning the same set of features*.

---
[1] As we mentioned, in this paper, we only focus on the simplest ensemble, i.e. taking an unweighted average of the output of independently trained models.

Furthermore, for linear functions over NTK feature mappings, we observe that the result obtained by ensemble can not be distilled at all into individual models, indicating the features selected by ensemble *is not contained* in the feature $\Phi_{W_0^{(i)}}(x)$ of any individual model. In contrast, in actual deep learning, ensemble *does not enlarge feature space*: an individual model $g_0(x)$ is still capable of learning the features of the ensemble model $G(x)$.

In sum, we believe to understand ensemble and knowledge distillation in deep learning, it is **inevitable** to study deep learning as a **feature learning process** (where features are trained), instead of feature selection process (where the features are prescribed and only the linear combinations are trained like in NTK). Moreover, we point out another fundamental difficulty here:

---
**Key challenge:**

If a single deep learning model is capable of— through knowledge distillation— learning the features of the ensemble model and achieving better test accuracy comparing to training the single model directly (and the same training accuracy, typically at global optimal of 100%), then why the single model *cannot learn* these features directly when we train the model to match the true data labels? What is the **dark knowledge** hidden in the output of ensemble (a.k.a. soft label)[2] comparing to the original hard label?

---

## 2.2 Ensemble in Deep Learning: a Feature Learning Process

Before addressing the key challenge, we point out that prior works are very limited with respect to studying neural network training as a feature learning process, due to the extreme non-convexity obstacle in optimization.

Most of the existing works proving that neural networks can learn features only focus on the case *when the input is Gaussian or Gaussian-like* [10, 12, 16, 37, 38, 45, 52, 53, 55, 55–57, 59, 68, 77–79, 83, 84, 86, 89]. However, as we demonstrate in Figure 2 on Page 4,

---
**Ensemble in DL might not improve test accuracy when inputs are Gaussian-like:**

Empirically, ensemble *does not* improve test accuracy in deep learning, in certain scenarios when the distribution of the input data is Gaussian or even mixture of Gaussians. This is true over various learner network structures (fully-connected, residual, convolution neural networks) and various labeling functions (when the labels are generated by linear functions, fully-connected, residual, convolutional networks, with/without label noise, with/without classification margin).

---

Some prior works also try to attribute the benefit of ensemble as reducing the *variance* of individual solutions [15, 61, 63, 81, 82] due to label noise or non-convex landscape of the training objective (so some individual models might simply not be trained very well by over-fitting to the label noise or stuck at a bad local minimal).

However, reducing variance can reduce test loss (typically cross-entropy), but not necessarily *test classification error*. As one example, taking the ensemble of random models can reduce test loss due to convexity of the cross-entropy loss, but the prediction of the ensemble is still random. As another example, the synthetic experiments in Figure 2 shows after applying ensemble over Gaussian-like inputs, the variance of the models is reduced while the test accuracy is *not improved*. We give many more empirical support to show that the variance (either from label noise or from the non-convex landscape) is usually not the cause for why ensemble works in deep learning, see Section 1.2. Moreover, we point out that (See Figure 1) in practice, typically the individual neural networks are trained *equally well*, meaning that they all have perfect training accuracy and almost identical test error, yet ensemble these models still improves the test accuracy significantly.

---

[2]For a $k$-class classification problem, the output of a model $g(x)$ is usually $k$-dimensional, and represents a soft-max probability distribution over the $k$ target classes. This is known as the *soft label*.

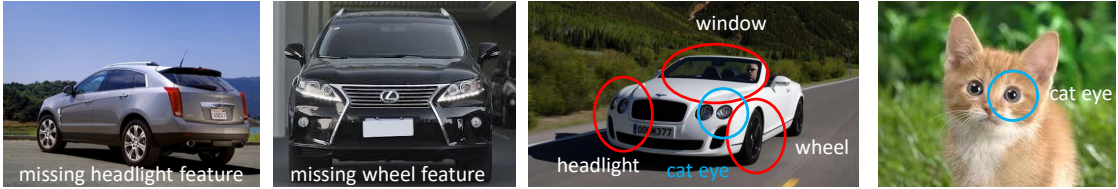

Figure 4: Illustration of images with multiple views (features) in the ImageNet dataset.

Hence, to understand the true benefit of ensemble in deep learning in theory, we would like to **consider a setting tied to practical deep learning**, where:

- The input distribution is more structured than standard Gaussian and there is no label noise. (From above discussions, ensemble cannot work for deep learning distribution-freely, nor even under Gaussian distribution).

- The individual neural networks all are well-trained, in the sense that the training accuracy in the end is 100%, and there is nearly no variance in the test accuracy for individual models. (So training and generalization never fails).

We would like to re-elaborate the *key challenge*: 'ensemble improves test accuracy' implies that different single models need to learn different sets of features; however, all these models have the same architecture, and trained using the same learning algorithm (SGD with momentum) with identical learning rates, and each the (learned) sets of features in each modellead to the perfect 100% training accuracy and an almost identical test accuracy. Thus, the difference of the features *must not be* due to 'difference in the data set', 'difference in the models', 'difference in the training algorithms', 'difference in the learning rates', 'failure in training occasionally', 'failure in generalization in some cases', etc. Additional principles need to be developed to incorporate the effect of ensemble in deep learning.

In this work, we propose to study a setting of data that we refer to as **multi-view**, where the above two conditions both hold when we train a two-layer neural networks with (smoothed-)ReLU activations. We also argue that the multi-view structure we consider is fairly common in the data sets used in practice, in particular for vision tasks. We give more details below.

### 2.3 Our Approach: Learning Multi-View Data

Let us first give a thought experiment to illustrate our approach, and we present the precise mathematical definition of the "multi-view" structure in Section 3. Consider a *binary* classification problem and four "features" $v_1, v_2, v_3, v_4$. The first two features correspond to the first class label, and the next two features correspond to the second class label. In the data distribution:

- When the label is class 1, then:[3]

$$\begin{cases} \text{both } v_1, v_2 \text{ appears with weight 1, one of } v_3, v_4 \text{ appears with weight 0.1} & \text{w.p. } 80\%; \\ \text{only } v_1 \text{ appears with weight 1, one of } v_3, v_4 \text{ appears with weight 0.1} & \text{w.p. } 10\%; \\ \text{only } v_2 \text{ appears with weight 1, one of } v_3, v_4 \text{ appears with weight 0.1} & \text{w.p. } 10\%. \end{cases}$$

---

[3]One can for simplicity think of "$v$ appears with weight $\alpha$ and $w$ appears with weight $\beta$" as data $= \alpha v + \beta w +$ noise.

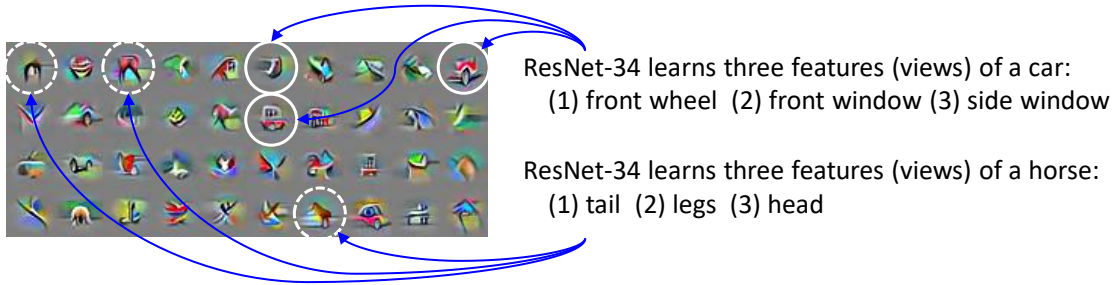

Figure 5: Visualization of the channels in layer-23 of a ResNet-34 trained on CIFAR-10, picture from [4].

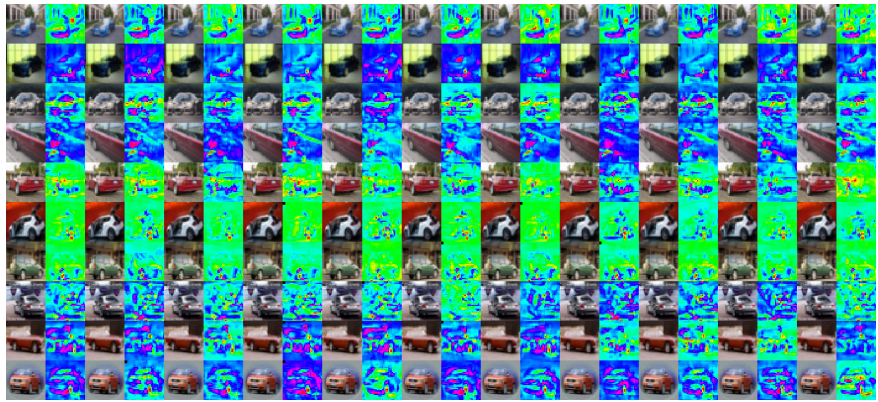

Figure 6: Ten independently trained ResNet-34 models (and their ensemble) detect car images through different reasonings, suggesting that the data has multi views, and independently trained neural networks do utilize this structure. The numerical experiments in Figure 7 also suggest the existence of multi views.

- When the label is class 2, then
$$\begin{cases} \text{both } v_3, v_4 \text{ appears with weight 1, one of } v_1, v_2 \text{ appears with weight 0.1} & \text{w.p. } 80\%; \\ \text{only } v_3 \text{ appears with weight 1, one of } v_1, v_2 \text{ appears with weight 0.1} & \text{w.p. } 10\%; \\ \text{only } v_4 \text{ appears with weight 1, one of } v_1, v_2 \text{ appears with weight 0.1} & \text{w.p. } 10\%. \end{cases}$$

We call the 80% of the data *multi-view data*: these are the data where multiple features exist and can be used to classify them correctly. We call the reset 20% of the data *single-view data*: some features for the correct labels are missing.

**Meaningfulness of our multi-view hypothesis.** Such "multi-view" structure is very common in many of the datasets where deep learning excels. In vision datasets in particular, as illustrated in Figure 4, a car image can be classified as a car by looking at the headlights, the wheels, or the windows. For a typical placement of a car in images, we can observe all these features, and it suffices to use one of the features to classify it as a car. However, there are some car images taken from a particular angle, where one or more of these features are missing. For example, an image of a car facing forward might be missing the wheel feature. Moreover, some car might also have a small fraction of "cat features": for example, the headlight might appear similar to cat eyes the ear of a cat. This can be used as the "dark knowledge" by the single model to learn from the ensemble.

In Figure 5, visualize the learned features from an actual neural network to show that they can indeed capture different views. In Figure 6, we plot the "heatmap" for some car images to illustrate that single models (trained from different random seeds) indeed pick up different parts of the input image to classify it as a car. In Figure 7, we manually delete for instance 7/8 of the channels in some intermediate layer of a ResNet, and show that the test accuracy may not be affected by much after ensemble— thus supporting that the multi-view hypothesis can indeed exist even in the intermediate layers of a neural network and ensemble is indeed collecting all these views.

| CIFAR100 | # input channels | | original | split to 2 | split to 4 | split to 8 | avg over 2 | avg over 4 | avg over 8 |
|---|---|---|---|---|---|---|---|---|---|
| ResNet-28 (a) | 16 | | 70.44±0.29% | 68.77±0.25% | 66.70±0.66% | - | 69.00±0.43% | 66.45±0.15% | - |
| ResNet-28 (b) | 32 | | 70.49±0.29% | 67.62±0.89% | 63.28±0.50% | - | 67.99±0.15% | 63.89±0.31% | - |
| ResNet-28-2 (a) | 32 | single | 76.09±0.23% | 74.50±0.68% | 72.47±1.78% | 70.84±1.32% | 75.31±0.23% | 73.69±0.34% | 71.60±0.34% |
| ResNet-28-2 (b) | 64 | model | 76.12±0.23% | 74.88±0.22% | 72.81±0.29% | 69.21±0.49% | 74.58±0.33% | 72.71±0.29% | 68.85±0.28% |
| ResNet-28-4 (a) | 64 | test | 79.10±0.18% | 78.57±0.29% | 77.94±0.43% | 76.88±0.35% | 78.42±0.35% | 78.14±0.16% | 77.52±0.20% |
| ResNet-28-4 (b) | 128 | accuracy | 78.53±0.16% | 77.72±0.20% | 76.62±0.29% | 74.93±0.40% | 77.95±0.22% | 76.88±0.33% | 75.17±0.25% |
| ResNet-28-10 (a) | 160 | | 81.23±0.23% | 81.03±0.17% | 80.53±0.09% | 80.12±0.26% | 80.58±0.28% | 81.06±0.22% | 80.63±0.22% |
| ResNet-28-10 (b) | 320 | | 80.76±0.27% | 80.41±0.24% | 80.09±0.16% | 79.02±0.22% | 80.54±0.23% | 80.03±0.16% | 79.38±0.27% |
| ResNet-28 (a) | 16 | | 75.52% | 74.07% | 73.63% | - | 74.05% | 70.98% | - |
| ResNet-28 (b) | 32 | | 74.47% | 73.58% | 72.17% | - | 71.97% | 68.03% | - |
| ResNet-28-2 (a) | 32 | ensemble | 80.33% | 79.73% | 79.58% | 78.75% | 79.24% | 78.19% | 76.31% |
| ResNet-28-2 (b) | 64 | model | 79.63% | 80.18% | 79.17% | 78.20% | 78.42% | 76.81% | 72.90% |
| ResNet-28-4 (a) | 64 | test | 82.64% | 82.81% | 82.56% | 82.24% | 82.26% | 82.12% | 81.71% |
| ResNet-28-4 (b) | 128 | accuracy | 81.84% | 82.06% | 81.89% | 81.74% | 81.28% | 80.63% | 79.14% |
| ResNet-28-10 (a) | 160 | | 84.05% | 84.08% | 83.65% | 83.51% | 83.79% | 84.12% | 83.69% |
| ResNet-28-10 (b) | 320 | | 83.10% | 83.40% | 83.81% | 83.53% | 83.21% | 83.00% | 82.19% |

Figure 7: Justify the multi-view hypothesis in practice. We regard some intermediate layer of a pre-trained ResNet as "input" with multiple channels (this pre-trained network stays fixed and shared for all individual models). Then, we train a new model either starting from this input (i.e. the "*original*" column), or from a fraction of the input (i.e., "*split into 4*" means using only 1/4 of the input channels), or from an average of the input (i.e., "*average over 4*" means averaging every four channels). Details in Appendix A.3.
**Observation 1.** Even when we significantly collapse the input channels (through averaging or throwing away most of them), most of the single model test accuracies do not drop by much. Moreover, it's known [64] that in ResNet, most channels are indeed learning different features (views) of the input, also see Figure 5 for an illustration. This indicates that many data can be classified correctly using different views.
**Observation 2.** Even when single model accuracy drops noticeably, ensemble accuracy does not change by much. We believe this is a strong evidence that there are multiple views in the data (even at intermediate layers), and **ensemble can collect all of them even when some models have missing views.**

**How individual neural networks learn.** Under the multi-view data defined above, if we train a neural network using the cross-entropy loss via gradient descent (GD) from random initialization, during the training process of the individual networks, we show that:

- The network will quickly pick up one of the feature $v \in \{v_1, v_2\}$ for the first label, and one of the features $v' \in \{v_3, v_4\}$ for the second label. So, 90% of the training examples, consisting of all the multi-view data and half of the single-view data (those with feature $v$ or $v'$), are classified correctly. Once classified correctly (with a large margin), these data begin to contribute negligible to gradient.

- Next, the network will memorize (using e.g. the noise in the data) the remaining 10% of the training examples without learning any new features, due to insufficient amount of left-over samples after the first phase, thus achieving training accuracy 100% but test accuracy 90%.

**How ensemble improves test accuracy.** We show that depending on the randomness of the initialization, each individual network will pick up $v_1$ or $v_2$ each with probability 0.5. Hence, we can prove that as long as we ensemble $\widetilde{O}(1)$ many independently trained models, w.h.p. the ensemble model will be able to pick up both features $\{v_1, v_2\}$, and both features $\{v_3, v_4\}$. Thus, all the data will be classified correctly.

**How knowledge distillation works.** Since ensemble learns all the features $v_1, v_2, v_3, v_4$, given a multi-view data with label 1, the ensemble will 1)$, where the 2 comes

from features $v_1, v_2$ and 0.1 comes from one of $v_3, v_4$. On the other hand, an individual model that only learns one of $v_3, v_4$ will actually output $\propto (2, 0)$ when the feature $v_3$ or $v_4$ in the data does not match the one learned by the model. Hence, by training the individual model to match the output of the ensemble, the individual model is *forced* to learn both features $v_3, v_4$, even though it has already perfectly classified the training data. **This is the "dark knowledge" hidden in the output of the ensemble model**.

(This theoretical finding is consistent with practice: Figure 3 suggests that models trained from knowledge distillation should have learned most of the features, and further computing their ensemble does not give much performance boost.)

## 2.4 Significance of Our Technique

Our work belongs to the generic framework where one can prove that certain aspects of the learning algorithm (in this paper, the randomness of the initialization) affects the order where the features are learned, which we believe is also one of the key ingredients to understand the role of the learning algorithm in terms of generalization in deep learning. This is **fundamentally different from convex optimization**, such as kernel method, where (with an $\ell_2$ regularization) there is an unique *global minimum* so the choice of optimization algorithm or the random seed of the initialization does not matter (thus, ensemble does not help at all). There are other works that consider other aspects, such as the choice of learning rate [58], that can affect the order where the features are learned in deep learning. In that work [58], the two "features" are asymmetric: a memorizable feature and a generalizable feature, so the learning rate will decide which feature to be picked. In our work, the features are "symmetric", so the randomness of the initialization will decide which feature to be picked. Our proof technique is fundamentally different from [58], as they only focus on the NTKs setting, where only a linear function over the prescribed (random) feature mapping is learned. As we have argued and shown empirically, the NTK setting cannot be used to explain ensemble and distillation in deep learning.

We believe that our workextends the reach of traditional optimization and statistical machine learning theory, where typically the statistics (generalization) is separated from optimization (training). As we have pointed out, such "separate" treatment might not be possible to understand (at least ensemble or knowledge distillation in) deep learning.

## 3 Problem Setup

In this paper, we consider the following data distribution with "multi-view", that allows us to formally prove our result on ensemble and knowledge distillation for two-layer neural networks. The data distribution is a straight-forward generalization of the intuitive setting in Section 2.3. For simplicity, in the main body, we use example choices of the parameters mainly a function of $k$ (such as $P = k^2$, $\gamma = \frac{1}{k^{1.5}}$, $\mu = \frac{k^{1.2}}{N}$, $\rho = k^{-0.01}$, $\sigma_0 = 1/\sqrt{k}$ as we shall see in this section), and we consider the case when $k$ is sufficiently large. In our formal statements of the theorems in the appendix, we shall give a much larger range of parameters for the theorems to hold.

### 3.1 Data Distribution and Notations

We consider learning a $k$-class classification problem over $P$-patch inputs, where each patch has dimension $d$. In symbols, each labelled data is represented by $(X, y)$ where $X = (x_1, x_2, \cdots, x_P) \in (\mathbb{R}^d)^P$ is the data vector and $y \in [k]$ is the data label. For simplicity, we focus on the case when $P = k^2$, and $d = \mathsf{poly}(k)$ for a large polynomial.

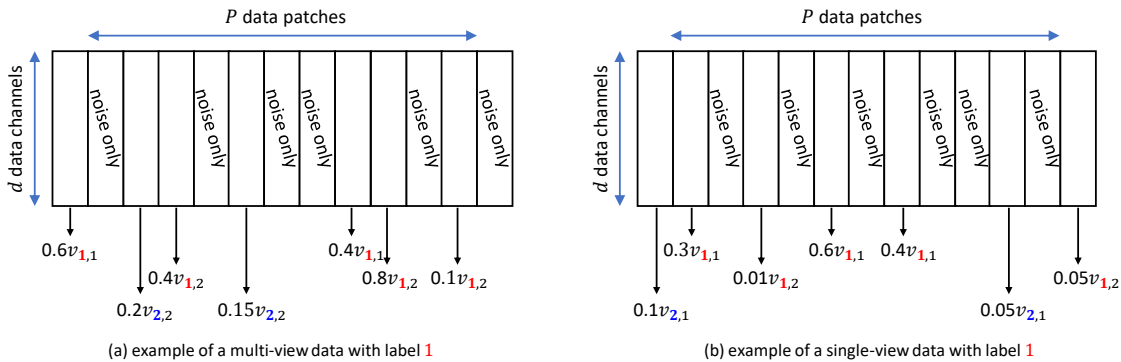

Figure 8: Illustration of a multi-view and a single-view data point; the feature vectors can also be combined with feature noise and random noise.

We consider the setting when $k$ is sufficiently large.[4] We use "w.h.p." to denote with probability at least $1 - e^{-\Omega(\log^2 k)}$, and use $\widetilde{O}, \widetilde{\Theta}, \widetilde{\Omega}$ notions to hide polylogarithmic factors in $k$.

We would like to assume that each class $j \in [k]$ has multiple associated features, say *two features* for simplicity of the math, represented by unit *feature vectors* $v_{j,1}, v_{j,2} \in \mathbb{R}^d$. For notation simplicity, we assume that all the features are orthogonal, although our work also extends to the "incoherent" case trivially. Namely,

$$\forall j, j' \in [k], \, \forall \ell, \ell' \in [k], \, \|v_{j,\ell}\|_2 = 1 \quad \text{and} \quad v_{j,\ell} \perp v_{j',\ell'} \text{ when } (j,\ell) \neq (j',\ell')$$

We denote by

$$\mathcal{V} \stackrel{\text{def}}{=} \{v_{j,1}, v_{j,2}\}_{j \in [k]} \quad \text{the set of all features.}$$

We now consider the following data and label distribution. Let $C_p$ be a global constant, $s \in [1, k^{0.2}]$ be a global parameter to control feature sparsity, $\sigma_p = \frac{1}{\sqrt{d}\,\text{polylog}(k)}$ be a parameter to control magnitude of the *random noise*, $\gamma = \frac{1}{k^{1.5}}$ be a parameter to control the *feature noise*. (Our proof in the appendix holds for a wider range of $\gamma$.)

To be concise, we define the *multi-view distribution* $\mathcal{D}_m$ and *single-view distribution* $\mathcal{D}_s$ together.

**Definition 3.1** (data distributions $\mathcal{D}_m$ and $\mathcal{D}_s$). *Given $\mathcal{D} \in \{\mathcal{D}_m, \mathcal{D}_s\}$, we define $(X, y) \sim \mathcal{D}$ as follows. First choose the label $y \in [k]$ uniformly at random. Then, the data vector $X$ is generated as follows (see also an illustration in Figure 8).*

1. *Sample a set $\mathcal{V}'$ of features uniformly at random from $\{v_{j',1}, v_{j',2}\}_{j' \neq y}$, each with probability $\frac{s}{k}$.*

2. *Denote $\mathcal{V}(X) = \mathcal{V}' \cup \{v_{y,1}, v_{y,2}\}$ as the set of feature vectors used in data $X$.*

3. *For each $v \in \mathcal{V}(X)$, pick $C_p$ many disjoint patches in $[P]$ and denote it as $\mathcal{P}_v(X)$ (the distribution of these patches can be arbitrary). We denote $\mathcal{P}(X) = \cup_{v \in \mathcal{V}(X)} \mathcal{P}_v(X)$.*

4. *If $\mathcal{D} = \mathcal{D}_s$ is the single-view distribution, pick a value $\widehat{\ell} = \widehat{\ell}(X) \in [2]$ uniformly at random.*

5. *For each $p \in \mathcal{P}_v(X)$ for some $v \in \mathcal{V}(X)$, we set*

$$x_p = z_p v + \sum_{v' \in \mathcal{V}} \alpha_{p,v'} v' + \xi_p$$

*Above, each $\alpha_{p,v'} \in [0, \gamma]$ is the feature noise, and $\xi_p \sim \mathcal{N}(0, \sigma_p^2 \mathbf{I})$ is an (independent) random Gaussian noise. The coefficients $z_p \geq 0$ satisfy that:*

---

[4] If we want to work with fixed $k$, say $k = 2$, our theorem can also be modified to that setting by increasing the number of features per class. In this case, a subset of features per class will be learned by each individual neural network. We keep our current setting with two features to simplify the notations.

*In the case of multi-view distribution $\mathcal{D} = \mathcal{D}_m$*

- $\sum_{p \in \mathcal{P}_v(X)} z_p \in [1, O(1)]$ when $v \in \{v_{y,1}, v_{y,2}\}$,
  and the marginal distribution of $\sum_{p \in \mathcal{P}_v(X)} z_p$ is *left-close*;[5]
- $\sum_{p \in \mathcal{P}_v(X)} z_p \in [\Omega(1), 0.4]$ when $v \in \mathcal{V}(X) \setminus \{v_{y,1}, v_{y,2}\}$,
  and the marginal distribution of $\sum_{p \in \mathcal{P}_v(X)} z_p$ is *right-close*.[6]

*In the case of single-view distribution $\mathcal{D} = \mathcal{D}_s$*

- $\sum_{p \in \mathcal{P}_v(X)} z_p \in [1, O(1)]$ when $v = v_{y,\widehat{\ell}}$,
- $\sum_{p \in \mathcal{P}_v(X)} z_p \in [\rho, O(\rho)]$ when $v = v_{y,3-\widehat{\ell}}$,     *we consider $\rho = k^{-0.01}$ for simplicity*
- $\sum_{p \in \mathcal{P}_v(X)} z_p \in [\Omega(\Gamma), \Gamma]$ when $v \in \mathcal{V}(X) \setminus \{v_{y,1}, v_{y,2}\}$.     *we consider $\Gamma = \frac{1}{\mathsf{polylog}(k)}$ for simplicity*

6. For each $p \in [P] \setminus \mathcal{P}(X)$, we set:

$$x_p = \sum_{v' \in \mathcal{V}} \alpha_{p,v'} v' + \xi_p$$

where each $\alpha_{p,v'} \in [0, \gamma]$ is the feature noise and $\xi_p \sim \mathcal{N}(0, \frac{\gamma^2 k^2}{d}\mathbf{I})$ is the (independent) Gaussian noise.

**Remark 3.2.** The distribution of how we pick $\mathcal{P}(X)$ and how to assign $\sum_{p \in \mathcal{P}_v(X)} z_p$ to each patch in $p \in \mathcal{P}_v(X)$ can be arbitrary (and can depend on other randomness in the data as well). Except the marginal distributions of the sum of some $z_p$'s are left-close or right-close, but we also do not have other restrictions on how $z_p$'s are distributed within the summation. In particular, we have allowed **different features** $v_{j,1}$, $v_{j,2}$ to show up with **different weights** in the data (for example, for multi-view data, some view $v_{y,1}$ can consistently have larger $z_p$ comparing to $v_{y,2}$.) Yet, we shall prove that the order to learn these features by the learner network *can still be flipped* depending on the randomness of network initialization. We also do not have any restriction on the distribution of the feature noise $\alpha_{p,v'}$ (they can depend on other randomness of the data distribution as well).

**Generality and significance of our data distribution.** Our setting is *very common and in fact a down-sized version of convolution networks* applied to image classification data. With a small kernel size, the good features in an image typically appears only on a few patches,[7] and most other patches are simply random noise or low-magnitude feature noises that are more or less irrelevant to the label.

More importantly, the above concept class (namely, labeled data distribution in Def. 3.1) is *not learnable by linear classifiers or constant degree polynomials*. Indeed, if we only use a linear classifier, then the total accumulated (low-magnitude) feature noise from all patches can be as large as $\gamma P \gg 1$ by our choice of $\gamma = \frac{1}{k^{1.5}}$ and $P = k^2$. This is much larger than the magnitude of the signal. On the other hand, by Markov brother's inequality, low-degree polynomials also lack the power to be approximately linear (to fit the signal) when the input is large, while being sub-linear when there are low-magnitude feature noises. We also conjecture that one can prove this concept class is not efficiently learnable by kernel methods in general, using the recent development of kernel lower bounds [2, 4]. Thus, we believe a (convolution) neural network with ReLU-like activation is in some sense necessary to learn this concept class.

---

[5] We say a distribution $p$ over a real interval $[a, b]$ for constants $a, b$ is *left-close*, if there is a $\varepsilon \leq \frac{1}{\mathsf{polylog}(k)}$ such that $\mathbf{Pr}_{z \sim p}[z \leq a + \varepsilon] \geq \frac{1}{\mathsf{polylog}(k)}$, and is *right-close* if $\mathbf{Pr}_{z \sim p}[z \geq b - \varepsilon] \geq \frac{1}{\mathsf{polylog}(k)}$. For instance, here $Z = \sum_{p \in \mathcal{P}_v(X)} z_p$ can be a uniform distribution over $[1, 2]$. This assumption is simply to avoid the case when the distribution is too screwed (for example with very small probability its $b$ and very large probability its $a$).

[6] For instance, $Z = \sum_{p \in \mathcal{P}_v(X)} z_p$ can be a uniform distribution over $[0.2, 0.4]$.

[7] For example, in image classification when the image is of size $64 \times 64$, at the first layer, each patch can be a sub-image of size $d = 48 = 4 \times 4 \times 3$ (3 RGB channels), and there are 256 patches. At the second layer, we can have higher dimension $d$ per patch, such as $d = 4 \times 4 \times 64$ when more channels are introduced. In convolution networks, there are typically over-laps between patches, we point out that our setting is more general: In fact, for example for a data $X = (a, b, c, d)$ with patches $(a, b, c)$ and $(b, c, d)$, we can simply define $x_1 = (a, b, c)$ and $x_2 = (b, c, d)$. Moreover, our $X$ can also be viewed as intermediate output of the previous convolution layer in a convolution network.

Our final data distribution $\mathcal{D}$, and the training data set $\mathcal{Z}$ are formally given as follows.

**Definition 3.3** ($\mathcal{D}$ and $\mathcal{Z}$). *We assume that the final distribution $\mathcal{D}$ consists of data from $\mathcal{D}_m$ w.p. $1 - \mu$ and from $\mathcal{D}_s$ w.p. $\mu$. We are given $N$ training samples from $\mathcal{D}$, and denote the training data set as $\mathcal{Z} = \mathcal{Z}_m \cup \mathcal{Z}_s$ where $\mathcal{Z}_m$ and $\mathcal{Z}_s$ respectively represent multi-view and single-view training data. We write $(X, y) \sim \mathcal{Z}$ as $(X, y)$ sampled uniformly at random from the empirical data set, and denote $N_s = |\mathcal{Z}_s|$. We again for simplicity focus on the setting when $\mu = \frac{1}{\mathsf{poly}(k)}$ and we are given samples $N = k^{1.2}/\mu$ so each label $i$ appears at least $\widetilde{\Omega}(1)$ in $\mathcal{Z}_s$. Although our result trivially applies to other choices of $N$.*

## 3.2 Learner Network

We consider a learner network using the following smoothed ReLU activation function $\widetilde{\mathsf{ReLU}}$:

**Definition 3.4.** *For an integer $q \geq 2$ and a threshold $\varrho = \frac{1}{\mathsf{polylog}(k)}$, the smoothed ReLU function*

$$\widetilde{\mathsf{ReLU}}(z) \stackrel{\text{def}}{=} \begin{cases} 0 & \text{if } z \leq 0; \\ \frac{z^q}{q\varrho^{q-1}} & \text{if } z \in [0, \varrho]; \\ z - (1 - \frac{1}{q})\varrho & \text{if } z \geq \varrho \end{cases}$$

Since $\widetilde{\mathsf{ReLU}}$ has continuous gradient and we denote it as $\widetilde{\mathsf{ReLU}}'(z)$. For simplicity, we focus on the choice $q = 4$, although our result applies to other constants $q \geq 3$ (see appendix) or most other forms of smoothing trivially. As mentioned in previous section, the (smoothed) ReLU activation has a desired property such that $\widetilde{\mathsf{ReLU}}(z)$ is linear when $z$ is large, but becomes much smaller when $z$ is small. This allows the network to effectively reduce the impact of low-magnitude feature noises from the input patches for better classification.

The *learner network* $F(X) = (F_1(X), \ldots, F_k(X)) \in \mathbb{R}^k$ is a two-layer convolutional network parameterized by $w_{i,r} \in \mathbb{R}^d$ for $i \in [k], r \in [m]$, satisfying

$$\forall i \in [k]: \quad F_i(X) = \sum_{r \in [m]} \sum_{p \in [P]} \widetilde{\mathsf{ReLU}}(\langle w_{i,r}, x_p \rangle)$$

Although there exists ground truth network with $m = 2$ that can classify the data correctly (e.g. $w_{i,r} = v_{i,r}$ for $r \in [2]$), in this paper, for efficient optimization purpose it is convenient to work on a moderate level of over-parameterization: $m \in [\mathsf{polylog}(k), k]$, which people also typically use in practice to train neural networks.

**Training a Single Model.** We learn the concept class (namely, the labeled data distribution) using gradient descent with learning rate $\eta > 0$, over the cross-entropy loss function $L$ using $N$ training data points $\mathcal{Z} = \{(X_i, y_i)\}_{i \in [N]}$. We denote the empirical loss as:

$$L(F) = \frac{1}{N} \sum_{i \in [N]} L(F; X_i, y_i) = \mathbb{E}_{(X,y) \sim \mathcal{Z}}[L(F; X, y)]$$

where $L(F; X, y) = -\log \frac{e^{F_y(X)}}{\sum_{j \in [k]} e^{F_j(X)}}$. We **randomly initialize** the network $F$ by letting each $w_{i,r}^{(0)} \sim \mathcal{N}(0, \sigma_0^2 I)$ for $\sigma_0^2 = \frac{1}{k}$, which is the most standard initialization people use in practice.

More specifically, to train a single model, at each iteration $t$ we update using gradient descent (GD):

$$w_{i,r}^{(t+1)} \leftarrow w_{i,r}^{(t)} - \eta \, \mathbb{E}_{(X,y)\sim\mathcal{Z}} \nabla_{w_{i,r}} L(F^{(t)}; X, y) \tag{3.1}$$

We run the algorithm for $T = \frac{\mathsf{poly}(k)}{\eta}$ many iterations.[8] We use $F^{(t)}$ to denote the model $F$ with hidden weights $\{w_{i,r}^{(t)}\}$ at iteration $t$.

**Notations.** We denote by $\mathbf{logit}_i(F, X) \stackrel{\text{def}}{=} \frac{e^{F_i(X)}}{\sum_{j\in[k]} e^{F_j(X)}}$. Using this, we can write down

$$\forall i \in [k], r \in [m]: \quad -\nabla_{w_{i,r}} L(F; X, y) = (\mathbb{1}_{i\neq y} - \mathbf{logit}_i(F, X))\nabla_{w_{i,r}} F_i(X) \ .$$

## 4 Main Theorems and Explanations

We now state the main theorems in this paper.[9] Recall the learner network and its learning process are given in Section 3.2, and the data distribution is in Section 3.1.

**Theorem 1** (single model)**.** *For every sufficiently large $k > 0$, every $m \in [\mathsf{polylog}(k), k]$, every $\eta \leq \frac{1}{\mathsf{poly}(k)}$, suppose we train a single model using the gradient descent update (3.1) starting from the random initialization defined in Section 3.2, then after $T = \frac{\mathsf{poly}(k)}{\eta}$ many iterations, with probability $\geq 1 - e^{-\Omega(\log^2 k)}$, the model $F^{(T)}$ satisfies:*

- *(training accuracy is perfect): meaning for all $(X, y) \in \mathcal{Z}$, all $i \in [k]\setminus\{y\}$: $F_y^{(T)}(X) > F_i^{(T)}(X)$.*
- *(test accuracy is bad but consistent): meaning that:*

$$\Pr_{(X,y)\sim\mathcal{D}}[\exists i \in [k] \setminus \{y\}: F_y^{(T)}(X) < F_i^{(T)}(X)] \in [0.49\mu, 0.51\mu] \ .$$

We shall give technical intuitions about why Theorem 1 holds in Appendix B. But, at a high-level, we shall construct a "lottery winning" set $\mathcal{M} \subseteq [k] \times [2]$ of cardinality $|\mathcal{M}| \in [k(1 - o(1)), k]$. It only depends on the random initialization of $F$. Then, with some effort we can prove that, for every $(i, \ell) \in \mathcal{M}$, at the end of the training $F^{(T)}$ will learn feature $v_{i,\ell}$ but not learn feature $v_{i,3-\ell}$. This means for those single-view data $(X, y)$ with $y = i$ and $\widehat{\ell}(X) = 3 - \ell$, the final network $F^{(T)}$ will predict its label wrong. This is why the final test accuracy is around $0.5\mu$.

Note the property that test accuracy consistently belongs to the range $[0.49\mu, 0.51\mu]$ should be reminiscent of message ⑤ in Figure 1, where multiple single models, although starting from different random initialization, in practice does have a relatively small variance in test accuracies.

**Ensemble.** Suppose $\{F^{[\ell]}\}_{\ell\in[K]}$ are $K = \widetilde{\Omega}(1)$ independently trained models of $F$ with $m = \mathsf{polylog}(k)$ for $T = O\big(\frac{\mathsf{poly}(k)}{\eta}\big)$ iterations (i.e., the same setting as Theorem 1 except we only need a small over-parameterization $m = \mathsf{polylog}(k)$). Let us define their ensemble

$$G(X) = \frac{\widetilde{\Theta}(1)}{K} \sum_{\ell} F^{[\ell]}(X) \tag{4.1}$$

Our next theorem states that the ensemble model has much higher test accuracy.

---

[8]Our result also trivially extends to the case when there is a weight decay (i.e. $\ell_2$ regularizer): $w_{i,r}^{(t+1)} \leftarrow (1 - \eta\lambda)w_{i,r}^{(t)} - \eta \, \mathbb{E}_{(X,y)\sim\mathcal{Z}} \nabla_{w_{i,r}} L(F^{(t)}; X, y)$ as long as $\lambda$ is not too large. We keep this basic version without weight decay to simplify the analysis.

[9]We shall restate these theorems in the appendix with more details and wider range of parameters.

**Theorem 2** (ensemble). *In the same setting as above, we have for the ensemble model $G$ in (4.1), with probability at least $1 - e^{-\Omega(\log^2 k)}$:*

- *(training accuracy is perfect): meaning for all $(X, y) \in \mathcal{Z}$, for all $i \in [k] \setminus \{y\}$: $G_y(X) > G_i(X)$.*
- *(test accuracy is good): meaning that:*

$$\Pr_{(X,y) \sim \mathcal{D}}[\exists i \in [k] \setminus \{y\} \colon G_y(X) < G_i(X)] \leq 0.001\mu \ .$$

As we discussed in Section 2.3, the reason Theorem 2 holds attributes to the fact that those lottery winning sets $\mathcal{M}$ depend on the random initialization of the networks; and therefore, when multiple models are put together, their "union" of $\mathcal{M}$ shall cover all possible features $\{v_{i,\ell}\}_{(i,\ell) \in [k] \times [2]}$. Moreover, our theorem only requires individual $K = \widetilde{\Omega}(1)$ models for ensemble, which is indeed "averaging the output of *a few* independently trained models".

**Roadmap.** We shall restate and prove the general versions of Theorem 1 and 2 in Appendix D, after establishing core lemmas in Appendix B and C.

### 4.1 Knowledge Distillation for Ensemble

We consider a knowledge distillation algorithm given the existing ensemble model $G$ (see (4.1)) as follows. For every label $i \in [k]$, let us define the truncated scaled logit as (for $\tau = \frac{1}{\log^2 k}$):

$$\mathbf{logit}_i^\tau(F, X) = \frac{e^{\min\{\tau^2 F_i(X), 1\}/\tau}}{\sum_{j \in [k]} e^{\min\{\tau^2 F_j(X), 1\}/\tau}} \tag{4.2}$$

(This logit function should be reminiscent of the logit function with temperature used by the original knowledge distillation work [42]; we use truncation instead which is easier to analyze.)

Now, we train a new network $F$ from random initialization (where the randomness is independent of all of those used in $F^{[\ell]}$). At every iteration $t$, we update each weight $w_{i,r}$ by:

$$w_{i,r}^{(t+1)} = w_{i,r}^{(t)} - \eta \nabla_{w_{i,r}} L(F^{(t)}) - \eta' \mathop{\mathbb{E}}_{(X,y) \sim \mathcal{Z}} \left[ \left( \mathbf{logit}_i^\tau(F^{(t)}, X) - \mathbf{logit}_i^\tau(G, X) \right)^- \nabla_{w_{i,r}} F_i^{(t)}(X) \right] \tag{4.3}$$

**Notation.** Throughout the paper we denote by $[a]^+ = \max\{0, a\}$ and $[a]^- = \min\{0, a\}$.

This knowledge distillation method (4.3) is almost identical to the one used in the original work [42], except we use a truncation during the training to make it more (theoretically) stable. Moreover, we update the distillation objective using a larger learning rate $\eta'$ comparing to $\eta$ of the cross-entropy objective. This is also consistent with the training schedule used in [42].

Let $F^{(t)}$ be the resulting network obtained by (4.3) at iteration $t$. We have the following theorem:

**Theorem 3** (ensemble distillation). *Consider the distillation algorithm (4.3) in which $G$ is the ensemble model defined in (4.1). For every $k > 0$, for $m = \mathsf{polylog}(k)$, for every $\eta \leq \frac{1}{\mathsf{poly}(k)}$, setting $\eta' = \eta \mathsf{poly}(k)$, after $T = \frac{\mathsf{poly}(k)}{\eta}$ many iterations with probability at least $1 - e^{-\Omega(\log^2 k)}$, for at least 90% of the iterations $t \leq T$:*

- *(training accuracy is perfect): meaning for all $(X, y) \in \mathcal{Z}$, all $i \in [k] \setminus \{y\}$: $F_y^{(t)}(X) > F_i^{(t)}(X)$.*
- *(test accuracy is good): meaning that:*

$$\Pr_{(X,y) \sim \mathcal{D}}[\exists i \in [k] \setminus \{y\} \colon F_y^{(t)}(X) < F_i^{(t)}(X)] \leq 0.001\mu \ .$$

We shall restate the general version of Theorem 3 in Appendix E, and prove it in Appendix F.

*Remark.* Theorem 3 necessarily means that the distilled model $F$ has learned all the features $\{v_{i,\ell}\}_{(i,\ell)\in[k]\times[2]}$ from the ensemble model $G$. This is consistent with our empirical findings in Figure 3: if one trains multiple individual models using knowledge distillation with different random seeds, then their ensemble gives no further performance boost.

### 4.2 Self Knowledge Distillation as Implicit Ensemble

Self distillation [35, 62] refers to training a single model to match the output of another single model. In this paper we also show that self-distillation can also improve test accuracy under our multi-view setting. Let us consider the following self-distillation algorithm.

Let us now consider $F = F^{(T)}, G = G^{(T)}$ be two single models trained in the same setting as Theorem 1 using independent random initializations (for simplicity, we override to notation a bit, so here $G$ is a single model to be distilled from, instead of the ensemble). We scale them up by a small factor $\widetilde{\Theta}(1)$ similar to (4.1). Then, starting from $F^{(T)}$, we apply the following updates for another $T'$ iterations:

$$w_{i,r}^{(t+1)} = w_{i,r}^{(t)} - \eta \mathop{\mathbb{E}}_{(X,y)\sim\mathcal{Z}} \left( \left(\mathbf{logit}_i^\tau(F^{(t)}, X) - \mathbf{logit}_i^\tau(G, X)\right)^- \nabla_{w_{i,r}} F_i^{(t)}(X) \right) \quad (4.4)$$

This objective is considered as "self-distillation" since $G$ is an individual model (trained using an identical manner as $F$, only from a different random initialization). In particular, if $F = G$, then $\mathbf{logit}_i^\tau(F^{(t)}, X) - \mathbf{logit}_i^\tau(G, X) = 0$ so the weights are no longer updated. However, as we will actually prove, this training objective will actually learn an $F$ that *has better generalization comparing to $G$*.

This time, *for simplicity*, let us make the following additional assumption on the data:

**Assumption 4.1** (balanced $\mathcal{D}_m$). *In Def. 3.1, for multi-view data $(X, y)$, we additionally assume that the marginal distributions of $\sum_{p \in \mathcal{P}_v(X)} z_p^q \in [1, 1 + o(1)]$ for $v \in \{v_{y,1}, v_{y,2}\}$.*

The rationale for this assumption is quite simple. Suppose $\mathcal{M}$ is the aforementioned "lottery winning" set of training a single model without knowledge distillation. Assumption 4.1 will ensure that each $(i, 1)$ and $(i, 2)$ will belong to $\mathcal{M}_F$ with relatively equal probability. If we train two models $F$ and $G$, their combined lottery winning set $\mathcal{M}_F \cup \mathcal{M}_G$ shall be of cardinality around $\frac{3}{2}k(1-o(1))$. Therefore, if we can distill the knowledge of $G$ to $F$, the test accuracy can be improved from $1 - \frac{1}{2}\mu$ to $1 - \frac{1}{4}\mu$. See the following theorem:[10]

**Theorem 4** (self-distillation). *Under this additional Assumption 4.1, consider the distillation algorithm (4.4) where $G$ is an independently trained single model (in the same setting as Theorem 1). For every $k > 0$, every $m \in [\mathsf{polylog}(k), k]$, every $\eta \leq \frac{1}{\mathsf{poly}(k)}$, after $T' = \frac{\mathsf{poly}(k)}{\eta}$ many iterations of algorithm (4.4), with probability at least $1 - e^{-\Omega(\log^2 k)}$:*

- *(training accuracy is perfect): meaning for all $(X, y) \in \mathcal{Z}$, all $i \neq y$: $F_y^{(T+T')}(X) > F_i^{(T+T')}(X)$.*
- *(test accuracy is better): meaning that:*

$$\mathop{\mathbf{Pr}}_{(X,y)\sim\mathcal{D}}[\exists i \in [k] \setminus \{y\} \colon F_y^{(T+T')}(X) < F_i^{(T+T')}(X)] \leq 0.26\mu$$

Recall from Theorem 1, all individual models should have test error at least $0.49\mu \gg 0.26\mu$. Hence the model $F$ generalizes better (comparing to both the original model $F$ before self-distillation and the individual model $G$) after self-distillation to the individual model $G$. We shall restate the general version of Theorem 4 in Appendix E, and prove it in Appendix F.

---

[10]One can trivially relax Assumption 4.1 so that the two views have different distributions but with a constant ratio between their expectations; in this way the improved accuracy is no longer $1 - \frac{1}{4}\mu$ but shall depend on this constant ratio. For simplicity of this paper, we do not state the result of that more general case.

**Why does self-distillation improve test accuracy? Self-distillation is performing implicit ensemble + knowledge distillation.** As the main idea of behind the proof, we actually show that self-distillation is performing implicit ensemble together with knowledge distillation. In particular, let $\mathcal{M}_F, \mathcal{M}_G \subseteq V$ be the features learned by individual models $F, G$ starting from (independent) random initializations $W_F^{(0)}$ and $W_G^{(0)}$ respectively when trained on the original data set, now, if we further train the individual model $F$ to match the output of individual model $G$, $F$ is actually going to learn a larger set of features $\mathcal{M}_G \cup \mathcal{M}_F$, where features in $\mathcal{M}_F$ come from gradient of the original objective, and features in $\mathcal{M}_G$ come from the gradient of the knowledge distillation objective w.r.t. $G$. This is equivalent to first ensemble $F$ and $G$, then train an additional model $H$ from random initialization to match the ensemble— Self-distillation implicitly merge "ensemble individual models $F$, $G$ and distill the ensemble to another individual model $H$" into "ensemble individual models $F$, $G$ and distill the ensemble to the individual model $F$" since $F$ and $H$ have the same structure. Then eventually it is merged directly into "training an individual model $F$ via self-distillation to match the output of an individual model $G$".

## 5 Discussion: Ensemble and Knowledge Distillation

In this work, we show, to the best of our knowledge, the first theoretical result towards understanding how ensemble work in deep learning. As our main contribution, we provide empirical evidence that ensemble might work very differently in deep learning comparing to ensemble in random feature models. Moreover, ensemble does not always improve test accuracy in deep learning, especially when the input data comes from Gaussian-like distribution.

Motivated by these empirical observations, we propose a generic structure of the data we refer to as multi-view, and prove that ensemble improves test accuracy for two-layer neural networks in this setting. Moreover, we also prove that ensemble model can be distilled into a single model. Meaning that, through training a single model to "simulate" the output of the ensemble over the same training data set, single model is able to match the test accuracy of the ensemble, and thus being superior to any single model that is clean, directly trained on the original data's labels.

We believe that our framework can be applied to other settings as well, for example, data augmentation using random cropping could be potentially regarded as another way to enforce the network to learn "multi-views". We hope that our new theoretical insights on how neural networks pick up features during training can also help in practice design new, principled approach to improve test accuracy of a neural network, potentially matching that of the ensemble.

# Appendix I: Experiment Details

## A  Experiment Details

Our real-life experiments use the CIFAR-10/100 datasets [49]. The SimpleCNN architecture we have used comes from [8], and the (pre-activation) ResNet architecture we have used comes from the wide resnet work [87]. For instance, SimpleCNN-10-3 stands for the 10-layer architecture in [8] but widened by a factor of 3; and ResNet-34-2 stands for the 34-layer wide resnet architecture in [87] and the widening factor is 2.

For training regular neural networks, it is well-known that SGD with momentum and 0.1 learning rate is a state-of-the-art training method. We use batch size 125, train for 140 epochs, and decay the learning rate thrice at epochs 80, 100 and 120 each by a factor 0.2.[11] We use standard random crop, random flip, normalization, and cutout augmentation [76] for the training data.

For training neural-kernel models (NTKs), we find Adam a better training algorithm with an initial learning rate 0.001. We use batch size 50, train for 200 epochs, and decay the learning rate twice at epochs 140 and 170 each by a factor 0.2. We use ZCA data preprocessing which has been reported very helpful for improving neural kernel methods' performance together with cutout augmentation [76].[12]

### A.1  Real-life Data: Single Model vs Ensemble vs Distillation

For our experiment in **Figure 1**, we compare the performance of neural kernel methods vs. real neural networks on the standard CIFAR-10/100 datasets.

When presenting the single-model accuracies in Figure 1, we simply run the training algorithms 10 times from independently randomly initialized seeds. The NTK models we present the best accuracies among the 10 runs, and for ResNet models we present the mean and standard deviations.

When presenting the ensemble accuracies in Figure 1, we simply take an average of the 10 independently training models' outputs and use that to predict test labels.

When presenting the "directly train $\sum_\ell f_\ell$" result in Figure 1, we directly train a larger network consists of averaging 10 single models (separately, independently initialized). We use the same training algorithm as that for training single models. For some of the NTK models, our 16GB GPU memory sometimes only allows us to train an average of *fewer than 10 single models*; and when we do so, we have put a $\diamond$ remark in Figure 1.[13]

When presenting the "knowledge distillation" result in Figure 1, we adopt the original knowledge distillation objective of [42]. It is very similar to (4.3) that we used in our theoretical proof (see (4.3). It has a weight parameter for the ratio between standard cross-entropy vs the distillation objective (known as $\frac{\eta'}{\eta}$ in (4.3)), and they have a temperature parameter that controls the distillation objective (that is very similar to our $\tau$ parameter in (4.3)). We have tuned both parameters in a reasonable range to get the best distillation accuracy.

When presenting the "self-distillation" result in Figure 1, we divide the training process of a single model into two stages: in the first stage it uses the original cross-entropy loss with hard training labels and records the best model in the checkpoints, and in the second stage it trains another single model from random initialization using the distillation objective of [42] to match the output of the previously recorded best model.

---

[11]Some standard training batch for such parameter settings can be found on https://github.com/bearpaw/pytorch-classification.

[12]ZCA data preprocessing does not help regular neural net training.

[13]In the case of ResNet16-5-NTK and ResNet10-10-NTK, it only allows us to train an average of 2 models; since this is somewhat meaningless, we simply report "out of memory" in Figure 1.

*Remark* A.1. We confirm two more experimental findings that we did not include in Figure 1. First, one can repeat self-distillation multiple times but the test accuracy gain becomes very incremental. Second, one can alternatively use a three-stage process for self-distillation like we did in our theoretical result (see Section 4.2): namely, train two independent single models $F$ and $G$, and then continue to train $G$ by distilling it to match the output of $F$. The resulting test accuracy is extremely close to that of the two-stage process.

### A.2  Real-life Data: Ensemble over Distillations of Ensemble

For our experiment in**Figure 3**, we have studied the process of (1) training 10 independent single models, (2) evaluating their ensemble, (3) training 10 independent single models using knowledge distillations to match the outputs of (2), and (4) evaluating their ensemble.

The process of (1) and (2) are identical to that in Section A.1.

To present (3), we first apply parameter tuning for the knowledge distillation objective (see Section A.1). Then, we fix the best-selected parameters and perform knowledge distillation 10 times. In other words, these 10 runs differ *only* in the random seeds used in their initialization and SGD, but are *identical* in learning rate, weight decay, knowledge distillation parameters, and all other parameters.

Finally, (4) is a simple (unweighted) average over the 10 models produced by (3).

### A.3  Real-life Data: Justifying the Multi-View Assumption

We also perform an experiment in**Figure 7** to justify that in real-life training, there is strong evidence that there are multiple views of the data— even at some intermediate layers— to justify the image labels.

Recall that ResNet has three blocks of layers. In the (a) version of the experiment, we take a pre-trained model, and view its output at the end of the *first* block as "input", to train a new model where the trainable parameters are the second and third blocks. In the (b) version of the experiment, ew view the pre-trained model's output at the end of the *second* block as "input", to train a new model where the trainable parameters are in the third block only.

Specifically, we consider ResNet-28-$M$ version (a) and (b) for $M \in \{1, 2, 4, 10\}$. For instance, the new "input" has $N = 32$ channels for the case of "ResNet-28-2 version (a)", and has $N = 320$ channels for the case of "ResNet-28-10 version (b)."

For each of the settings above, we

- split the input into 8 chunks (with $N/8$ channels) and train 1 model each, totaling 8 models;
- split the input into 4 chunks (with $N/4$ channels) and train 2 models each, totaling 8 models;
- split the input into 2 chunks (with $N/2$ channels) and train 4 models each, totaling 8 models;
- average the input into $N/8$ channels (by averaging over every 8 channels) and train 8 models;
- average the input into $N/4$ channels (by averaging over every 4 channels) and train 8 models;
- average the input into $N/2$ channels (by averaging over every 2 channels) and train 8 models.

We call those 8 models "single models" and present their accuracies in the first half of the rows of Figure 7.

Next, we also present the ensemble accuracy of these 8 single models in the second half of the rows of Figure 7. (Note for the 8 single models, also use 8 different seeds for the upper-layer pre-trained models. This allows us to compare the ensemble accuracies in a more fair manner.)

### A.4 Synthetic Data: Whether Ensemble Improves Accuracy over Gaussian-Like Data

Recall in **Figure 2** we have shown that ensemble does not seem to improve test accuracy on Gaussian-like data. We explain how we perform this experiment.

**Synthetic data generation.** We generate synthetic data with $k = 10$ labels.

- We consider inputs that are generated as either Gaussian or mixture of Gaussian with different means.
- We consider inputs that are either uniformly generated, or generated through rejection sampling (so as to make different labels to have roughly the same number of data).
- We consider data that are either without label noise, or with 10% of the label randomly flipped.
- We consider data that are generated from a relatively small (but unknown to the learner) ground-truth network, that are either linear, or fully-connected (e.g. fc2 for 2 layers), or convolutional (e.g. conv3 for 3 layers), or residual (e.g. res3 for 3-layered residual and resconv3 for 3-layered residual convolutional).
- We consider data that are either generated as above, or generated with margin across labels.
- Finally, for each of the settings above, we select a dimension $d$ so that the single-model testing accuracy is around $60\% \sim 80\%$.

**Learner networks.** We also consider fully-connected, convolutional, as well as residual networks with $m = 200$ neurons to learn the given data distribution. For each data/learner pair, we use SGD with momentum 0.9, and tune the learning rate together with weight decay parameters so as to maximize test accuracy. We run for 10 single models and compare their (best) accuracy to their ensemble accuracy.

**Result: single vs ensemble.** Our detailed comparison tables are in Figure 9 (for non-convolutional inputs) and Figure 10 (for convolutional inputs). To make the result more easily interpretable, we have included in Figure 2 an abbreviated table which, for each data distribution, picks the best single and best ensemble model across all learner networks. It is clear from these reported results that, for a plethora of settings of Gaussian-like datasets, the accuracy given by ensemble barely exceeds that of single models.

**Result: accuracy consistency on single models.** For our synthetic datasets, we have also computed the mean and standard deviation for the 10 trained single models from different random initializations. We observe that their standard deviation is also negligible comparing the already not-so-great accuracy: for instance, a standard deviation of 1.0% is quite small comparing to a 70% test accuracy model on the test data. See Figure 11.

|  |  | fc2 learner | fc3 learner | fc4 learner | res3 learner | res4 learner | input dimension |
|---|---|---|---|---|---|---|---|
| **without label noise** / uniform sampling | gauss, linear | 79.4% (79.4%) 0.0% | 79.9% (79.2%) -0.7% | 78.0% (77.8%) -0.3% | 80.3% (79.2%) -1.1% | 80.1% (79.6%) -0.5% | 100 |
|  | gauss, fc2 | 65.4% (64.9%) -0.4% | 65.0% (64.2%) -0.8% | 64.2% (64.4%) 0.2% | 67.7% (65.1%) -2.6% | 64.7% (64.2%) -0.5% | 100 |
|  | gauss, fc3 | 66.6% (66.7%) 0.1% | 68.9% (68.2%) -0.6% | 68.6% (69.0%) 0.5% | 67.7% (67.5%) -0.2% | 68.4% (68.1%) -0.3% | 52 |
|  | gauss, res3 | 67.1% (67.0%) -0.1% | 68.8% (68.0%) -0.8% | 66.7% (66.3%) -0.5% | 69.1% (67.1%) -2.0% | 67.5% (67.8%) 0.3% | 100 |
|  | gauss, linear, margin | 79.0% (77.9%) -1.2% | 78.8% (77.8%) -1.0% | 77.8% (77.3%) -0.5% | 77.8% (77.3%) -0.5% | 78.7% (78.0%) -0.8% | 292 |
|  | gauss, fc2, margin | 80.6% (78.7%) -1.9% | 78.9% (79.0%) 0.2% | 78.6% (78.5%) -0.1% | 78.8% (78.7%) -0.1% | 79.0% (78.8%) -0.3% | 100 |
|  | gauss, fc3, margin | 75.2% (75.8%) 0.5% | 76.0% (75.2%) -0.8% | 75.9% (75.6%) -0.4% | 76.0% (75.4%) -0.7% | 75.8% (75.4%) -0.3% | 100 |
|  | gauss, res3, margin | 80.5% (80.8%) 0.3% | 80.4% (80.0%) -0.3% | 80.4% (79.2%) -1.2% | 80.7% (80.8%) 0.1% | 80.1% (79.5%) -0.6% | 100 |
|  | mixture, linear | 80.7% (80.0%) -0.7% | 80.5% (79.7%) -0.8% | 78.8% (78.4%) -0.4% | 80.7% (80.1%) -0.5% | 79.8% (79.1%) -0.8% | 100 |
|  | mixture, fc2 | 65.1% (64.5%) -0.5% | 65.2% (64.9%) -0.4% | 64.2% (64.5%) 0.3% | 67.7% (64.7%) -3.0% | 64.1% (63.5%) -0.5% | 100 |
|  | mixture, fc3 | 62.8% (62.8%) 0.0% | 64.0% (63.7%) -0.3% | 63.4% (64.2%) 0.8% | 63.3% (62.7%) -0.6% | 63.7% (64.4%) 0.7% | 80 |
|  | mixture, res3 | 69.9% (70.4%) 0.5% | 70.7% (71.2%) 0.5% | 70.2% (70.4%) 0.1% | 70.2% (71.2%) 1.0% | 70.6% (70.7%) 0.1% | 100 |
|  | mixture, linear, margin | 77.5% (77.2%) -0.3% | 77.8% (77.0%) -0.8% | 78.0% (76.4%) -1.6% | 77.8% (76.9%) -0.9% | 79.0% (76.6%) -2.4% | 292 |
|  | mixture, fc2, margin | 79.9% (78.0%) -1.9% | 78.8% (78.4%) -0.5% | 76.5% (76.3%) -0.2% | 80.4% (77.4%) -3.0% | 77.4% (77.6%) 0.2% | 100 |
|  | mixture, fc3, margin | 79.1% (78.6%) -0.5% | 79.9% (80.1%) 0.2% | 79.5% (79.4%) -0.1% | 79.6% (79.6%) 0.0% | 80.4% (79.4%) -0.9% | 100 |
|  | mixture, res3, margin | 82.8% (82.8%) 0.0% | 84.7% (83.9%) -0.8% | 83.7% (83.7%) 0.0% | 84.1% (83.5%) -0.6% | 83.1% (82.9%) -0.3% | 100 |
| rejection sampling | gauss, linear | 78.9% (78.6%) -0.4% | 78.7% (78.5%) -0.2% | 76.7% (76.7%) 0.0% | 78.9% (78.3%) -0.7% | 78.6% (78.4%) -0.2% | 100 |
|  | gauss, fc2 | 66.3% (64.1%) -2.2% | 66.0% (64.5%) -1.5% | 64.2% (63.8%) -0.4% | 66.0% (64.3%) -1.8% | 63.9% (63.6%) -0.3% | 100 |
|  | gauss, fc3 | 69.0% (68.6%) -0.4% | 75.5% (75.2%) -0.3% | 76.8% (76.6%) -0.2% | 72.6% (72.7%) 0.1% | 74.8% (74.7%) -0.1% | 24 |
|  | gauss, res3 | 69.3% (68.3%) -1.0% | 69.3% (69.0%) -0.3% | 67.7% (67.9%) 0.2% | 68.8% (68.2%) -0.5% | 68.3% (68.8%) 0.5% | 100 |
|  | gauss, linear, margin | 77.7% (77.3%) -0.4% | 77.6% (77.0%) -0.6% | 76.9% (76.4%) -0.5% | 78.4% (77.2%) -1.2% | 77.5% (77.3%) -0.2% | 292 |
|  | gauss, fc2, margin | 78.4% (76.6%) -1.8% | 77.0% (76.5%) -0.5% | 76.5% (76.5%) 0.0% | 76.8% (76.4%) -0.3% | 76.5% (76.2%) -0.3% | 100 |
|  | gauss, fc3, margin | 72.8% (72.2%) -0.6% | 76.6% (76.2%) -0.4% | 76.9% (77.0%) 0.2% | 75.1% (75.7%) 0.7% | 75.9% (76.7%) 0.8% | 100 |
|  | gauss, res3, margin | 83.2% (82.7%) -0.5% | 84.6% (84.0%) -0.6% | 83.2% (82.4%) -0.8% | 82.9% (83.0%) 0.0% | 83.8% (83.2%) -0.6% | 100 |
|  | mixture, linear | 80.5% (80.0%) -0.5% | 80.4% (80.0%) -0.4% | 80.2% (79.9%) -0.3% | 80.7% (80.7%) 0.0% | 79.4% (79.4%) -0.1% | 100 |
|  | mixture, fc2 | 67.6% (66.6%) -1.0% | 67.0% (66.8%) -0.2% | 66.7% (66.4%) -0.3% | 66.3% (66.0%) -0.3% | 66.3% (66.9%) 0.6% | 100 |
|  | mixture, fc3 | 65.6% (65.4%) -0.2% | 73.1% (73.1%) 0.0% | 73.2% (72.7%) -0.5% | 68.6% (68.8%) 0.2% | 70.6% (70.5%) -0.1% | 32 |
|  | mixture, res3 | 68.3% (67.8%) -0.5% | 69.4% (69.4%) 0.0% | 68.4% (68.6%) 0.2% | 69.9% (68.4%) -1.5% | 69.0% (68.5%) -0.5% | 100 |
|  | mixture, linear, margin | 79.9% (79.3%) -0.6% | 79.7% (79.5%) -0.2% | 79.0% (79.2%) 0.2% | 79.9% (80.0%) 0.0% | 80.0% (79.7%) -0.3% | 292 |
|  | mixture, fc2, margin | 75.9% (75.7%) -0.2% | 76.9% (76.9%) 0.0% | 76.4% (76.3%) -0.1% | 78.4% (77.0%) -1.5% | 76.2% (76.2%) 0.0% | 100 |
|  | mixture, fc3, margin | 71.9% (71.7%) -0.3% | 73.3% (72.8%) -0.5% | 72.8% (72.9%) 0.1% | 71.9% (72.2%) 0.3% | 72.1% (71.8%) -0.3% | 100 |
|  | mixture, res3, margin | 82.6% (81.7%) -0.9% | 84.4% (83.7%) -0.7% | 82.4% (82.4%) 0.1% | 82.4% (82.1%) -0.3% | 83.4% (82.9%) -0.5% | 100 |
| **With 10% label noise** / uniform sampling | gauss, linear | 74.3% (74.1%) -0.2% | 72.2% (71.1%) -1.1% | 68.0% (68.7%) 0.7% | 70.8% (70.4%) -0.4% | 68.9% (69.3%) 0.4% | 100 |
|  | gauss, fc2 | 63.9% (63.2%) -0.8% | 64.3% (62.0%) -2.3% | 60.3% (61.1%) 0.7% | 63.1% (62.5%) -0.6% | 61.3% (61.6%) 0.4% | 100 |
|  | gauss, fc3 | 66.1% (65.6%) -0.5% | 66.5% (66.4%) -0.1% | 64.7% (66.0%) 1.3% | 66.4% (66.3%) -0.2% | 66.0% (65.3%) -0.7% | 40 |
|  | gauss, res3 | 66.7% (65.9%) -0.8% | 66.4% (65.9%) -0.4% | 64.1% (65.2%) 1.1% | 68.7% (65.4%) -3.3% | 63.9% (63.6%) -0.3% | 80 |
|  | gauss, linear, margin | 82.1% (81.7%) -0.4% | 80.9% (80.5%) -0.4% | 78.4% (78.9%) 0.5% | 80.0% (80.3%) 0.3% | 79.8% (80.1%) 0.3% | 144 |
|  | gauss, fc2, margin | 75.9% (75.2%) -0.8% | 76.0% (75.1%) -0.9% | 73.4% (74.6%) 1.1% | 77.4% (75.3%) -2.1% | 73.7% (74.8%) 1.1% | 100 |
|  | gauss, fc3, margin | 70.7% (70.8%) 0.0% | 70.7% (70.3%) -0.4% | 68.4% (69.0%) 0.6% | 69.9% (70.1%) 0.2% | 69.3% (69.5%) 0.2% | 100 |
|  | gauss, res3, margin | 74.4% (74.0%) -0.4% | 73.1% (72.7%) -0.3% | 71.0% (71.5%) 0.5% | 75.5% (72.2%) -3.3% | 70.3% (71.1%) 0.8% | 100 |
|  | mixture, linear | 73.3% (73.1%) -0.2% | 73.6% (74.0%) 0.4% | 69.1% (70.2%) 1.1% | 72.9% (73.1%) 0.3% | 71.8% (72.6%) 0.8% | 100 |
|  | mixture, fc2 | 69.1% (66.7%) -2.4% | 68.6% (66.5%) -2.1% | 65.6% (65.8%) 0.2% | 70.1% (66.6%) -3.5% | 64.1% (64.6%) 0.5% | 80 |
|  | mixture, fc3 | 63.0% (61.9%) -1.2% | 61.3% (62.4%) 1.1% | 60.2% (60.0%) -0.2% | 61.5% (60.9%) -0.6% | 58.8% (58.7%) 0.0% | 80 |
|  | mixture, res3 | 66.8% (63.6%) -3.2% | 66.9% (63.5%) -3.4% | 61.0% (62.2%) 1.2% | 66.0% (63.8%) -2.3% | 61.6% (62.9%) 1.4% | 100 |
|  | mixture, linear, margin | 80.7% (79.7%) -1.0% | 79.4% (80.0%) 0.7% | 78.1% (77.0%) -1.1% | 78.8% (79.0%) 0.2% | 78.5% (78.6%) 0.1% | 144 |
|  | mixture, fc2, margin | 73.7% (73.9%) 0.1% | 73.2% (73.2%) 0.0% | 72.2% (72.9%) 0.7% | 72.6% (73.1%) 0.5% | 72.7% (73.1%) 0.3% | 100 |
|  | mixture, fc3, margin | 73.5% (73.5%) 0.0% | 73.9% (74.6%) 0.8% | 71.9% (73.1%) 1.2% | 71.9% (72.7%) 0.9% | 71.8% (72.1%) 0.3% | 100 |
|  | mixture, res3, margin | 75.9% (76.2%) 0.4% | 76.7% (76.4%) -0.3% | 75.9% (76.4%) 0.5% | 74.5% (76.4%) -0.2% | 74.9% (74.9%) 0.3% | 100 |
| rejection sampling | gauss, linear | 72.9% (72.2%) -0.8% | 72.7% (71.5%) -1.2% | 66.6% (67.1%) 0.5% | 72.2% (70.5%) -1.8% | 72.0% (67.7%) -4.3% | 100 |
|  | gauss, fc2 | 62.7% (62.2%) -0.5% | 62.8% (62.2%) -0.6% | 60.6% (60.4%) -0.2% | 63.7% (62.2%) -1.5% | 64.6% (63.5%) -1.1% | 80 |
|  | gauss, fc3 | 65.2% (64.5%) -0.7% | 69.9% (69.1%) -0.8% | 72.5% (72.0%) -0.5% | 67.2% (67.3%) 0.0% | 69.6% (69.5%) -0.1% | 24 |
|  | gauss, res3 | 67.6% (66.5%) -1.1% | 68.1% (68.1%) 0.0% | 66.9% (67.8%) 1.0% | 64.3% (64.9%) 0.6% | 64.5% (64.5%) 0.0% | 80 |
|  | gauss, linear, margin | 81.6% (80.2%) -1.4% | 81.3% (78.8%) -2.5% | 76.7% (76.7%) 0.0% | 78.1% (78.0%) -0.1% | 77.4% (77.4%) 0.1% | 144 |
|  | gauss, fc2, margin | 72.1% (72.2%) 0.1% | 72.1% (71.5%) -0.6% | 71.5% (71.6%) 0.1% | 74.2% (71.6%) -2.6% | 74.7% (71.3%) -3.5% | 100 |
|  | gauss, fc3, margin | 66.6% (66.9%) 0.3% | 70.4% (70.5%) 0.1% | 64.4% (64.5%) 0.1% | 68.0% (66.1%) 0.6% | 63.8% (64.5%) 0.7% | 100 |
|  | gauss, res3, margin | 76.8% (73.6%) -3.2% | 74.1% (74.3%) 0.3% | 74.3% (74.8%) 0.5% | 70.6% (71.1%) 0.5% | 70.1% (71.1%) 1.0% | 100 |
|  | mixture, linear | 74.2% (73.7%) -0.5% | 73.8% (73.5%) -0.3% | 72.5% (72.6%) 0.1% | 73.5% (72.4%) -1.1% | 69.6% (70.7%) 1.1% | 100 |
|  | mixture, fc2 | 62.3% (62.3%) 0.0% | 66.0% (63.3%) -2.8% | 66.2% (62.0%) -4.2% | 62.3% (63.1%) 0.8% | 61.6% (61.5%) -0.1% | 80 |
|  | mixture, fc3 | 71.9% (71.3%) -0.5% | 75.3% (74.8%) -0.5% | 78.1% (78.6%) 0.5% | 73.4% (73.5%) 0.1% | 74.4% (73.9%) -0.5% | 20 |
|  | mixture, res3 | 66.0% (65.9%) -0.1% | 68.5% (69.5%) 1.1% | 68.8% (69.4%) 0.5% | 64.6% (65.9%) 1.3% | 63.5% (64.9%) 1.4% | 80 |
|  | mixture, linear, margin | 84.1% (81.5%) -2.6% | 82.0% (82.3%) 0.3% | 82.7% (82.4%) -0.3% | 80.8% (80.7%) -0.1% | 80.5% (80.6%) 0.1% | 144 |
|  | mixture, fc2, margin | 75.7% (71.4%) -4.3% | 71.6% (70.9%) -0.7% | 71.0% (71.4%) 0.4% | 69.9% (69.7%) -0.2% | 69.0% (70.1%) 1.1% | 100 |
|  | mixture, fc3, margin | 65.2% (65.5%) 0.3% | 66.8% (66.4%) -0.5% | 66.1% (66.3%) 0.2% | 67.5% (64.0%) -3.5% | 62.6% (62.8%) 0.2% | 100 |
|  | mixture, res3, margin | 75.6% (72.4%) -3.2% | 74.1% (74.5%) 0.4% | 72.4% (73.2%) 0.8% | 76.4% (72.6%) -3.8% | 75.0% (71.9%) -3.1% | 100 |

Figure 9: For synthetic Gaussian-like data, ensemble barely helps on improving test accuracy. In this table, we give a closer look at how ensemble performs with respect to each individual learner network. (For a more concise view, see Figure 2; for convolutional data, see Figure 10.)

Figure 10: For synthetic Gaussian-like data, ensemble barely helps on improving test accuracy. In this table, we give a closer look at how ensemble performs with respect to each individual learner network, when the data is generated from a *target convolutional network*. This gives evidence that, having convolutional data may alone be necessarily for ensemble to work either.

|  |  | learner networks | | | | | | | |
| --- | --- | --- | --- | --- | --- | --- | --- | --- | --- |
|  |  | synthetic data | fc2 | fc3 | fc4 | res3 | res4 | conv2 | conv3 | resconv3 |
| without label noise | uniform sampling | gauss, linear | 79.0±0.2% | 79.4±0.3% | 77.5±0.4% | 79.5±0.3% | 79.4±0.3% | 74.4±0.6% | 73.7±0.7% | 74.3±0.5% |
| | | gauss, fc2 | 65.0±0.2% | 64.5±0.3% | 63.8±0.3% | 64.7±1.2% | 64.3±0.3% | 62.1±0.4% | 62.1±0.5% | 62.5±0.5% |
| | | gauss, fc3 | 66.4±0.1% | 68.4±0.3% | 68.0±0.3% | 67.3±0.3% | 67.3±0.5% | 59.4±0.3% | 62.6±0.8% | 61.1±0.6% |
| | | gauss, res3 | 66.4±0.3% | 67.3±0.6% | 66.1±0.6% | 67.3±0.7% | 66.8±0.5% | 60.9±0.4% | 60.3±0.7% | 60.9±0.5% |
| | | gauss, linear, margin | 78.2±0.3% | 78.0±0.3% | 77.2±0.4% | 77.0±0.3% | 77.8±0.4% | 70.3±1.0% | 69.6±0.8% | 69.7±0.9% |
| | | gauss, fc2, margin | 78.5±0.8% | 78.4±0.3% | 78.2±0.4% | 78.3±0.4% | 78.3±0.3% | 75.0±0.3% | 74.6±0.4% | 74.0±0.3% |
| | | gauss, fc3, margin | 75.0±0.1% | 75.2±0.4% | 75.2±0.4% | 75.2±0.4% | 75.3±0.4% | 67.9±0.4% | 68.2±0.7% | 66.6±0.8% |
| | | gauss, res3, margin | 80.1±0.3% | 80.0±0.3% | 77.7±3.0% | 80.2±0.3% | 79.6±0.3% | 71.9±0.4% | 71.3±0.6% | 71.8±0.5% |
| | | mixture, linear | 80.0±0.3% | 79.8±0.3% | 78.1±0.5% | 80.2±0.3% | 79.3±0.3% | 75.3±0.5% | 72.8±0.5% | 74.8±0.5% |
| | | mixture, fc2 | 64.2±0.4% | 64.2±0.5% | 63.0±0.5% | 64.4±1.2% | 63.5±0.4% | 60.9±0.3% | 60.6±0.5% | 59.9±0.4% |
| | | mixture, fc3 | 62.3±0.3% | 63.4±0.4% | 63.1±0.3% | 62.5±0.5% | 63.0±0.4% | 58.3±0.3% | 60.2±0.6% | 60.2±0.6% |
| | | mixture, res3 | 69.5±0.4% | 70.3±0.4% | 69.9±0.3% | 69.7±0.3% | 69.8±0.5% | 63.2±0.4% | 61.5±0.6% | 63.1±0.5% |
| | | mixture, linear, margin | 76.8±0.3% | 76.8±0.5% | 76.4±0.7% | 77.0±0.5% | 76.9±0.8% | 69.7±0.7% | 69.8±0.6% | 69.7±1.0% |
| | | mixture, fc2, margin | 77.2±1.1% | 77.3±0.6% | 75.9±0.4% | 77.1±1.2% | 76.4±0.4% | 74.0±0.4% | 73.2±0.4% | 73.9±0.5% |
| | | mixture, fc3, margin | 78.4±0.3% | 79.6±0.2% | 79.1±0.4% | 79.1±0.3% | 79.1±0.6% | 71.8±0.6% | 72.0±0.4% | 72.1±0.5% |
| | | mixture, res3, margin | 82.5±0.2% | 83.7±0.5% | 82.5±1.2% | 83.2±0.5% | 82.7±0.3% | 74.7±0.4% | 73.9±0.6% | 73.4±0.5% |
| | rejection sampling | gauss, linear | 78.2±0.5% | 78.0±0.4% | 76.2±0.3% | 78.4±0.3% | 78.0±0.3% | 73.9±0.5% | 72.3±0.9% | 73.4±0.7% |
| | | gauss, fc2 | 63.7±1.0% | 63.9±0.8% | 63.1±0.7% | 63.8±0.9% | 63.0±0.4% | 59.6±0.2% | 59.3±0.6% | 59.4±0.7% |
| | | gauss, fc3 | 68.4±0.3% | 75.1±0.2% | 76.1±0.3% | 72.2±0.3% | 73.6±0.6% | 56.8±0.3% | 65.1±0.4% | 63.6±0.5% |
| | | gauss, res3 | 67.9±0.5% | 68.9±0.3% | 67.0±0.5% | 67.9±0.3% | 67.5±0.4% | 58.3±0.3% | 59.7±0.6% | 59.6±0.7% |
| | | gauss, linear, margin | 77.2±0.3% | 76.9±0.3% | 76.3±0.4% | 77.3±0.4% | 77.0±0.3% | 70.4±0.9% | 69.5±0.5% | 68.5±0.8% |
| | | gauss, fc2, margin | 76.4±0.8% | 76.4±0.4% | 76.2±0.3% | 76.4±0.3% | 76.1±0.3% | 72.2±0.4% | 70.3±0.7% | 72.0±0.5% |
| | | gauss, fc3, margin | 72.5±0.2% | 76.2±0.3% | 75.8±0.5% | 74.6±0.3% | 74.6±0.4% | 58.9±0.8% | 63.7±1.0% | 62.9±0.5% |
| | | gauss, res3, margin | 82.5±0.3% | 83.7±0.4% | 82.4±0.4% | 82.7±0.1% | 82.6±0.5% | 72.5±0.5% | 72.2±0.6% | 73.7±0.7% |
| | | mixture, linear | 80.2±0.3% | 80.0±0.2% | 78.4±1.6% | 80.1±0.3% | 78.8±0.3% | 74.5±0.6% | 73.9±1.0% | 74.7±0.5% |
| | | mixture, fc2 | 66.2±0.5% | 66.3±0.4% | 66.0±0.4% | 66.0±0.3% | 65.6±0.5% | 62.6±0.5% | 62.0±0.6% | 62.6±0.6% |
| | | mixture, fc3 | 65.1±0.3% | 72.2±0.7% | 72.8±0.3% | 67.7±0.4% | 69.8±0.6% | 52.1±0.3% | 61.8±0.4% | 60.0±0.4% |
| | | mixture, res3 | 67.7±0.3% | 68.9±0.2% | 67.7±0.4% | 67.6±0.9% | 67.9±0.5% | 59.8±0.4% | 59.5±0.6% | 59.8±0.5% |
| | | mixture, linear, margin | 79.3±0.3% | 79.2±0.3% | 78.4±0.3% | 79.5±0.2% | 79.5±0.3% | 71.2±0.7% | 71.0±0.7% | 70.7±1.2% |
| | | mixture, fc2, margin | 75.6±0.2% | 76.2±0.4% | 76.0±0.3% | 76.0±0.9% | 75.7±0.3% | 72.8±0.4% | 71.9±0.5% | 72.4±0.7% |
| | | mixture, fc3, margin | 71.4±0.3% | 72.8±0.5% | 71.5±0.6% | 71.5±0.4% | 71.4±0.4% | 62.0±0.3% | 63.4±0.8% | 61.9±0.4% |
| | | mixture, res3, margin | 81.4±0.5% | 83.1±0.5% | 81.7±0.3% | 81.6±0.4% | 82.3±0.5% | 71.3±0.6% | 71.0±0.7% | 71.3±0.4% |
| With 10% label noise | uniform sampling | gauss, linear | 73.2±0.5% | 71.1±0.5% | 67.3±0.5% | 69.8±0.4% | 68.2±0.5% | 67.3±0.4% | 66.0±0.7% | 67.5±0.4% |
| | | gauss, fc2 | 62.8±0.5% | 61.7±1.0% | 59.6±0.5% | 61.7±0.6% | 60.0±0.7% | 61.3±0.6% | 60.7±0.3% | 61.1±0.4% |
| | | gauss, fc3 | 65.7±0.2% | 66.2±0.2% | 63.9±0.6% | 65.6±0.5% | 65.0±0.6% | 60.5±0.4% | 61.4±0.5% | 60.9±0.3% |
| | | gauss, res3 | 66.4±0.2% | 65.6±0.5% | 63.4±0.5% | 64.8±1.4% | 62.6±0.6% | 62.3±0.9% | 61.7±0.8% | 62.5±0.6% |
| | | gauss, linear, margin | 81.5±0.3% | 80.1±0.4% | 77.9±0.3% | 79.4±0.4% | 78.8±0.5% | 75.0±0.4% | 73.8±0.6% | 74.9±0.6% |
| | | gauss, fc2, margin | 75.3±0.3% | 74.6±0.5% | 73.1±0.3% | 74.5±1.1% | 73.2±0.4% | 73.6±0.2% | 72.9±0.6% | 73.1±0.6% |
| | | gauss, fc3, margin | 70.4±0.2% | 70.3±0.3% | 67.9±0.5% | 69.2±0.3% | 68.1±0.7% | 66.4±0.3% | 66.0±0.5% | 66.2±0.5% |
| | | gauss, res3, margin | 72.8±0.7% | 72.3±0.4% | 70.1±0.4% | 71.0±1.6% | 69.4±0.6% | 68.4±0.4% | 67.3±0.7% | 68.6±0.5% |
| | | mixture, linear | 72.3±0.5% | 72.6±0.5% | 68.2±0.6% | 71.2±1.0% | 70.3±0.9% | 69.4±0.5% | 67.4±0.9% | 68.0±0.6% |
| | | mixture, fc2 | 66.2±1.1% | 65.2±1.2% | 62.2±1.4% | 64.4±2.0% | 63.5±0.4% | 64.2±0.3% | 63.9±0.4% | 63.7±0.6% |
| | | mixture, fc3 | 61.1±0.8% | 59.9±0.7% | 59.0±0.5% | 59.3±0.9% | 57.9±0.6% | 57.9±0.5% | 58.1±0.7% | 57.5±0.3% |
| | | mixture, res3 | 63.2±1.3% | 62.2±1.9% | 60.3±0.4% | 61.3±1.8% | 60.3±0.7% | 59.2±0.3% | 60.0±0.8% | 59.8±0.9% |
| | | mixture, linear, margin | 79.2±0.5% | 78.8±0.4% | 76.5±1.6% | 77.9±0.6% | 77.5±0.6% | 72.4±0.6% | 71.8±0.9% | 72.0±0.8% |
| | | mixture, fc2, margin | 72.8±0.5% | 72.5±0.3% | 71.2±0.6% | 71.9±0.5% | 71.8±0.4% | 70.6±0.5% | 69.8±0.5% | 70.2±0.4% |
| | | mixture, fc3, margin | 72.4±0.5% | 72.9±0.6% | 70.8±0.7% | 71.4±0.4% | 70.6±0.5% | 69.5±0.5% | 69.2±0.6% | 69.1±0.6% |
| | | mixture, res3, margin | 75.5±0.3% | 75.8±0.6% | 73.9±1.7% | 74.2±0.6% | 73.6±0.6% | 72.0±0.4% | 71.7±0.7% | 71.3±0.6% |
| | rejection sampling | gauss, linear | 71.8±0.5% | 70.2±1.0% | 66.0±0.4% | 68.6±1.4% | 65.8±2.3% | 65.1±0.7% | 64.0±0.9% | 65.3±0.7% |
| | | gauss, fc2 | 62.2±0.3% | 62.0±0.6% | 59.2±0.6% | 60.7±1.1% | 60.1±1.7% | 59.9±0.5% | 59.0±0.4% | 59.8±0.4% |
| | | gauss, fc3 | 64.6±0.4% | 69.3±0.4% | 71.3±0.7% | 66.9±0.2% | 68.3±0.9% | 54.7±0.5% | 62.7±1.1% | 60.0±0.7% |
| | | gauss, res3 | 66.5±0.5% | 67.5±0.3% | 66.1±0.5% | 63.8±0.4% | 63.3±0.7% | 59.2±0.4% | 58.6±0.9% | 59.7±0.6% |
| | | gauss, linear, margin | 80.3±0.6% | 78.3±1.2% | 76.2±0.4% | 77.5±0.4% | 76.9±0.2% | 73.4±0.8% | 72.2±0.6% | 73.4±1.0% |
| | | gauss, fc2, margin | 71.8±0.2% | 71.6±0.3% | 70.1±1.2% | 71.0±1.2% | 69.9±1.7% | 69.4±0.6% | 68.9±0.5% | 69.5±0.5% |
| | | gauss, fc3, margin | 66.2±0.2% | 69.5±0.5% | 63.5±0.5% | 64.6±0.4% | 63.1±0.5% | 58.3±0.5% | 60.5±1.0% | 59.3±0.5% |
| | | gauss, res3, margin | 73.4±1.2% | 73.6±0.4% | 73.5±0.5% | 70.1±0.3% | 69.1±0.6% | 67.6±0.3% | 68.1±0.6% | 68.4±0.8% |
| | | mixture, linear | 72.3±0.8% | 70.8±1.2% | 69.2±3.1% | 70.0±1.3% | 68.9±0.6% | 67.9±0.5% | 66.3±0.5% | 67.4±0.7% |
| | | mixture, fc2 | 61.9±0.3% | 62.2±1.4% | 59.8±2.3% | 61.6±0.7% | 59.6±0.8% | 59.3±0.7% | 59.8±0.4% | 59.9±0.7% |
| | | mixture, fc3 | 71.4±0.2% | 74.9±0.3% | 77.2±0.6% | 72.6±0.4% | 73.6±0.5% | 61.7±0.2% | 69.0±0.2% | 67.3±0.5% |
| | | mixture, res3 | 64.9±0.6% | 67.6±0.7% | 67.0±1.0% | 63.9±0.4% | 62.2±0.7% | 60.4±0.6% | 60.6±0.8% | 59.8±0.7% |
| | | mixture, linear, margin | 81.1±1.1% | 81.5±0.4% | 80.7±2.0% | 80.3±0.3% | 79.9±0.4% | 75.0±1.2% | 73.4±1.2% | 74.1±1.1% |
| | | mixture, fc2, margin | 70.8±1.8% | 70.5±0.6% | 69.9±0.6% | 69.0±0.6% | 68.2±0.6% | 68.2±0.6% | 68.2±0.6% | 67.5±0.7% |
| | | mixture, fc3, margin | 64.7±0.3% | 65.4±0.7% | 65.0±0.6% | 63.3±1.5% | 61.5±0.6% | 60.7±0.7% | 61.4±0.6% | 60.0±0.7% |
| | | mixture, res3, margin | 71.6±1.4% | 73.3±0.6% | 71.2±0.7% | 70.6±2.1% | 69.3±2.2% | 68.1±0.4% | 67.8±0.7% | 67.6±0.6% |

Figure 11: For synthetic Gaussian-like data, we compute the standard deviations of single models test accuracies over 10 runs. This standard deviation is generally bigger than what we see on the CIFAR10/100 datasets (see Figure 1), yet ensemble still offers nearly no benefit.

# Appendix II: Complete Proofs

## B Single Model: Proof Plan and Induction Hypothesis

Our main proof relies on an induction hypothesis for every iteration $t = 0, 1, 2, \ldots, T$. Before we state it, let us introduce several notations. Let us denote

$$\Lambda_i^{(t)} \stackrel{\text{def}}{=} \max_{r \in [m], \ell \in [2]} [\langle w_{i,r}^{(t)}, v_{i,\ell} \rangle]^+ \quad \text{and} \quad \Lambda_{i,\ell}^{(t)} \stackrel{\text{def}}{=} \max_{r \in [m]} [\langle w_{i,r}^{(t)}, v_{i,\ell} \rangle]^+ \tag{B.1}$$

Suppose $m \leq \mathsf{poly}(k)$. For every $i \in [k]$, let us denote

$$\mathcal{M}_i^{(0)} \stackrel{\text{def}}{=} \left\{ r \in [m] \,\Big|\, \exists \ell \in [2] \colon \langle w_{i,r}^{(0)}, v_{i,\ell} \rangle \geq \Lambda_{i,\ell}^{(0)} \left( 1 - O\big(\tfrac{1}{\log k}\big) \right) \right\}$$

**Intuition.** If a neuron $r \in [m]$ is not in $\mathcal{M}_i^{(0)}$, it means that for both $\ell \in \{1, 2\}$, the correlation $\langle w_{i,r}^{(0)}, v_{i,\ell} \rangle$ at the random initialization is, by a non-trivial factor, smaller than $\Lambda_{i,\ell}^{(0)}$— the largest correlation between $\langle w_{i,r'}, v_{i,\ell} \rangle$ among all neurons. In words, this means the magnitude of $v_{i,1}$ and $v_{i,2}$ inside the random initialization $w_{i,r}^{(0)}$ is non-trivially lagging behind, comparing to other neurons. We shall prove that, through the course of the training, those neurons $r$ will lose the lottery and not learn anything useful for the output label $i \in [k]$. (This corresponds to Induction Hypothesis B.3i later.)

**Fact B.1.** *With probability at least $1 - e^{-\Omega(\log^5 k)}$, we have $|\mathcal{M}_i^{(0)}| \leq m_0 \stackrel{\text{def}}{=} O(\log^5 k)$.*

(The proof of Fact B.1 follows from standard analysis on Gaussian variables, see Proposition G.1.)

Suppose we denote by $S_{i,\ell} \stackrel{\text{def}}{=} \mathbb{E}_{(X,y) \sim \mathcal{Z}_m} \left[ \mathbb{1}_{y=i} \sum_{p \in P_{v_{i,\ell}}(X)} z_p^q \right]$. Then, define

$$\mathcal{M} \stackrel{\text{def}}{=} \left\{ (i, \ell^*) \in [k] \times [2] \,\Big|\, \Lambda_{i,\ell^*}^{(0)} \geq \Lambda_{i, 3-\ell^*}^{(0)} \left( \tfrac{S_{i, 3-\ell^*}}{S_{i,\ell^*}} \right)^{\frac{1}{q-2}} \left( 1 + \tfrac{1}{\log^2(m)} \right) \right\} \tag{B.2}$$

**Intuition.** If $(i, \ell) \in \mathcal{M}$, it means the feature $v_{i,\ell}$ has much higher chance than $v_{i,3-\ell}$ to be learned by the model. (This is because, after an appropriate scaling factor defined by the training data, $v_{i,\ell}$ correlates more with the network's random initialization comparing to $v_{i,3-\ell}$. As we mentioned earlier, we shall prove that for each $i \in [k]$, at least one of the two features $v_{i,1}$ or $v_{i,2}$. But more importantly, our later Induction Hypothesis B.3e ensures that, during the entire training process, if $(i, 3 - \ell) \in \mathcal{M}$, then $v_{i,\ell}$ must be missing from the learner network.

Our next proposition states that, for every $i \in [k]$, with descent probability at least one of $(i, 1)$ or $(i, 2)$ shall be in $\mathcal{M}$— this implies the test accuracy on single-view data cannot exceed 49.99%, as one of the views shall be missing. However, our next proposition also ensures that the "weaker" feature among the two, still has some non-negligible chance to be picked up by the random initialization. This is behind the reason that why ensemble works in our later proofs.

**Proposition B.2.** *Suppose $m \leq \mathsf{poly}(k)$. We have the following properties about $\mathcal{M}$.*

- *For every $i \in [k]$, at most one of $(i, 1)$ or $(i, 2)$ is in $\mathcal{M}$ (obvious).*
- *For every $i \in [k]$, suppose $S_{i,\ell} \geq S_{i,3-\ell}$, then*
    - $\mathbf{Pr}\left[(i, 3 - \ell) \in \mathcal{M}\right] \geq m^{-O(1)}$.
    - $\mathbf{Pr}\left[(i, \ell) \in \mathcal{M} \text{ or } (i, 3 - \ell) \in \mathcal{M}\right] \geq 1 - o(1)$

(Proposition B.2 is a result of the anti-concentration of the maximum of Gaussian, see Appendix G.)

We are now ready to state our induction hypothesis.

**Induction Hypothesis B.3.** *For every $\ell \in [2]$, for every $r \in [m]$, for every $(X, y) \in \mathcal{Z}_m$ and $i \in [k]$, or for every $(X, y) \in \mathcal{Z}_s$ and $i \in [k] \setminus \{y\}$:*

(a) *For every $p \in \mathcal{P}_{v_{i,\ell}}(X)$, we have: $\langle w_{i,r}^{(t)}, x_p \rangle = \langle w_{i,r}^{(t)}, v_{i,\ell} \rangle z_p \pm \widetilde{o}(\sigma_0)$.*

(b) *For every $p \in \mathcal{P}(X) \setminus (\mathcal{P}_{v_{i,1}}(X) \cup \mathcal{P}_{v_{i,2}}(X))$, we have: $|\langle w_{i,r}^{(t)}, x_p \rangle| \leq \widetilde{O}(\sigma_0)$.*

(c) *For every $p \in [P] \setminus \mathcal{P}(X)$, we have: $|\langle w_{i,r}^{(t)}, x_p \rangle| \leq \widetilde{O}(\sigma_0 \gamma k)$.*

*In addition, for every $(X, y) \in \mathcal{Z}_s$, every $i \in [k]$, every $r \in [m]$, every $\ell \in [2]$,*

(d) *For every $p \in \mathcal{P}_{v_{i,\ell}}(X)$, we have: $\langle w_{i,r}^{(t)}, x_p \rangle = \langle w_{i,r}^{(t)}, v_{i,\ell} \rangle z_p + \langle w_{i,r}^{(t)}, \xi_p \rangle \pm \widetilde{O}(\sigma_0 \gamma k)$*

(e) *For every $p \in \mathcal{P}_{v_{i,\ell}}(X)$, if $(i, 3-\ell) \in \mathcal{M}$ we have: $|\langle w_{i,r}^{(t)}, x_p \rangle| \leq \widetilde{O}(\sigma_0)$.*

(f) *For every $p \in \mathcal{P}_{v_{i,\ell}}(X)$, if $r \in [m] \setminus \mathcal{M}_i^{(0)}$ we have: $|\langle w_{i,r}^{(t)}, x_p \rangle| \leq \widetilde{O}(\sigma_0)$.*

*Moreover, we have for every $i \in [k]$,*

(g) $\Lambda_i^{(t)} \geq \Omega(\sigma_0)$ *and* $\Lambda_i^{(t)} \leq \widetilde{O}(1)$.

(h) *for every $r \in [m]$, every $\ell \in [2]$, it holds that $\langle w_{i,r}^{(t)}, v_{i,\ell} \rangle \geq -\widetilde{O}(\sigma_0)$.*

(i) *for every $r \in [m] \setminus \mathcal{M}_i^{(0)}$, every $\ell \in [2]$, it holds that $\langle w_{i,r}^{(t)}, v_{i,\ell} \rangle \leq \widetilde{O}(\sigma_0)$.*

**Intuition.** The first three items in Induction Hypothesis B.3 essentially say that, when studying the correlation between $w_{i,r}$ with a multi-view data, or between $w_{i,r}$ with a single-view data (but $y \neq i$), the correlation is about $\langle w_{i,r}, v_{i,1} \rangle$ and $\langle w_{i,r}, v_{i,2} \rangle$ and the remaining terms are sufficiently small. (Of course, this requires a careful proof.) We shall later prove that at least one of $\Lambda_{i,1}^{(t)}$ or $\Lambda_{i,2}^{(t)}$ is large after training. Therefore, using the first three items, we can argue that all multi-view data are classified correctly.

The middle three items in Induction Hypothesis B.3 essentially say that, when studying the correlation between $w_{i,r}$ with a single-view data $(X, y)$ with $y = i$, then the correlation also has a significant noise term $\langle w_{i,r}^{(t)}, \xi_p \rangle$. This term shall become useful for us to argue that single-view data can be all memorized (through for instance memorizing the noise).

The remaining items in Induction Hypothesis B.3 are just some regularization statements.

## C  Single Model: Technical Proofs

We devote this section to prove that Induction Hypothesis B.3 holds for every iteration $t \leq T$, and in the next Section D, we state how the induction hypothesis easily implies our main theorems for single model and ensemble model.

**Parameter C.1.** We state the range of parameters for our proofs in this section to hold.

- $\varrho = \frac{1}{\mathsf{polylog}(k)}$ \hfill (recall $\varrho$ is the threshold for the smoothed ReLU activation)

- $\Gamma = \frac{1}{\mathsf{polylog}(k)}$ \hfill (recall $\Gamma$ controls off-target feature magnitude in Def. 3.1)

- $q \geq 3$ and $\sigma_0^{q-2} = \frac{1}{k}$ \hfill (recall $w_{i,r}^{(0)} \sim \mathcal{N}(0, \sigma_0^2 I)$ gives the initialization magnitude)

- $N_s \leq \widetilde{o}(k/\rho)$ and $N_s \leq \frac{k^2}{s}\rho^{q-1}$. \hfill (recall $N_s$ is the size of single-view training data)

- $\gamma \leq \widetilde{O}(\frac{\sigma_0}{k})$ and $\gamma^q \leq \widetilde{\Theta}\left(\frac{1}{k^{q-1}mP}\right)$ \hfill (recall $\gamma$ controls feature noise in Def. 3.1)

- polylog$(k) \leq s \leq k^{0.2}$                                                                         (recall $s$ controls feature sparsity in Def. 3.1)
- $\rho^{q-1} \geq \frac{1}{k}$                              (recall $\rho$ controls on-target feature magnitude of single-view data in Def. 3.1)
- $N \geq N_s \cdot \mathsf{poly}(k)$, $\eta T \geq N \cdot \mathsf{poly}(k)$, and $\sqrt{d} \geq \eta T \cdot \mathsf{poly}(k)$.
- $\mathsf{polylog}(k) \leq m \leq \widetilde{O}(\frac{1}{s\sigma_0^q})$.

**Example.** A reasonable set of parameters is, up to polylogarithmic factors:

$$q = 4, \quad \sigma_0 = \frac{1}{\sqrt{k}}, \quad \rho = \frac{1}{k^{0.2}}, \quad m \leq k, \quad s \leq k^{0.2}, \quad N_s \leq k^{1.2}, \quad P \leq k^2, \quad \gamma \leq \frac{1}{k^{1.5}}.$$

**Theorem C.2.** *Under Parameter C.1, for any $m \in \left[\widetilde{\Omega}(1), \widetilde{O}(\frac{1}{s\sigma_0^q})\right]$ and sufficiently small $\eta \leq \frac{1}{\mathsf{poly}(k)}$, our Induction Hypothesis B.3 holds for all iterations $t = 0, 1, \ldots, T$.*

## C.1 Gradient Calculations and Function Approximation

**Gradient calculations.** Recall $\mathbf{logit}_i(F, X) \overset{\text{def}}{=} \frac{e^{F_i(X)}}{\sum_{j \in [k]} e^{F_j(X)}}$. Recall also

**Fact C.3.** *Given data point $(X, y) \in \mathcal{D}$, for every $i \in [k]$, $r \in [m]$,*

$$-\nabla_{w_{i,r}} L(F; X, y) = (1 - \mathbf{logit}_i(F, X)) \sum_{p \in [P]} \widetilde{\mathsf{ReLU}}'(\langle w_{i,r}, x_p \rangle) x_p \qquad \text{when } i = y \qquad \text{(C.1)}$$

$$-\nabla_{w_{i,r}} L(F; X, y) = -\mathbf{logit}_i(F, X) \sum_{p \in [P]} \widetilde{\mathsf{ReLU}}'(\langle w_{i,r}, x_p \rangle) x_p \qquad \text{when } i \neq y \qquad \text{(C.2)}$$

Now, we also have the following observations:

**Claim C.4.** *If Induction Hypothesis B.3 holds at iteration $t$, and if $s \leq \widetilde{O}(\frac{1}{\sigma_0^q m})$ and $\gamma \leq \widetilde{O}(\frac{1}{\sigma_0 k(mP)^{1/q}})$, then*

- *for every $(X, y) \in \mathcal{Z}$ and every $i \in [k]$: $\mathbf{logit}_i(F^{(t)}, X) = O\left(\frac{e^{O(\Lambda_i^{(t)})m_0}}{e^{O(\Lambda_i^{(t)})m_0} + k}\right)$*
- *for every $(X, y) \in \mathcal{Z}_s$ and $i \in [k] \setminus \{y\}$: $\mathbf{logit}_i(F^{(t)}, X) = O\left(\frac{1}{k}\right)\left(1 - \mathbf{logit}_y(F^{(t)}, X)\right)$*

*Proof of Claim C.4.* Recall $F_i^{(t)}(X) = \sum_{r \in [m]} \sum_{p \in [P]} \widetilde{\mathsf{ReLU}}(\langle w_{i,r}^{(t)}, x_p \rangle)$. For every $r \notin \mathcal{M}_i^{(0)}$, using Induction Hypothesis B.3i we have

$$\sum_{p \in [P]} \widetilde{\mathsf{ReLU}}(\langle w_{i,r}^{(t)}, x_p \rangle) \leq \widetilde{O}(\sigma_0^q) \cdot s + \widetilde{O}((\sigma_0 \gamma k)^q) \cdot P \leq \frac{1}{m \mathsf{polylog}(k)}$$

so they sum up to at most $\frac{1}{\mathsf{polylog}(k)}$. For any $r \in \mathcal{M}_i^{(0)}$, we have

$$\sum_{p \in [P]} \widetilde{\mathsf{ReLU}}(\langle w_{i,r}^{(t)}, x_p \rangle) \leq \sum_{\ell \in [2]} \left[\langle w_{i,r}^{(t)}, v_{i,\ell} \rangle\right]^+ \cdot \left(\sum_{p \in \mathcal{P}_{v_i,\ell}(X)} z_p\right) + \widetilde{o}(\sigma_0) + \widetilde{O}(\sigma_0^q) \cdot s + \widetilde{O}((\sigma_0 \gamma k)^q) \cdot P$$

$$\leq \sum_{\ell \in [2]} \left[\langle w_{i,r}^{(t)}, v_{i,\ell} \rangle\right]^+ \cdot \left(\sum_{p \in \mathcal{P}_{v_i,\ell}(X)} z_p\right) + O\left(\frac{1}{m_0}\right)$$

Recall from Def. 3.1 we have $\sum_{p \in \mathcal{P}_{v_i,\ell}(X)} z_p \leq O(1)$; and furthermore when $(X, y) \in \mathcal{Z}_s$ and $i \in [k] \setminus \{y\}$ we have $\sum_{p \in \mathcal{P}_{v_i,\ell}(X)} z_p \leq \Gamma = \frac{1}{\mathsf{polylog}(k)}$. In the former case, we have

$$0 \leq F_i^{(t)}(X) \leq m_0 \cdot \Lambda_i^{(t)} \cdot O(1) + O(1)$$

and this proves the first bound; in the latter case we have

$$0 \leq F_i^{(t)}(X) \leq m_0 \cdot \Lambda_i^{(t)} \cdot \Gamma + O(1) \leq O(1) \tag{C.3}$$

and this proves the second bound. □

**Definition C.5.** *For each data point $X$, we consider a value $V_{i,r,\ell}(X)$ given as:*

$$V_{i,r,\ell}(X) \stackrel{\text{def}}{=} \mathbb{1}_{v_{i,\ell} \in \mathcal{V}(X)} \sum_{p \in \mathcal{P}_{v_{i,\ell}}(X)} \widetilde{\mathsf{ReLU}}'(\langle w_{i,r}, x_p \rangle) z_p$$

**Definition C.6.** *We define four error terms that shall be used frequently in our proofs.*

$$\mathcal{E}_1 := \widetilde{O}(\sigma_0^{q-1}) \gamma s \qquad\qquad \mathcal{E}_{2,i,r}(X) := O(\gamma(V_{i,r,1}(X) + V_{i,r,2}(X)))$$
$$\mathcal{E}_3 := \widetilde{O}(\sigma_0 \gamma k)^{q-1} \gamma P \qquad \mathcal{E}_{4,j,\ell}(X) := \widetilde{O}(\sigma_0)^{q-1} \mathbb{1}_{v_{j,\ell} \in \mathcal{V}(X)}$$

we first bound the positive gradient (namely for $i = y$):

**Claim C.7** (positive gradient). *Suppose Induction Hypothesis B.3 holds at iteration $t$. For every $(X, y) \in \mathcal{Z}$, every $r \in [m]$, every $\ell \in [2]$, and $i = y$, we have*

(a) $\langle -\nabla_{w_{i,r}} L(F^{(t)}; X, y), v_{i,\ell} \rangle \geq \left( V_{i,r,\ell}(X) - \widetilde{O}(\sigma_p P) \right) \left( 1 - \mathbf{logit}_i(F^{(t)}, X) \right)$

(b) $\langle -\nabla_{w_{i,r}} L(F^{(t)}; X, y), v_{i,\ell} \rangle \leq (V_{i,r,\ell}(X) + \mathcal{E}_1 + \mathcal{E}_3) \left( 1 - \mathbf{logit}_i(F^{(t)}, X) \right)$

(c) *For every $j \in [k] \setminus \{i\}$,*

$$\left| \langle \nabla_{w_{i,r}} L(F^{(t)}, X, y), v_{j,\ell} \rangle \right| \leq \left( 1 - \mathbf{logit}_i(F^{(t)}, X) \right) (\mathcal{E}_{2,i,r}(X) + \mathcal{E}_1 + \mathcal{E}_3 + \mathcal{E}_{4,j,\ell}(X))$$

*Proof of Claim C.7.* We drop the superscript $^{(t)}$ for notational simplicity. Using the gradient formula from (C.1) (in the case of $i = y$), and the orthogonality among feature vectors, we have

$$\langle -\nabla_{w_{i,r}} L(F; X, y), v_{j,\ell} \rangle = (1 - \mathbf{logit}_i(F, X)) \times$$

$$\left( \mathbb{1}_{v_{j,\ell} \in \mathcal{V}(X)} \sum_{p \in \mathcal{P}_{v_{j,\ell}}(X)} \widetilde{\mathsf{ReLU}}'(\langle w_{i,r}, x_p \rangle) z_p + \sum_{p \in [P]} \widetilde{\mathsf{ReLU}}'(\langle w_{i,r}, x_p \rangle) \alpha_{p,v_{j,\ell}} \pm \sum_{p \in [P]} |\langle v_{j,\ell}, \xi_p \rangle| \right)$$

Using the randomness of $\xi_p$, we have (recalling $\sigma_p = \frac{1}{\sqrt{d}\mathsf{polylog}(k)}$ and $\gamma \leq \frac{1}{k}$)

$$\sum_{p \in [P]} |\langle v_{j,\ell}, \xi_p \rangle| \leq \widetilde{O}(\sigma_p \cdot s + \frac{\gamma k}{\sqrt{d}} \cdot P) \ll \widetilde{O}(\sigma_p \cdot P)$$

When $j = i$ we have $v_{i,\ell} \in \mathcal{V}(X)$ so this proves Claim C.7a. Using Induction Hypothesis B.3, we have

- For every $p \in \mathcal{P}_{v_{i,1}}(X) \cup \mathcal{P}_{v_{i,2}}(X)$, we have: $\widetilde{\mathsf{ReLU}}'(\langle w_{i,r}, x_p \rangle) \in [0, 1]$.

- For every $p \in \mathcal{P}(X) \setminus \left( \mathcal{P}_{v_{i,1}}(X) \cup \mathcal{P}_{v_{i,2}}(X) \right)$, we have: $\widetilde{\mathsf{ReLU}}'(\langle w_{i,r}^{(t)}, x_p \rangle) \in [0, \widetilde{O}(\sigma_0^{q-1})]$.

- For every $p \in [P] \setminus \mathcal{P}(X)$, we have: $\widetilde{\mathsf{ReLU}}'(\langle w_{i,r}^{(t)}, x_p \rangle) \in [0, \widetilde{O}((\sigma_0 \gamma k)^{q-1})]$.

Using the sparsity from Def. 3.1, we have $|\mathcal{P}(X) \setminus (\mathcal{P}_{v_{i,1}}(X) \cup \mathcal{P}_{v_{i,2}}(X))| \leq \widetilde{O}(s)$. Combining this with $\alpha_{p,v} \leq \gamma$, and setting $j = i$, this proves Claim C.7b.

Finally, when $j \neq i$, using Induction Hypothesis B.3 we additionally have

- When $v_{j,\ell} \in \mathcal{V}(X)$ and $p \in \mathcal{P}_{v_{j,\ell}}(X)$, we have $\widetilde{\mathsf{ReLU}}'(\langle w_{i,r}, x_p \rangle) \leq \widetilde{O}(\sigma_0^{q-1})$
- For $p \in \mathcal{P}_{v_{i,1}}(X) \cup \mathcal{P}_{v_{i,2}}(X)$, we have a more precise bound using Induction Hypothesis B.3a:

$$\sum_{p \in \mathcal{P}_{v_{i,\ell}}(X)} \widetilde{\mathsf{ReLU}}'(\langle w_{i,r}, x_p \rangle) \leq \sum_{p \in \mathcal{P}_{v_{i,\ell}}(X)} \widetilde{\mathsf{ReLU}}'(\langle w_{i,r}, v_{i,\ell} \rangle + \widetilde{o}(\sigma_0))$$

$$\leq \sum_{p \in \mathcal{P}_{v_{i,\ell}}(X)} \widetilde{\mathsf{ReLU}}'(\langle w_{i,r}, v_{i,\ell} \rangle + \widetilde{o}(\sigma_0)) z_p \leq \sum_{p \in \mathcal{P}_{v_{i,\ell}}(X)} \widetilde{\mathsf{ReLU}}'(\langle w_{i,r}, x_p \rangle) z_p + \widetilde{o}(\sigma_0) = V_{i,r,\ell}(X)$$

Putting them together proves Claim C.7c. $\square$

We also have the following claim about the negative gradient (namely for $i \neq y$), whose proof is completely symmetric to that of Claim C.7 so we ignore here.

**Claim C.8** (negative gradient). *Suppose Induction Hypothesis B.3 holds at iteration $t$. For every $(X, y) \in \mathcal{Z}$, every $r \in [m]$, every $\ell \in [2]$, and $i \in [k] \setminus \{y\}$, we have*

(a) $\langle -\nabla_{w_{i,r}} L(F^{(t)}, X, y), v_{i,\ell} \rangle \geq -\mathbf{logit}_i(F^{(t)}, X)\left(\mathcal{E}_1 + \mathcal{E}_3 + \mathbb{1}_{v_{i,\ell} \in \mathcal{P}(X)} V_{i,r,\ell}(X)\right)$

(b) *For every $j \in [k]$:* $\langle -\nabla_{w_{i,r}} L(F^{(t)}, X, y), v_{j,\ell} \rangle \leq \mathbf{logit}_i(F^{(t)}, X) \widetilde{O}(\sigma_p P)$

(c) *For every $j \in [k] \setminus \{i\}$:* $\langle -\nabla_{w_{i,r}} L(F^{(t)}, X, y), v_{j,\ell} \rangle \geq -\mathbf{logit}_i(F^{(t)}, X)(\mathcal{E}_1 + \mathcal{E}_3 + \mathcal{E}_{4,j,\ell}(X))$

**Function approximations.** Let us denote

$$\Phi_{i,\ell}^{(t)} \stackrel{\text{def}}{=} \sum_{r \in [m]} [\langle w_{i,r}^{(t)}, v_{i,\ell} \rangle]^+ \quad \text{and} \quad \Phi_i^{(t)} \stackrel{\text{def}}{=} \sum_{\ell \in [2]} \Phi_{i,\ell}^{(t)} \tag{C.4}$$

One can easily derive that

**Claim C.9** (function approximation). *Suppose Induction Hypothesis B.3 holds at iteration $t$ and supposes $\leq \widetilde{O}(\frac{1}{\sigma_0^q m})$ and $\gamma \leq \widetilde{O}(\frac{1}{\sigma_0 k (mP)^{1/q}})$. Let $Z_{i,\ell}^{(t)}(X) \stackrel{\text{def}}{=} \mathbb{1}_{v_{i,\ell} \in \mathcal{V}(X)} \left(\sum_{p \in \mathcal{P}_{v_{i,\ell}}(X)} z_p\right)$, we have:*

- *for every $t$, every $(X, y) \in \mathcal{Z}_m$ and $i \in [k]$, or for every $(X, y) \in \mathcal{Z}_s$ and $i \in [k] \setminus \{y\}$,*

$$F_i^{(t)}(X) = \sum_{\ell \in [2]} \left(\Phi_{i,\ell}^{(t)} \times Z_{i,\ell}^{(t)}(X)\right) \pm \widetilde{O}(\sigma_0 + \sigma_0^q sm + (\sigma_0 \gamma k)^q \cdot Pm)$$

$$= \sum_{\ell \in [2]} \left(\Phi_{i,\ell}^{(t)} \times Z_{i,\ell}^{(t)}(X)\right) \pm O\left(\frac{1}{\mathsf{polylog}(k)}\right)$$

- *for every $(X, y) \sim \mathcal{D}$, with probability at least $1 - e^{-\Omega(\log^2 k)}$ it satisfies for every $i \in [k]$,*

$$F_i^{(t)}(X) = \sum_{\ell \in [2]} \left(\Phi_{i,\ell}^{(t)} \times Z_{i,\ell}^{(t)}(X)\right) \pm O\left(\frac{1}{\mathsf{polylog}(k)}\right)$$

## C.2 Useful Claims as Consequences of the Induction Hypothesis

In this sub-section we state some consequences of our Induction Hypothesis B.3. They shall be useful in our later proof of the induction hypothesis.

### C.2.1 Correlation Growth

**Claim C.10** (growth). *Suppose Induction Hypothesis B.3 holds at iteration $t$, then for every $i \in [k]$, suppose $\Lambda_i^{(t)} = O(1/m_0)$, then it satisfies*

$$\Lambda_i^{(t+1)} = \Lambda_i^{(t)} + \Theta\left(\frac{\eta}{k}\right) \widetilde{\mathsf{ReLU}}'(\Lambda_i^{(t)})$$

*Proof of Claim C.10.* Recall $\Lambda_i^{(t)} \stackrel{\text{def}}{=} \max_{r \in [m], \ell \in [2]} [\langle w_{i,r}^{(t)}, v_{i,\ell} \rangle]^+$.

Now, let us take any $r \in [m]$ and $\ell \in [2]$ so that $\langle w_{i,r}^{(t)}, v_{i,\ell} \rangle \geq \widetilde{\Omega}(\sigma_0)$. We first show a lower bound on the increment. By Claim C.7 and Claim C.8,

$$\langle w_{i,r}^{(t+1)}, v_{i,\ell} \rangle \geq \langle w_{i,r}^{(t)}, v_{i,\ell} \rangle + \eta \mathop{\mathbb{E}}_{(X,y) \sim \mathcal{Z}} \left[ \mathbb{1}_{y=i} \left( V_{i,r,\ell}(X) - \widetilde{O}(\sigma_p P) \right) \left( 1 - \mathbf{logit}_i(F^{(t)}, X) \right) \right.$$
$$\left. - \mathbb{1}_{y \neq i} \left( \mathcal{E}_1 + \mathcal{E}_3 + \mathbb{1}_{v_{i,\ell} \in \mathcal{P}(X)} V_{i,r,\ell}(X) \right) \mathbf{logit}_i \left( F^{(t)}, X \right) \right]$$
(C.5)

Recall $V_{i,r,\ell}(X) = \sum_{p \in \mathcal{P}_{v_{i,\ell}}(X)} \widetilde{\mathsf{ReLU}}'(\langle w_{i,r}^{(t)}, x_p \rangle) z_p$. Using Induction Hypothesis B.3, we know that as long as $(X,y) \in \mathcal{Z}_m$, or when $(X,y) \in \mathcal{Z}_s$ but $i \neq y$, it satisfies

$$V_{i,r,\ell}(X) = \sum_{p \in \mathcal{P}_{v_{i,\ell}}(X)} \widetilde{\mathsf{ReLU}}'\left( \langle w_{i,r}^{(t)}, v_{i,\ell} \rangle z_p \pm \widetilde{o}(\sigma_0) \right) z_p$$

- When $i = y$ is the correct label, at least when $(X,y) \in \mathcal{Z}_m$, we have $\sum_{p \in \mathcal{P}_{v_{i,\ell}}(X)} z_p \geq 1$, and together with $|\mathcal{P}_{v_{i,\ell}}| \leq C_p = O(1)$, this tells us $V_{i,r,\ell}(X) \geq \Omega(1) \cdot \widetilde{\mathsf{ReLU}}'\left( \langle w_{i,r}^{(t)}, v_{i,\ell} \rangle \right)$.

- When $i \neq y$ is the wrong label and when $v_{i,\ell} \in \mathcal{P}(X)$, we can use $z_p \leq O(1)$ to derive that $V_{i,r,\ell}(X) \leq O(1) \cdot \widetilde{\mathsf{ReLU}}'\left( \langle w_{i,r}^{(t)}, v_{i,\ell} \rangle \right)$.

Together with $\mathbf{logit}_i \left( F^{(t)}, X \right) \leq O(\frac{1}{k})$ from Claim C.4, we can derive that

$$\langle w_{i,r}^{(t+1)}, v_{i,\ell} \rangle \geq \langle w_{i,r}^{(t)}, v_{i,\ell} \rangle + \eta \mathop{\mathbb{E}}_{(X,y) \sim \mathcal{Z}} \left[ \mathbb{1}_{y=i} \cdot \Omega(1) - O(1) \cdot \mathbb{1}_{y \neq i} \mathbb{1}_{v_{i,\ell} \in \mathcal{P}(X)} \frac{1}{k} \right] \cdot \widetilde{\mathsf{ReLU}}'\left( \langle w_{i,r}^{(t)}, v_{i,\ell} \rangle \right)$$
$$- \eta \widetilde{O}\left( \frac{\sigma_p P + \mathcal{E}_1 + \mathcal{E}_3}{k} \right)$$

Finally, recall the property of our distribution $\mathbf{Pr}_{(X,y) \sim \mathcal{D}}\left[ v_{i,\ell} \in \mathcal{P}(X) \mid i \neq y \right] = \frac{s}{k} \ll o(1)$, we derive that

$$\langle w_{i,r}^{(t+1)}, v_{i,\ell} \rangle \geq \langle w_{i,r}^{(t)}, v_{i,\ell} \rangle + \Omega\left(\frac{\eta}{k}\right) \widetilde{\mathsf{ReLU}}'\left( \langle w_{i,r}^{(t)}, v_{i,\ell} \rangle \right)$$

As for the lower bound, using Claim C.7 and Claim C.8 again, we have

$$\langle w_{i,r}^{(t+1)}, v_{i,\ell} \rangle \leq \langle w_{i,r}^{(t)}, v_{i,\ell} \rangle + \eta \mathop{\mathbb{E}}_{(X,y) \sim \mathcal{Z}} \left[ \mathbb{1}_{y=i} \left( V_{i,r,\ell}(X) + \mathcal{E}_1 + \mathcal{E}_3 \right) \left( 1 - \mathbf{logit}_i(F^{(t)}, X) \right) \right.$$
$$\left. - \mathbb{1}_{y \neq i} \left( \widetilde{O}(\sigma_p P) \right) \mathbf{logit}_i \left( F^{(t)}, X \right) \right]$$

so a completely symmetric argument also shows

$$\langle w_{i,r}^{(t+1)}, v_{i,\ell} \rangle \leq \langle w_{i,r}^{(t)}, v_{i,\ell} \rangle + O\left(\frac{\eta}{k}\right) \widetilde{\mathsf{ReLU}}'\left( \langle w_{i,r}^{(t)}, v_{i,\ell} \rangle \right) \qquad \square$$

Claim C.10 immediately gives the following corollary (using $\Omega(\sigma_0) \leq \Lambda_i^{(0)} \leq \widetilde{O}(\sigma_0)$):

**Claim C.11.** *Suppose Induction Hypothesis B.3 holds for every iteration. Define thresholds (noticing $\Lambda_\varnothing^- \leq \Lambda_\varnothing^+$*

$$\Lambda_\varnothing^- \stackrel{\text{def}}{=} \Theta\left(\frac{\varrho}{\log k}\right) = \widetilde{\Theta}(1) \quad \text{and} \quad \Lambda_\varnothing^+ \stackrel{\text{def}}{=} \Theta\left(\frac{1}{m_0}\right) = \widetilde{\Theta}(1)$$

*Let $T_{0,i}$ be the first iteration so that $\Lambda_i^{(t)} \geq 2\Lambda_\varnothing^-$, and $T_0 \stackrel{\text{def}}{=} \Theta(\frac{k}{\eta \sigma_0^{q-2}})$ (noticing $T_0 \geq T_{0,i}$) Then,*

- *for every $i \in [k]$ and $t \geq T_0$, it satisfies $\Lambda_i^{(t)} \geq \Lambda_\varnothing^+$*
- *for every $i \in [k]$ and $t \geq T_{0,i}$, it satisfies $\Lambda_i^{(t)} \geq \Lambda_\varnothing^-$*

### C.2.2 Single-View Error Till the End

In this subsection we present a claim to bound the "convergence" (namely, the $(1 - \mathbf{logit}_y(F^{(t)}, X))$ part) for every single-view data from $T_0$ till the end.

**Claim C.12** (single view till the end). *Suppose Induction Hypothesis B.3 holds for all iterations $t < T$ and $\gamma \leq \widetilde{O}(\sigma_0 k)$. We have that*

*(a) for every $(X, y) \in \mathcal{Z}_s$, for every $r \in [m]$, every $\ell \in [2]$, every $p \in \mathcal{P}_{v_{y,\ell}}(X)$*

$$\sum_{t=T_0}^T \left(1 - \mathbf{logit}_y(F^{(t)}, X)\right) \widetilde{\mathsf{ReLU}}'(\langle w_{y,r}, x_p \rangle) \leq \widetilde{O}\left(\frac{N}{\eta}\right)$$

*(b) for every $(X, y) \in \mathcal{Z}_s$,*

$$\sum_{t=T_0}^T \left(1 - \mathbf{logit}_y(F^{(t)}, X)\right) \leq \widetilde{O}\left(\frac{N}{\eta \rho^{q-1}}\right)$$

Before proving Claim C.12, we first establish a simple claim to bound how the (correlation with the) noise term grows on single view data. This is used to show that the learner learns most single-view data through *memorization*.

**Claim C.13** (noise lower bound). *Suppose Induction Hypothesis B.3 holds at iteration $t$.*

*(a) For every $(X, y) \in \mathcal{Z}_s$, every $\ell \in [2]$, for every $p \in \mathcal{P}_{v_{y,\ell}}(X)$,*

$$\langle w_{i,r}^{(t+1)}, \xi_p \rangle \geq \langle w_{i,r}^{(t)}, \xi_p \rangle - \frac{\eta}{\sqrt{d}} + \widetilde{\Omega}\left(\frac{\eta}{N}\right) \widetilde{\mathsf{ReLU}}'(\langle w_{i,r}^{(t)}, x_p \rangle) \left(1 - \mathbf{logit}_i(F^{(t)}, X^*)\right) \geq \cdots \geq -\frac{\eta T}{\sqrt{d}}$$

*(b) For every $(X, y) \in \mathcal{Z}_s$, every $\ell \in [2]$,*

$$\sum_{p \in P_{v_{y,\ell}}(X)} \langle w_{y,r}^{(t+1)}, \xi_p \rangle \geq \sum_{p \in P_{v_{y,\ell}}(X)} \langle w_{y,r}^{(t)}, \xi_p \rangle - \frac{O(\eta)}{\sqrt{d}}$$
$$+ \widetilde{\Omega}\left(\frac{\eta}{N}\right) \widetilde{\mathsf{ReLU}}'\left(\rho \cdot \langle w_{y,r}^{(t)}, v_{y,\ell} \rangle - \widetilde{O}(\eta T/\sqrt{d} + \sigma_0 \gamma k)\right) \left(1 - \mathbf{logit}_y(F^{(t)}, X)\right)$$

*Proof of Claim C.13.* For every $(X^*, y^*) \in \mathcal{Z}_s$, every $i \in [k]$, every $\ell \in [2]$, and every $p^* \in P_{v_{i,\ell}}(X^*)$, one can calculate that

$$\langle w_{i,r}^{(t+1)}, \xi_{p^*} \rangle = \langle w_{i,r}^{(t)}, \xi_{p^*} \rangle + \eta \mathop{\mathbb{E}}_{(X,y) \sim \mathcal{Z}} \left[ \mathbb{1}_{y=i} \left(\sum_{p \in [P]} \widetilde{\mathsf{ReLU}}'(\langle w_{i,r}^{(t)}, x_p \rangle) \langle x_p, \xi_{p^*} \rangle\right) \left(1 - \mathbf{logit}_i(F^{(t)}, X)\right) \right.$$
$$\left. - \mathbb{1}_{y \neq i} \left(\sum_{p \in [P]} \widetilde{\mathsf{ReLU}}'(\langle w_{i,r}^{(t)}, x_p \rangle) \langle x_p, \xi_{p^*} \rangle\right) \mathbf{logit}_i(F^{(t)}, X) \right]$$

Note when $X \neq X^*$, we have $|\langle x_p, \xi_{p^*}\rangle| \leq \widetilde{O}(\sigma_p) \leq o(\frac{1}{\sqrt{d}})$; and when $X = X^*$ but $p \neq p^*$, we also have $|\langle x_p, \xi_{p^*}\rangle| \leq \widetilde{O}(\sigma_p) \leq o(\frac{1}{\sqrt{d}})$. Therefore, when $i = y^*$,

$$\langle w_{i,r}^{(t+1)}, \xi_{p^*}\rangle = \langle w_{i,r}^{(t)}, \xi_{p^*}\rangle + \widetilde{\Theta}\left(\frac{\eta}{N}\right)\widetilde{\mathsf{ReLU}}'(\langle w_{i,r}^{(t)}, x_{p^*}\rangle)\left(1 - \mathbf{logit}_i(F^{(t)}, X^*)\right) \pm \frac{\eta}{\sqrt{d}} \quad (C.6)$$

Using the non-negativity of $\widetilde{\mathsf{ReLU}}'$ we arrive at the first conclusion. Next, using Induction Hypothesis B.3d, we have $\langle w_{i,r}^{(t)}, x_{p^*}\rangle = \langle w_{i,r}^{(t)}, v_{i,\ell}\rangle z_{p^*} + \langle w_{i,r}^{(t)}, \xi_{p^*}\rangle \pm \widetilde{O}(\sigma_0\gamma k)$. Also, recall from Def. 3.1 that $\sum_{p^* \in P_{v_{i,\ell}}(X^*)} z_{p^*} \geq \Omega(\rho)$. Therefore, when summing over constantly many $p^* \in P_{v_{i,\ell}}(X^*)$ we have

$$\sum_{p^* \in P_{v_{i,\ell}}(X^*)} \langle w_{i,r}^{(t+1)}, \xi_{p^*}\rangle \geq \sum_{p^* \in P_{v_{i,\ell}}(X^*)} \langle w_{i,r}^{(t)}, \xi_{p^*}\rangle - \frac{O(\eta)}{\sqrt{d}}$$
$$+ \widetilde{\Omega}\left(\frac{\eta}{N}\right)\widetilde{\mathsf{ReLU}}'\left(\rho \cdot \langle w_{i,r}^{(t)}, v_{i,\ell}\rangle - \widetilde{O}(\eta T/\sqrt{d} + \sigma_0\gamma k)\right)\left(1 - \mathbf{logit}_i(F^{(t)}, X^*)\right)$$

This arrives at our second conclusion. □

*Proof of Claim C.12.* We now prove Claim C.12 using Claim C.13. Let us denote $i = y$.

Claim C.12a is in fact a direct corollary of Claim C.13a, because once the summation has reached $\widetilde{\Omega}(\frac{N}{\eta})$ at some iteration $t = t_0$, then according to Claim C.13a, we must have already satisfied

$$\forall t \geq t_0: \quad \langle w_{i,r}^{(t)}, \xi_p\rangle \geq \mathsf{polylog}(k)$$

but according to $\langle w_{i,r}^{(t)}, x_p\rangle = \langle w_{i,r}^{(t)}, v_{i,\ell}\rangle z_p + \langle w_{i,r}^{(t)}, \xi_p\rangle \pm \widetilde{O}(\sigma_0\gamma k)$ from Induction Hypothesis B.3c, and using $\langle w_{i,r}^{(t)}, v_{i,\ell}\rangle \geq -1$ from Induction Hypothesis B.3h, we immediately have

$$F_i^{(t)}(X) \geq \langle w_{i,r}^{(t)}, x_{p^*}\rangle - O(1) \geq \mathsf{polylog}(k)$$

while at the same time, one can easily derive (recall (C.3)) that $F_j^{(t)}(X) \leq m_0 \cdot \Lambda_i^{(t)} \cdot \Gamma \leq O(1)$ for every $j \neq i$. Therefore, we have $1 - \mathbf{logit}_y(F^{(t)}, X) \leq e^{-\log^5 k}$ for every $t \geq t_1$. This proves the Claim C.12a.

Next, we move to Claim C.12b. We prove by way of contradiction and suppose

$$\sum_{t \geq T_0}\left(1 - \mathbf{logit}_i\left(F^{(t)}, X\right)\right) \geq \widetilde{\Omega}\left(\frac{N}{\eta\rho^{q-1}}\right)$$

Using $\Lambda_i^{(t)} = \max_{r \in [m], \ell \in [2]}[\langle w_{i,r}^{(t)}, v_{i,\ell}\rangle]^+ \geq \widetilde{\Omega}(1)$ from Claim C.11 and the definition of $\mathcal{M}_i^{(0)}$, we have

$$\sum_{(r,\ell) \in \mathcal{M}_i^{(0)} \times [2]} \mathbb{1}_{\langle w_{i,r}^{(t)}, v_{i,\ell}\rangle \geq \widetilde{\Omega}(1)} \sum_{t \geq T_0}\left(1 - \mathbf{logit}_i\left(F^{(t)}, X\right)\right) \geq \widetilde{\Omega}\left(\frac{N}{\eta\rho^{q-1}}\right)$$

Note that when $\langle w_{i,r}^{(t)}, v_{i,\ell}\rangle \geq \widetilde{\Omega}(1)$ and $\sum_{p \in \mathcal{P}_{v_{i,\ell}}(X)}\langle w_{i,r}^{(t)}, \xi_p\rangle \geq \mathsf{polylog}(k)$ simultaneously hold, there must exists some $p^* \in \mathcal{P}_{v_{i,\ell}}(X)$ so that $\langle w_{i,r}^{(t)}, \xi_{p^*}\rangle \geq \mathsf{polylog}(k)$, but according to Induction Hypothesis B.3d, we have (noticing $\widetilde{\mathsf{ReLU}}$ is in the linear regime now because $\langle w_{i,r}^{(t)}, v_{i,\ell}\rangle \geq 0$)

$$F_i^{(t)}(X) \geq \langle w_{i,r}^{(t)}, x_{p^*}\rangle - O(1) = \langle w_{i,r}^{(t)}, v_{i,\ell}\rangle z_{p^*} + \langle w_{i,r}^{(t)}, \xi_{p^*}\rangle \pm \widetilde{O}(\sigma_0\gamma k) - O(1) \geq \mathsf{polylog}(k)$$

In contrast, one can derive (recall (C.3)) that $F_j^{(t)}(X) \leq m_0 \cdot \Lambda_i^{(t)} \cdot \Gamma \leq O(1)$ for every $j \neq i$. This means $1 - \mathbf{logit}_i\left(F^{(t)}, X\right) \ll e^{-\log^5 k}$. In other words,

$$\sum_{t \geq T_0} \sum_{(r,\ell) \in \mathcal{M}_i^{(0)} \times [2]} \mathbb{1}_{\langle w_{i,r}^{(t)}, v_{i,\ell}\rangle \geq \widetilde{\Omega}(1)} \mathbb{1}_{\sum_{p \in \mathcal{P}_{v_{i,\ell}}(X)} \langle w_{i,r}^{(t)}, \xi_p\rangle \leq \mathsf{polylog}(k)} \left(1 - \mathbf{logit}_i\left(F^{(t)}, X\right)\right) \geq \widetilde{\Omega}\left(\frac{N}{\eta \rho^{q-1}}\right)$$

Now we partition the iterations between $T_0$ and $T$ into $4m_0$ stages of consecutive iterations, denoted by $\mathcal{T}_1, \ldots \mathcal{T}_{4m_0}$, so that each of them have a similar partial sum in the above summation. In symbols: $\forall g \in [4m_0]$:

$$\sum_{t \in \mathcal{T}_g} \sum_{(r,\ell) \in \mathcal{M}_i^{(0)} \times [2]} \mathbb{1}_{\langle w_{i,r}^{(t)}, v_{i,\ell}\rangle \geq \widetilde{\Omega}(1)} \mathbb{1}_{\sum_{p \in \mathcal{P}_{v_{i,\ell}}(X)} \langle w_{i,r}^{(t)}, \xi_p\rangle \leq \mathsf{polylog}(k)} \left(1 - \mathbf{logit}_i\left(F^{(t)}, X\right)\right) \geq \widetilde{\Omega}\left(\frac{N}{\eta \rho^{q-1}}\right)$$
(C.7)

Let us first look at stage $\mathcal{T}_1$. By averaging, there exists some $(r, \ell) = (r_1^*, \ell_1^*) \in \mathcal{M}_i^{(0)} \times [2]$ so that

$$\sum_{t \in \mathcal{T}_1} \mathbb{1}_{\langle w_{i,r}^{(t)}, v_{i,\ell}\rangle \geq \widetilde{\Omega}(1)} \mathbb{1}_{\sum_{p \in \mathcal{P}_{v_{i,\ell}}(X)} \langle w_{i,r}^{(t)}, \xi_p\rangle \leq \mathsf{polylog}(k)} \left(1 - \mathbf{logit}_i\left(F^{(t)}, X\right)\right) \geq \widetilde{\Omega}\left(\frac{N}{\eta \rho^{q-1}}\right)$$

Applying Claim C.13b, we know that after stage $\mathcal{T}_1$ (namely, for any $t \in [\max \mathcal{T}_1, T]$), it satisfies

$$\text{for } (r, \ell) = (r_1^*, \ell_1^*) \qquad \sum_{p \in \mathcal{P}_{v_{i,\ell}}(X)} \langle w_{i,r}^{(t)}, \xi_p\rangle \geq \widetilde{\Omega}\left(\frac{N}{\eta \rho^{q-1}}\right) \cdot \widetilde{\Omega}\left(\frac{\eta}{N}\rho^{q-1}\right) > \mathsf{polylog}(k)$$

Continuing to stage $\mathcal{T}_2$, by averaging again, we can find some other $(r, \ell) = (r_2^*, \ell_2^*) \in \mathcal{M}_i^{(0)} \times [2]$ so that

$$\sum_{t \in \mathcal{T}_2} \mathbb{1}_{\langle w_{i,r}^{(t)}, v_{i,\ell}\rangle \geq \widetilde{\Omega}(1)} \mathbb{1}_{\sum_{p \in \mathcal{P}_{v_{i,\ell}}(X)} \langle w_{i,r}^{(t)}, \xi_p\rangle \leq \mathsf{polylog}(k)} \left(1 - \mathbf{logit}_i\left(F^{(t)}, X\right)\right) \geq \widetilde{\Omega}\left(\frac{N}{\eta \rho^{q-1}}\right)$$

From the conclusion of of the previous stage, it must satisfy that $(r_2^*, \ell_2^*) \neq (r_1^*, \ell_1^*)$. A similar derivation also tells us that after stage $\mathcal{T}_2$ (namely, for any $t \in [\max \mathcal{T}_2, T]$), it satisfies

$$\text{for } (r, \ell) = (r_2^*, \ell_2^*) \qquad \sum_{p \in \mathcal{P}_{v_{i,\ell}}(X)} \langle w_{i,r}^{(t)}, \xi_p\rangle > \mathsf{polylog}(k)$$

We continue this argument until we finish stage $\mathcal{T}_{2m_0}$. At this point, we know for every $t \in [\max \mathcal{T}_{2m_0}, T]$

$$\text{for all } (r, \ell) \in \mathcal{M}_i^{(0)} \times [2] \qquad \sum_{p \in \mathcal{P}_{v_{i,\ell}}(X)} \langle w_{i,r}^{(t)}, \xi_p\rangle > \mathsf{polylog}(k)$$

This contradicts (C.7) for any $g > 2m_0$. $\square$

### C.2.3 Multi-View Error Till the End

In this subsection we present a claim to bound the "convergence" (namely, the $\left(1 - \mathbf{logit}_y\left(F^{(t)}, X\right)\right)$ part) for the average multi-view data from $T_0$ till the end.

**Claim C.14** (multi-view till the end). *Suppose Induction Hypothesis B.3 holds for every iteration $t < T$, and suppose $N_s \leq \frac{k^2 \rho^{q-1}}{s}$, then*

$$\sum_{t=T_0}^{T} \mathbb{E}_{(X,y) \sim \mathcal{Z}_m}\left[1 - \mathbf{logit}_y\left(F^{(t)}, X\right)\right] \leq \widetilde{O}\left(\frac{k}{\eta}\right) + \widetilde{O}\left(\frac{sN_s}{\eta k \rho^{q-1}}\right) \leq \widetilde{O}\left(\frac{k}{\eta}\right)$$

In fact, Claim C.14 is a direct corollary of the following claim, combined with $\Lambda_i^{(t)} = \widetilde{O}(1)$ from Induction Hypothesis B.3g, and with the convergence Claim C.12b for single-view data.

**Claim C.15.** *Suppose Induction Hypothesis B.3 holds at iteration $t$ and $t \geq T_0$, then*
$$\sum_{i \in [k]} \Lambda_i^{(t+1)} \geq \sum_{i \in [k]} \Lambda_i^{(t)} + \Omega(\eta) \times \mathop{\mathbb{E}}_{(X,y) \sim \mathcal{Z}_m} \left[ 1 - \mathbf{logit}_y \left( F^{(t)}, X \right) \right]$$
$$- \eta O\left(\frac{s}{k} \frac{N_s}{N}\right) \mathop{\mathbb{E}}_{(X,y) \sim \mathcal{Z}_s} \left[ 1 - \mathbf{logit}_y \left( F^{(t)}, X \right) \right]$$

*Proof of Claim C.15.* Recall $\Lambda_i^{(t)} \stackrel{\text{def}}{=} \max_{r \in [m], \ell \in [2]} [\langle w_{i,r}^{(t)}, v_{i,\ell} \rangle]^+$. Let us take $r, \ell$ to be this argmax so that Claim C.11 tells us $\langle w_{i,r}^{(t)}, v_{i,\ell} \rangle \geq \Lambda_\varnothing^+ = \widetilde{\Theta}(1)$. By Claim C.7 and Claim C.8,
$$\langle w_{i,r}^{(t+1)}, v_{i,\ell} \rangle \geq \langle w_{i,r}^{(t)}, v_{i,\ell} \rangle + \eta \mathop{\mathbb{E}}_{(X,y) \sim \mathcal{Z}} \left[ \mathbb{1}_{y=i} \left( V_{i,r,\ell}(X) - \widetilde{O}(\sigma_p P) \right) \left( 1 - \mathbf{logit}_i(F^{(t)}, X) \right) \right.$$
$$\left. - \mathbb{1}_{y \neq i} \left( \mathcal{E}_1 + \mathcal{E}_3 + \mathbb{1}_{v_{i,\ell} \in \mathcal{P}(X)} V_{i,r,\ell}(X) \right) \mathbf{logit}_i \left( F^{(t)}, X \right) \right]$$

Recall $V_{i,r,\ell}(X) = \sum_{p \in \mathcal{P}_{v_{i,\ell}}(X)} \widetilde{\mathsf{ReLU}}'(\langle w_{i,r}^{(t)}, x_p \rangle) z_p$. Using Induction Hypothesis B.3a, we know that as long as $(X, y) \in \mathcal{Z}_m$, or when $(X, y) \in \mathcal{Z}_s$ but $i \neq y$, it satisfies
$$V_{i,r,\ell}(X) = \sum_{p \in \mathcal{P}_{v_{i,\ell}}(X)} \widetilde{\mathsf{ReLU}}' \left( \langle w_{i,r}^{(t)}, v_{i,\ell} \rangle z_p \pm \widetilde{o}(\sigma_0) \right) z_p$$

Since $\langle w_{i,r}^{(t)}, v_{i,\ell} \rangle \geq \Lambda_\varnothing^+ = \Theta(\frac{1}{m_0}) \gg \varrho$ (see Claim C.11) and since $|\mathcal{P}_{v_{i,\ell}}(X)| \leq O(1)$, for most of $p \in \mathcal{P}_{v_{i,\ell}}$ we must be already in the linear regime of $\widetilde{\mathsf{ReLU}}$ so
$$0.9 \sum_{p \in \mathcal{P}_{v_{i,\ell}}(X)} z_p \leq V_{i,r,\ell}(X) \leq \sum_{p \in \mathcal{P}_{v_{i,\ell}}(X)} z_p$$

According to our choice of the distribution (see Def. 3.1):

- When $(X, y) \in \mathcal{Z}_m$ and $y = i$, we have $V_{i,r,\ell}(X) \geq 0.9$.
- When $(X, y) \in \mathcal{Z}_s$ and $y = i$, we have $V_{i,r,\ell}(X) \geq 0$.
- When $(X, y) \in \mathcal{Z}_m$, $y \neq i$ and $v_{i,\ell} \in \mathcal{P}(X)$, we have $V_{i,r,\ell}(X) \leq 0.4$.
- When $(X, y) \in \mathcal{Z}_s$, $y \neq i$ and $v_{i,\ell} \in \mathcal{P}(X)$, we have $V_{i,r,\ell}(X) \leq \Gamma \leq 1$.

Together, we derive that
$$\langle w_{i,r}^{(t+1)}, v_{i,\ell} \rangle \geq \langle w_{i,r}^{(t)}, v_{i,\ell} \rangle + \eta \mathop{\mathbb{E}}_{(X,y) \sim \mathcal{Z}_m} \left[ 0.89 \cdot \mathbb{1}_{y=i} \left( 1 - \mathbf{logit}_i(F^{(t)}, X) \right) \right]$$
$$- \eta \mathop{\mathbb{E}}_{(X,y) \sim \mathcal{Z}_m} \left[ \mathbb{1}_{y \neq i} (\mathcal{E}_1 + \mathcal{E}_3 + 0.4 \mathbb{1}_{v_{i,\ell} \in \mathcal{P}(X)}) \mathbf{logit}_i \left( F^{(t)}, X \right) \right]$$
$$- O\left(\frac{\eta N_s}{N}\right) \mathop{\mathbb{E}}_{(X,y) \sim \mathcal{Z}_s} \left[ \mathbb{1}_{y=i} \cdot \widetilde{O}(\sigma_p P) \left( 1 - \mathbf{logit}_y \left( F^{(t)}, X \right) \right) \right]$$
$$- O\left(\frac{\eta N_s}{kN}\right) \mathop{\mathbb{E}}_{(X,y) \sim \mathcal{Z}_s} \left[ \mathbb{1}_{y \neq i} \left( \mathcal{E}_1 + \mathcal{E}_3 + \mathbb{1}_{v_{i,\ell} \in \mathcal{P}(X)} \right) \left( 1 - \mathbf{logit}_y \left( F^{(t)}, X \right) \right) \right]$$
(C.8)

Above, we have applied Claim C.4 which says for $(X, y) \in \mathcal{Z}_s$, it holds that $\mathbf{logit}_i(F^{(t)}, X) \leq O\left(\frac{1}{k}\right) \left( 1 - \mathbf{logit}_y(F^{(t)}, X) \right)$.

Finally, substituting $\mathbb{1}_{v_{i,\ell} \in \mathcal{P}(X)}$ with the naive upper bound $\mathcal{E}_1 + \mathcal{E}_3 + 0.4 \mathbb{1}_{v_{i,\ell} \in \mathcal{P}(X)} \leq 0.41$, we

have

$$\langle w_{i,r}^{(t+1)}, v_{i,\ell} \rangle \geq \langle w_{i,r}^{(t)}, v_{i,\ell} \rangle + \eta \mathop{\mathbb{E}}_{(X,y) \sim \mathcal{Z}_m} \left[ 0.89 \cdot \mathbb{1}_{y=i} \left( 1 - \mathbf{logit}_i(F^{(t)}, X) \right) - 0.41 \cdot \mathbb{1}_{y \neq i} \mathbf{logit}_i \left( F^{(t)}, X \right) \right]$$

$$- O\left(\frac{\eta N_s}{kN}\right) \mathop{\mathbb{E}}_{(X,y) \sim \mathcal{Z}_s} \left[ \left( k \mathbb{1}_{y=i} \cdot \widetilde{O}(\sigma_p P) + \mathbb{1}_{y \neq i} \left( \mathcal{E}_1 + \mathcal{E}_3 + \mathbb{1}_{v_{i,\ell} \in \mathcal{P}(X)} \right) \right) \left( 1 - \mathbf{logit}_y \left( F^{(t)}, X \right) \right) \right]$$

Summing up over all $i \in [k]$, and using $v_{i,\ell} \in \mathcal{P}(X)$ with probability $\frac{s}{k}$ when $i \neq y$, we finish the proof. $\square$

### C.2.4 Multi-View Individual Error

Our next claim states that up to a polynomial factor, the error on any individual multi-view data is bounded by the training error.

**Claim C.16** (multi-view individual error). *For every $t \geq 0$, every $(X, y) \in \mathcal{Z}_m$,*

$$1 - \mathbf{logit}_y \left( F^{(t)}, X \right) \leq \widetilde{O}\left(\frac{k^4}{s^2}\right) \cdot \mathop{\mathbb{E}}_{(X,y) \sim \mathcal{Z}_m} \left[ 1 - \mathbf{logit}_y \left( F^{(t)}, X \right) \right]$$

*(The same also holds w.p. $\geq 1 - e^{-\Omega(\log^2 k)}$ for every $(X, y) \sim \mathcal{D}_m$ on the left hand side.)*

*Furthermore, if $\mathbb{E}_{(X,y) \sim \mathcal{Z}_m} \left[ 1 - \mathbf{logit}_y \left( F^{(t)}, X \right) \right] \leq \frac{1}{k^4}$ is sufficiently small, we have $0.4 \Phi_i^{(t)} - \Phi_j^{(t)} \leq -\Omega(\log k)$ for every pair $i, j \in [k]$.*

*Proof.* For a data point $(X, y) \in \mathcal{Z}_m$, let us denote by $\mathcal{H}(X)$ be the set of all $i \in [k] \setminus \{y\}$ such that,

$$\sum_{\ell \in [2]} \sum_{p \in \mathcal{P}_{v_{i,\ell}}(X)} z_p \geq 0.8 - \frac{1}{100 \log(k)}, \quad \sum_{\ell \in [2]} \sum_{p \in \mathcal{P}_{v_{y,\ell}}(X)} z_p \leq 2 + \frac{1}{100 \log(k)}$$

Now, suppose $1 - \mathbf{logit}_y \left( F^{(t)}, X \right) = \xi(X)$, then using $\min\{1, \beta\} \leq 2\left(1 - \frac{1}{1+\beta}\right)$, we have

$$\min \left\{ 1, \sum_{i \in [k] \setminus \{y\}} e^{F_i^{(t)}(X) - F_y^{(t)}(X)} \right\} \leq 2\xi(X)$$

By Claim C.9 and our definition of $\mathcal{H}(X)$, this implies that

$$\min \left\{ 1, \sum_{i \in \mathcal{H}(X)} e^{0.4 \Phi_i^{(t)} - \Phi_y^{(t)}} \right\} \leq 4\xi(X)$$

If we denote by $\psi = \mathbb{E}_{(X,y) \sim \mathcal{Z}_m} \left[ 1 - \mathbf{logit}_y \left( F^{(t)}, X \right) \right]$, then

$$\mathop{\mathbb{E}}_{(X,y) \sim \mathcal{Z}_m} \left[ \min \left\{ 1, \sum_{i \in \mathcal{H}(X)} e^{0.4 \Phi_i^{(t)} - \Phi_y^{(t)}} \right\} \right] \leq 4\psi$$

$$\implies \mathop{\mathbb{E}}_{(X,y) \sim \mathcal{Z}_m} \left[ \sum_{i \in \mathcal{H}(X)} \min \left\{ \frac{1}{k}, e^{0.4 \Phi_i^{(t)} - \Phi_y^{(t)}} \right\} \right] \leq 4\psi$$

Notice that we can rewrite the LHS so that

$$\mathop{\mathbb{E}}_{(X,y) \sim \mathcal{Z}_m} \left[ \sum_{j \in [k]} \mathbb{1}_{j=y} \sum_{i \in [k]} \mathbb{1}_{i \in \mathcal{H}(X)} \min \left\{ \frac{1}{k}, e^{0.4 \Phi_i^{(t)} - \Phi_j^{(t)}} \right\} \right] \leq 4\psi$$

$$\implies \sum_{j \in [k]} \sum_{i \in [k]} \mathbb{1}_{i \neq j} \mathop{\mathbb{E}}_{(X,y) \sim \mathcal{Z}_m} \left[ \mathbb{1}_{j=y} \mathbb{1}_{i \in \mathcal{H}(X)} \right] \min \left\{ \frac{1}{k}, e^{0.4 \Phi_i^{(t)} - \Phi_j^{(t)}} \right\} \leq 4\psi$$

Note however, that for every $i \neq j \in [k]$, the probability of generating a multi-view sample $(X, y) \in \mathcal{Z}_m$ with $y = j$ and $i \in \mathcal{H}(X)$ is at least $\widetilde{\Omega}\left(\frac{1}{k} \cdot \frac{s^2}{k^2}\right)$. This implies

$$\sum_{j \in [k]} \sum_{i \in [k] \setminus i} \min \left\{ \frac{1}{k}, e^{0.4 \Phi_i^{(t)} - \Phi_j^{(t)}} \right\} \leq \widetilde{O}\left(\frac{k^3}{s^2} \psi\right) \quad \text{(C.9)}$$

Finally, using $1 - \frac{1}{1+\beta} \leq \min\{1, \beta\}$, it is easy to see for every $(X, y) \in \mathcal{Z}_m$

$$1 - \mathbf{logit}_y\left(F^{(t)}, X\right) = \min\left\{1, \sum_{i \in [k] \setminus \{y\}} 2e^{0.4\Phi_i^{(t)} - \Phi_y^{(t)}}\right\} \leq k \cdot \sum_{i \in [k] \setminus \{y\}} \min\left\{\frac{1}{k}, e^{0.4\Phi_i^{(t)} - \Phi_y^{(t)}}\right\} \leq \widetilde{O}\left(\frac{k^4}{s^2}\psi\right)$$

We complete the proof.

Note that if one replaces $(X, y) \in \mathcal{Z}_m$ with $(X, y) \in \mathcal{D}_m$, we also have

$$1 - \mathbf{logit}_y\left(F^{(t)}, X\right) = \min\left\{1, \sum_{i \in [k] \setminus \{y\}} 2e^{0.4\Phi_i^{(t)} - \Phi_y^{(t)}}\right\}$$

with high probability, so the same result also holds.

Note also (C.9) implies if $\mathbb{E}_{(X,y) \sim \mathcal{Z}_m}\left[1 - \mathbf{logit}_y\left(F^{(t)}, X\right)\right] \leq \frac{1}{k^4}$ is sufficiently small, we have $0.4\Phi_i^{(t)} - \Phi_j^{(t)} \leq -\Omega(\log k)$ for every pair $i \neq j$. Using the non-negativity of $\Phi_i^{(t)}$, we know the relationship holds also when $i = j$. □

### C.2.5 Multi-View Error in Stage 2

As we shall see later, our final proof is divided into three stages for each index $i \in [k]$: the first stage is for $t \leq T_{0,i}$, the second stage is for all $t \in [T_{0,i}, T_0]$, and the third iteration is for $t > T_0$. We have the following claim to bound the maximum error of multi-view data during the second stage.

**Claim C.17** (multi-view stage 2). *Suppose Induction Hypothesis B.3 holds for every iteration $t \leq T_0$, and $\Upsilon = \widetilde{\Theta}(\frac{1}{k^{0.2}})$ is a parameter. Then, for every $i \in [k]$*

(a)
$$\sum_{t=T_{0,i}}^{T_0} \mathbb{E}_{(X,y) \sim \mathcal{Z}_m}\left[\mathbb{1}_{y=i}\left(1 - \mathbf{logit}_y\left(F^{(t)}, X\right)\right)\right] \leq O\left(\frac{s}{k}T_0\Upsilon\right) + \widetilde{O}\left(\frac{1}{\eta}\right)$$

(b) *for every $t \in [T_{0,i}, T_0]$, every $j \in [k] \setminus \{i\}$, every $\ell \in [2]$,*

$$\mathbb{E}_{(X,y) \sim \mathcal{Z}_m}\left[\mathbb{1}_{y \neq i}\mathbf{logit}_i\left(F^{(t)}, X\right)\right] \leq O(\frac{1}{k})$$

$$\mathbb{E}_{(X,y) \sim \mathcal{Z}_m}\left[\mathbb{1}_{y \neq i}\mathbb{1}_{v_{j,\ell} \in \mathcal{P}(X)}\mathbf{logit}_i\left(F^{(t)}, X\right)\right] \leq O(\frac{s}{k^2})$$

In order to prove Claim C.17 we first establish the following claim.

**Claim C.18.** *Let $\Upsilon$ be any $\Upsilon \in \left[\frac{1}{k}, \frac{1}{s}\right]$, and recall $\Phi_i^{(t)} \stackrel{\text{def}}{=} \sum_{r \in [m], \ell \in [2]}[\langle w_{i,r}^{(t)}, v_{i,\ell} \rangle]^+$ from (C.4). Then, letting $T_1 \stackrel{\text{def}}{=} \widetilde{\Theta}\left(\frac{k^{2.5}\Upsilon^{2.5}}{\eta}\right)$ and suppose Induction Hypothesis B.3 holds for all iterations $t \leq T_1$. Then,*

$$\forall t \leq T_1, \quad \forall i \in [k]: \quad e^{0.4\Phi_i^{(t)}} \leq k\Upsilon .$$

*(Note when $\Upsilon \leq \widetilde{O}(k^{-0.2})$ we have $T_0 \leq T_1$.)*

*Proof of Claim C.18.* Recall from Induction Hypothesis B.3i that for those $r \in [m] \setminus \mathcal{M}_i^{(0)}$ and $\ell \in [2]$, it satisfies $[\langle w_{i,r}^{(t)}, v_{i,\ell} \rangle]^+ \leq \widetilde{O}(\sigma_0)$ and their summation does not exceed $\frac{1}{\text{polylog}(k)}$ due to our choice of $m$. Thus, to prove this claim, it suffices to slightly abuse the notation and think of

$$\widehat{\Phi}^{(t)} = \max_{i \in [k]} \sum_{(r,\ell) \in \mathcal{M}_i^{(0)} \times [2]}[\langle w_{i,r}^{(t)}, v_{i,\ell} \rangle]^+$$

Let $i$ be this argmax in $\widehat{\Phi}^{(t)}$. For every $(r, \ell) \in \mathcal{M}_i^{(0)} \times [2]$, by Claim C.7 and Claim C.8,

$$\langle w_{i,r}^{(t+1)}, v_{i,\ell} \rangle \leq \langle w_{i,r}^{(t)}, v_{i,\ell} \rangle + \eta \mathop{\mathbb{E}}_{(X,y)\sim\mathcal{Z}} \Big[ \mathbb{1}_{y=i} (V_{i,r,\ell}(X) + \mathcal{E}_1 + \mathcal{E}_3) \Big( 1 - \mathbf{logit}_i(F^{(t)}, X) \Big)$$
$$+ \mathbb{1}_{y\neq i} \Big( \widetilde{O}(\sigma_p P) \Big) \mathbf{logit}_i \Big( F^{(t)}, X \Big) \Big] \quad \text{(C.10)}$$

where recall

$$V_{i,r,\ell}(X) = \sum_{p \in \mathcal{P}_{v_{i,\ell}}(X)} \widetilde{\mathsf{ReLU}}'(\langle w_{i,r}, x_p\rangle) z_p \leq \sum_{p \in \mathcal{P}_{v_{i,\ell}}(X)} z_p \leq O(1)$$

Therefore,

$$\langle w_{i,r}^{(t+1)}, v_{i,\ell} \rangle \leq \langle w_{i,r}^{(t)}, v_{i,\ell} \rangle + O(\eta) \Big( \mathop{\mathbb{E}}_{(X,y)\sim\mathcal{Z}} \big[ \mathbb{1}_{y=i} (1 - \mathbf{logit}_y(F^{(t)}, X)) \big] + \widetilde{O}(\sigma_p P) \Big)$$

Note that single-view data contribute to at most $O(\frac{\eta N_s}{kN})$ on the RHS of (C.10), so we only focus on those $(X, y) \in \mathcal{Z}_m$. By Claim C.9 we know $F_y^{(t)}(X) \geq \Phi_y^{(t)} - \frac{1}{\mathsf{polylog}(k)}$ and for $j \in [k] \setminus \{y\}$,

- W.p. $1 - (1 - \frac{s}{k})^2$, both $v_{j,1}, v_{j,2} \notin \mathcal{P}(X)$, and in this case $F_j^{(t)}(X) \leq \frac{1}{\mathsf{polylog}(k)}$;
- W.p. $(1 - \frac{s}{k})^2$, at least one of $v_{j,1}, v_{j,2} \in \mathcal{P}(X)$, and in this case $F_j^{(t)}(X) \leq 0.4\Phi_j^{(t)} + \frac{1}{\mathsf{polylog}(k)}$;

Together, and using the inequality $1 - \mathbf{logit}_y(F^{(t)}, X) \leq \frac{\sum_{j\neq y} e^{F_j^{(t)}(X)}}{e^{F_y^{(t)}(X)}}$, and conclude that

$$\mathop{\mathbb{E}}_{(X,y)\sim\mathcal{Z}_m} \big[ \mathbb{1}_{i=y}(1 - \mathbf{logit}_y(F^{(t)}, X)) \big] \leq \frac{1}{k} \cdot O\left( \frac{se^{0.4\widehat{\Phi}^{(t)}} + k}{e^{\widehat{\Phi}^{(t)}}} \right)$$

Summing up over all $(r, \ell) \in \mathcal{M}_i^{(0)} \times [2]$ with $|\mathcal{M}_i^{(0)}| \leq m_0 \leq \widetilde{O}(1)$, we have

$$\widehat{\Phi}^{(t+1)} \leq \widehat{\Phi}^{(t)} + \eta \frac{1}{k} \widetilde{O} \left( \frac{se^{0.4\widehat{\Phi}^{(t)}} + k}{e^{\widehat{\Phi}^{(t)}}} + \frac{N_s}{N} \right)$$

This implies whenever $e^{\widehat{\Phi}^{(t)}} = \Omega\left( k^{2.5} \Upsilon^{2.5} \right)$, we have

$$\widehat{\Phi}^{(t+1)} \leq \widehat{\Phi}^{(t)} + \eta \widetilde{O}\Big( \frac{1}{k^{2.5}\Upsilon^{2.5}} \Big)$$

This finishes the proof of Claim C.18. $\square$

*Proof of Claim C.17.* We first prove Claim C.17a and the proof of Claim C.17b is only simpler.

Recall from (C.8) in the proof of Claim C.15 that as long as $t \geq T_{0,i}$ (so $\Lambda_i^{(t)} \geq \Lambda_\varnothing$),

$$\Lambda_i^{(t+1)} \geq \Lambda_i^{(t)} + \eta \mathop{\mathbb{E}}_{(X,y)\sim\mathcal{Z}_m} \Big[ 0.89 \cdot \mathbb{1}_{y=i} \Big( 1 - \mathbf{logit}_i(F^{(t)}, X) \Big) \Big]$$
$$- \eta \mathop{\mathbb{E}}_{(X,y)\sim\mathcal{Z}_m} \Big[ \mathbb{1}_{y\neq i}(\mathcal{E}_1 + \mathcal{E}_3 + 0.4\mathbb{1}_{v_{i,1} \text{ or } v_{i,2} \in \mathcal{P}(X)})\mathbf{logit}_i \Big( F^{(t)}, X \Big) \Big]$$
$$- O\Big(\frac{\eta N_s}{N}\Big) \mathop{\mathbb{E}}_{(X,y)\sim\mathcal{Z}_s} \Big[ \mathbb{1}_{y=i} \cdot \widetilde{O}(\sigma_p P)\Big( 1 - \mathbf{logit}_y \Big( F^{(t)}, X \Big) \Big) \Big]$$
$$- O\Big(\frac{\eta N_s}{kN}\Big) \mathop{\mathbb{E}}_{(X,y)\sim\mathcal{Z}_s} \Big[ \mathbb{1}_{y\neq i} \Big( \mathcal{E}_1 + \mathcal{E}_3 + \mathbb{1}_{v_{i,1} \text{ or } v_{i,2} \in \mathcal{P}(X)} \Big) \Big( 1 - \mathbf{logit}_y \Big( F^{(t)}, X \Big) \Big) \Big]$$

Applying Claim C.9, we have for $(X, y) \in \mathcal{Z}_m$ and $i \neq y$, if at least one of $\{v_{i,1}, v_{i,2}\}$ is in $\mathcal{P}(X)$, then $F_i^{(t)}(X) \leq 0.4\Phi_i^{(t)} + \frac{1}{\mathsf{polylog}(k)}$. Therefore,

$$\mathop{\mathbb{E}}_{(X,y)\sim \mathcal{Z}_m} \left[\mathbb{1}_{y\neq i}\mathbb{1}_{v_{i,1} \text{ or } v_{i,2}\in \mathcal{P}(X)}\mathbf{logit}_i\left(F^{(t)}, X\right)\right]$$

$$\leq \mathop{\mathbb{E}}_{(X,y)\sim \mathcal{Z}_m}\left[\mathbb{1}_{y\neq i}\mathbb{1}_{v_{i,1} \text{ or } v_{i,2}\in \mathcal{P}(X)} \frac{1}{1+\sum_{j\in[k]} e^{F_j^{(t)}(X)-0.4\Phi_i^{(t)}}}\right]$$

$$\leq \mathop{\mathbb{E}}_{(X,y)\sim \mathcal{Z}_m}\left[\mathbb{1}_{y\neq i}\mathbb{1}_{v_{i,1} \text{ or } v_{i,2}\in \mathcal{P}(X)}O(\Upsilon)\right] \leq O(\frac{s\Upsilon}{k}) \quad (C.11)$$

Above, the last inequality uses $F_j^{(t)}(X) \geq 0$ and Claim C.18 which says $e^{0.4\Phi_i^{(t)}} \leq O(k\Upsilon)$. Also, for obvious reason

$$\mathop{\mathbb{E}}_{(X,y)\sim \mathcal{Z}_s}\left[\left(k\mathbb{1}_{y=i}\cdot \widetilde{O}(\sigma_p P) + \mathbb{1}_{y\neq i}\left(\mathcal{E}_1 + \mathcal{E}_3 + \mathbb{1}_{\{v_{i,1},v_{i,2}\}\in\mathcal{P}(X)}\right)\right)\left(1 - \mathbf{logit}_y\left(F^{(t)}, X\right)\right)\right] \leq O(\frac{s}{k})$$

Together, we arrive at the conclusion that

$$\Lambda_i^{(t+1)} \geq \Lambda_i^{(t)} + \Omega(\eta)\left(\mathop{\mathbb{E}}_{(X,y)\sim \mathcal{Z}_m}\left[\mathbb{1}_{y=i}\left(1-\mathbf{logit}_i(F^{(t)}, X)\right)\right] - O\left(\frac{s\Upsilon}{k} + \frac{s}{k^2}\frac{N_s}{N}\right)\right)$$

Using $\Lambda_i^{(t)} \leq \widetilde{O}(1)$ from Induction Hypothesis B.3g we immediate finish the proof of Claim C.17a.

As for the first part of Claim C.17b, note that when $v_{i,1} \notin P(x)$ and $v_{i,2} \notin P(x)$, we have $\mathbf{logit}_i(F^{(t)}, X) \leq O(\frac{1}{k})$; but if one of them belongs to $P(X)$ the probability is $\frac{s}{k}$ and we have (C.11). They together (and using $s\Upsilon \leq 1$) imply the first part of Claim C.17b.

As for the second part of Claim C.17b, it is similar to the first part once we take into account $\mathbf{Pr}[v_{j,\ell} \in \mathcal{P}(X)] = \frac{s}{k}$. $\square$

## C.3 Tensor Power Method Bound

In this subsection we establish a lemma for comparing the growth speed of two sequences of updates of the form $x_{t+1} \leftarrow x_t + \eta C_t x_t^{q-1}$. This should be reminiscent of the classical analysis of the growth of eigenvalues on the (incremental) tensor power method of degree $q$.

**Lemma C.19.** *Let $q \geq 3$ be a constant and $x_0, y_0 = o(1)$. Let $\{x_t, y_t\}_{t\geq 0}$ be two positive sequences updated as*

- $x_{t+1} \geq x_t + \eta C_t x_t^{q-1}$ *for some $C_t = \Theta(1)$, and*
- $y_{t+1} \leq y_t + \eta S C_t y_t^{q-1}$ *for some constant $S = \Theta(1)$.*

*Suppose $x_0 \geq y_0 S^{\frac{1}{q-2}}(1 + \frac{1}{\mathsf{polylog}(k)})$, then we must have for every $A = O(1)$, let $T_x$ be the first iteration such that $x_t \geq A$, then*

$$y_{T_x} \leq O(y_0 \cdot \mathsf{polylog}(k))$$

We first establish a claim before proving Lemma C.19.

**Claim C.20.** *Consider an increasing sequence $x_t \geq 0$ defined as $x_{t+1} = x_t + \eta C_t x_t^{q-1}$ for some $C_t = \Theta(1)$, then we have for every $A > x_0$, every $\delta \in (0,1)$, and every $\eta \in (0,1)$:*

$$\sum_{t\geq 0, x_t \leq A} \eta C_t \geq \left[\frac{\delta(1+\delta)^{-1}}{(1+\delta)^{q-2}-1}\left(1 - \left(\frac{(1+\delta)x_0}{A}\right)^{q-2}\right) - \frac{O(\eta A^{q-1})}{x_0}\frac{\log\left(\frac{A}{x_0}\right)}{\log(1+\delta)}\right]\cdot \frac{1}{x_0^{q-2}}$$

$$\sum_{t\geq 0, x_t \leq A} \eta C_t \leq \left[\frac{(1+\delta)^{q-2}}{(q-2)} + \frac{O(\eta A^{q-1})}{x_0}\frac{\log\left(\frac{A}{x_0}\right)}{\log(1+\delta)}\right]\cdot \frac{1}{x_0^{q-2}}$$

*Proof of Claim C.20.* For every $g = 0, 1, 2, \ldots$, let $\mathcal{T}_g$ be the first iteration such that $x_t \geq (1+\delta)^g x_0$. Let $b$ be the smallest integer such that $(1 + \delta)^b x_0 \geq A$. Suppose for notation simplicity that we replace $x_t$ with exactly $A$ whenever $x_t \geq A$.

By the definition of $\mathcal{T}_g$, we have

$$\sum_{t \in [\mathcal{T}_g, \mathcal{T}_{g+1})} \eta C_t [(1+\delta)^g x_0]^{(q-1)} \leq x_{\mathcal{T}_{g+1}} - x_{\mathcal{T}_g} \leq \delta(1+\delta)^g x_0 + O(\eta A^{q-1})$$

$$\sum_{t \in [\mathcal{T}_g, \mathcal{T}_{g+1})} \eta C_t [(1+\delta)^{g+1} x_0]^{(q-1)} \geq x_{\mathcal{T}_{g+1}} - x_{\mathcal{T}_g} \geq \delta(1+\delta)^g x_0 - O(\eta A^{q-1})$$

These imply that

$$\sum_{t \in [\mathcal{T}_g, \mathcal{T}_{g+1})} \eta C_t \leq \frac{\delta}{(1+\delta)^{g(q-2)}} \frac{1}{x_0^{q-2}} + \frac{O(\eta A^{q-1})}{x_0^{q-1}}$$

$$\sum_{t \in [\mathcal{T}_g, \mathcal{T}_{g+1})} \eta C_t \geq \frac{\delta}{(1+\delta)^{g(q-2)}(1+\delta)^{q-1}} \frac{1}{x_0^{q-2}} - \frac{O(\eta A^{q-1})}{x_0^{q-1}}$$

Recall $b$ is the smallest integer such that $(1+\delta)^b x_0 \geq A$, so we can calculate

$$\sum_{t \geq 0, x_t \leq A} \eta C_t \leq \sum_{g=0}^{b-1} \frac{\delta}{(1+\delta)^{g(q-2)}} \frac{1}{x_0^{q-2}} + \frac{O(\eta A^{q-1})}{x_0^{q-1}} b = \frac{\delta}{1 - \frac{1}{(1+\delta)^{q-2}}} \frac{1}{x_0^{q-2}} + \frac{O(\eta A^{q-1})}{x_0^{q-1}} b$$

$$= \frac{\delta (1+\delta)^{q-2}}{(1+\delta)^{q-2} - 1} \frac{1}{x_0^{q-2}} + \frac{O(\eta A^{q-1})}{x_0^{q-1}} b \leq \frac{(1+\delta)^{q-2}}{(q-2)} \frac{1}{x_0^{q-2}} + \frac{O(\eta A^{q-1})}{x_0^{q-1}} b$$

$$\sum_{t \geq 0, x_t \leq A} \eta C_t \geq \sum_{q=0}^{b-2} \frac{\delta}{(1+\delta)^{g(q-2)}(1+\delta)^{q-1}} \frac{1}{x_0^{q-2}} - \frac{O(\eta A^{q-1})}{x_0^{q-1}} b$$

$$\geq \frac{\delta(1+\delta)^{-1} \left(1 - \frac{1}{(1+\delta)^{(q-2)(b-1)}}\right)}{(1+\delta)^{q-2} - 1} \frac{1}{x_0^{q-2}} - \frac{O(\eta A^{q-1})}{x_0^{q-1}} b$$

$$\geq \frac{\delta(1+\delta)^{-1} \left(1 - \left(\frac{(1+\delta)x_0}{A}\right)^{q-2}\right)}{(1+\delta)^{q-2} - 1} \frac{1}{x_0^{q-2}} - \frac{O(\eta A^{q-1})}{x_0^{q-1}} b \qquad \square$$

*Proof of Lemma C.19.* Let us apply Claim C.20 twice, once for the $x_t$ sequence with $C_t$ and threshold $A$, and the other time for the $y_t$ sequence with $C'_t = SC_t$ and threshold $A' = y_0 \cdot \mathsf{polylog}(k)$. Let $T_x$ be the first iteration $t$ in which $x_t \geq A$, and $T_y$ be the first iteration $t$ in which $y_t \geq A'$.

According to Claim C.20, we know

$$\sum_{t=0}^{T_x} \eta C_t \leq \left[\frac{(1+\delta)^{q-2}}{(q-2)} + \frac{O(\eta A^{q-1})}{x_0} \frac{\log\left(\frac{A}{x_0}\right)}{\log(1+\delta)}\right] \cdot \frac{1}{x_0^{q-2}}$$

$$\leq \frac{1 + O(\delta)}{(q-2)x_0^{q-2}} + O\left(\frac{\eta \log(1/x_0)}{\delta x_0^{q-1}}\right)$$

$$\sum_{t=0}^{T_y} \eta C'_t = \sum_{t=0}^{T_y} \eta SC_t \geq \left[\frac{\delta(1+\delta)^{-1}}{(1+\delta)^{q-2} - 1} \left(1 - \left(\frac{(1+\delta)y_0}{A'}\right)^{q-2}\right) - \frac{O(\eta(A')^{q-1})}{y_0} \frac{\log\left(\frac{A'}{y_0}\right)}{\log(1+\delta)}\right] \cdot \frac{1}{y_0^{q-2}}$$

$$\geq \frac{1 - O(\delta + \frac{1}{\mathsf{polylog}(k)})}{(q-2)y_0^{q-2}} - \tilde{O}\left(\frac{\eta}{\delta}\right)$$

Therefore, choosing $\delta = \frac{1}{\mathsf{polylog}(k)}$ and $\eta \leq \frac{x_0}{\log(1/x_0)\mathsf{polylog}(k)}$, together with the assumption $x_0 \geq y_0 S^{\frac{1}{q-2}}(1 + \frac{1}{\mathsf{polylog}(k)})$, we immediately have that $T_x < T_y$ so we finish the proof. $\square$

## C.4 Main Lemmas for Proving the Induction Hypothesis

In this subsection, we begin to provide key technical lemmas that, when combined together, shall prove Induction Hypothesis B.3. (We combine the analysis in the next subsection, Section C.5.)

### C.4.1 Lambda Lemma

Recall $\Lambda_i^{(t)} \stackrel{\text{def}}{=} \max_{r \in [m], \ell \in [2]} [\langle w_{i,r}^{(t)}, v_{i,\ell} \rangle]^+$. Our first lemma shows that $\Lambda_i^{(t)}$ cannot go above $\widetilde{O}(1)$.

**Lemma C.21.** *Suppose Induction Hypothesis B.3 holds for all iterations $< t$ and suppose $N_s \leq \widetilde{o}(k/\rho)$. Then, letting $\Phi_{i,\ell}^{(t)} \stackrel{\text{def}}{=} \sum_{r \in [m]} [\langle w_{i,r}^{(t)}, v_{i,\ell} \rangle]^+$, we have*

$$\forall i \in [k], \forall \ell \in [2]: \qquad \Phi_{i,\ell}^{(t)} \leq \widetilde{O}(1)$$

*This implies $\Lambda_i^{(t)} \leq \widetilde{O}(1)$ as well.*

*Proof of Lemma C.21.* We make a simple observation:

- For those $r \in [m] \setminus \mathcal{M}_i^{(0)}$ and $\ell \in [2]$, recall Induction Hypothesis B.3i says $[\langle w_{i,r}^{(t)}, v_{i,\ell} \rangle]^+ \leq \widetilde{O}(\sigma_0)$ and their summation does not exceed $\frac{1}{\mathsf{polylog}(k)}$ due to our choice of $m$.

Therefore, to prove this claim, it suffices to slightly abuse the notation and prove $\overline{\Phi}_{i,\ell}^{(t)} \leq \widetilde{O}(1)$ for

$$\overline{\Phi}_{i,\ell}^{(t)} \stackrel{\text{def}}{=} \sum_{r \in \mathcal{M}_i^{(0)}} [\langle w_{i,r}^{(t)}, v_{i,\ell} \rangle]^+ = \Phi_{i,\ell}^{(t)} \pm \frac{1}{\mathsf{poly}(k)}$$

Recall gradient descent update gives

$$\langle w_{i,r}^{(t+1)}, v_{i,\ell} \rangle = \langle w_{i,r}^{(t)}, v_{i,\ell} \rangle + \eta \mathop{\mathbb{E}}_{(X,y) \sim \mathcal{Z}} \left[ \langle -\nabla_{w_{i,r}} L(F^{(t)}; X, y), v_{i,\ell} \rangle \right]$$

when taking the positive part, we know there exists $\Delta_{i,r,\ell}^{(t)} \in [0, 1]$ such that

$$[\langle w_{i,r}^{(t+1)}, v_{i,\ell} \rangle]^+ = [\langle w_{i,r}^{(t)}, v_{i,\ell} \rangle]^+ + \eta \Delta_{i,r,\ell}^{(t)} \mathop{\mathbb{E}}_{(X,y) \sim \mathcal{Z}} \left[ \langle -\nabla_{w_{i,r}} L(F^{(t)}; X, y), v_{i,\ell} \rangle \right]$$

Now in each iteration $t$, for every pair $(i, \ell)$, we define a special subset $\mathcal{Z}_{s,i,\ell}$ of the single-view data as

$$\mathcal{Z}_{s,i,\ell} \stackrel{\text{def}}{=} \left\{ (X, y) \in \mathcal{Z}_s \mid y = i \wedge \widehat{\ell}(X) = 3 - \ell \right\}$$

For analysis purpose, from iteration $t = 0$ onwards, we define two sequences of quantities

- define $A_{i,\ell}^{(0)} \stackrel{\text{def}}{=} \sum_{r \in \mathcal{M}_i^{(0)}} [\langle w_{i,r}^{(0)}, v_{i,\ell} \rangle]^+$ and $B_{i,\ell}^{(0)} = 0$.
- when $t \geq 0$, define

$$A_{i,\ell}^{(t+1)} \stackrel{\text{def}}{=} A_{i,\ell}^{(t)} + \eta \sum_{r \in \mathcal{M}_i^{(0)}} \Delta_{i,r,\ell}^{(t)} \mathop{\mathbb{E}}_{(X,y) \sim \mathcal{Z}} \left[ \mathbb{1}_{(X,y) \notin \mathcal{Z}_{s,i,\ell}} \cdot \langle -\nabla_{w_{i,r}} L(F^{(t)}; X, y), v_{i,\ell} \rangle \right]$$

$$B_{i,\ell}^{(t+1)} \stackrel{\text{def}}{=} B_{i,\ell}^{(t)} + \eta \sum_{r \in \mathcal{M}_i^{(0)}} \Delta_{i,r,\ell}^{(t)} \mathop{\mathbb{E}}_{(X,y) \sim \mathcal{Z}} \left[ \mathbb{1}_{(X,y) \in \mathcal{Z}_{s,i,\ell}} \cdot \langle -\nabla_{w_{i,r}} L(F^{(t)}; X, y), v_{i,\ell} \rangle \right]$$

For obvious reason, we have $\overline{\Phi}_{i,\ell}^{(t)} = A_{i,\ell}^{(t)} + B_{i,\ell}^{(t)}$.

**Bound the $B$ sequence.** Applying Claim C.7, we have

$$B_{i,\ell}^{(t+1)} = B_{i,\ell}^{(t)} + \eta \sum_{r \in \mathcal{M}_i^{(0)}} \Delta_{i,r,\ell}^{(t)} \mathop{\mathbb{E}}_{(X,y) \sim \mathcal{Z}} \left[ \mathbb{1}_{(X,y) \in \mathcal{Z}_{s,i,\ell}} \cdot (V_{i,r,\ell}(X) \pm (\mathcal{E}_1 + \mathcal{E}_3)) \left( 1 - \mathbf{logit}_i(F^{(t)}, X) \right) \right]$$

but according to the definition of $\mathcal{Z}_{s,i,\ell}$ (which implies $\widehat{\ell}(X) = 3 - \ell$), we must have

$$0 \leq V_{i,r,\ell}(X) = \sum_{p \in \mathcal{P}_{v_{i,\ell}}(X)} \widetilde{\mathsf{ReLU}}'(\langle w_{i,r}, x_p \rangle) z_p \leq O(\rho) \cdot \sum_{p \in \mathcal{P}_{v_{i,\ell}}(X)} \widetilde{\mathsf{ReLU}}'(\langle w_{i,r}, x_p \rangle)$$

This means, we must have

$$|B_{i,\ell}^{(t+1)} - B_{i,\ell}^{(t)}|$$
$$\leq O\Big(\frac{\eta \rho N_s}{N}\Big) \sum_{r \in \mathcal{M}_i^{(0)}} \mathbb{E}_{(X,y) \sim \mathcal{Z}_s} \Big[ \mathbb{1}_{(X,y) \in \mathcal{Z}_{s,i,\ell}} \big(1 - \mathbf{logit}_i(F^{(t)}, X)\big) \Big(\mathcal{E}_1 + \mathcal{E}_3 + \sum_{p \in \mathcal{P}_{v_{i,\ell}}(X)} \widetilde{\mathsf{ReLU}}'(\langle w_{i,r}, x_p \rangle)\Big) \Big]$$

Applying Claim C.12a (and noticing $T_0 \leq O(\frac{N}{\eta})$), we conclude that (using our parameter assumption)

$$\forall t \geq 0: \quad |B_{i,\ell}^{(t)}| \leq \widetilde{O}\Big(\frac{\rho N_s}{k}\Big) < \frac{1}{\mathsf{polylog}(k)}$$

**Bound the $A$ sequence.** So far we have derived that $\Phi_{i,\ell}^{(t)} = A_{i,\ell}^{(t)} \pm \frac{1}{\mathsf{polylog}(k)}$ (because $|B_{i,\ell}^{(t)}|, |\Phi_{i,\ell}^{(t)} - \overline{\Phi}_{i,\ell}^{(t)}| \leq \frac{1}{\mathsf{polylog}(k)}$. So, it suffices to bound the $A$ sequence. Let us denote by

$$\Phi^{(t)} \stackrel{\text{def}}{=} \max_{i \in [k], \ell \in [2]} \Phi_{i,\ell}^{(t)}$$

in the remainder of the proof. Suppose we are now at some iteration $t \geq T_0$, let

$$(i, \ell) = \arg\max_{i \in [k], \ell \in [2]} \{A_{i,\ell}^{(t)}\}$$

Applying Claim C.7 and Claim C.8, and using $V_{i,r,\ell}(X) \in [0,1]$, we have

$$A_{i,\ell}^{(t+1)} \leq A_{i,\ell}^{(t)} + O(\eta)\Big(\mathbb{E}_{(X,y) \sim \mathcal{Z}}\big[\mathbb{1}_{y=i}\mathbb{1}_{(X,y) \notin \mathcal{Z}_{s,i,\ell}}(1 - \mathbf{logit}_y(F^{(t)}, X))\big] + \widetilde{O}(\sigma_p P)\Big)$$

Observe that, whenever $A_{i,\ell}^{(t)} > \mathsf{polylog}(k)$, we also have $\Phi^{(t)} \geq \mathsf{polylog}(k)$ and thus

- for every $(X, y) \in \mathcal{Z}_m$ with $y = i$, recall from Claim C.9 that

$$F_j^{(t)}(X) = \sum_{\ell \in [2]} \Big(\Phi_{j,\ell}^{(t)} \times \mathbb{1}_{v_{j,\ell} \in \mathcal{V}(X)} \Big(\sum_{p \in \mathcal{P}_{v_{j,\ell}}(X)} z_p\Big)\Big) \pm O\Big(\frac{1}{\mathsf{polylog}(k)}\Big)$$

By our choice of the distribution, this implies

  - $F_j^{(t)}(X) \leq 0.8001 \Phi^{(t)}$ for $j \neq i$, and
  - $F_i^{(t)}(X) \geq 0.9999 \Phi^{(t)}$ because $(i, \ell)$ is the argmax of $A_{i,\ell}^{(t)}$ and $A_{i,\ell}^{(t)}$ is close to $\Phi_{i,\ell}^{(t)}$.

- for every $(X, y) \in \mathcal{Z}_s$ with $y = i$ and $\widehat{\ell}(X) = \ell$, we can also use Claim C.9 to derive

  - $F_j^{(t)}(X) \leq 0.8001 \Phi^{(t)}$ for $j \neq i$.

  However, on the $i$-th output, for every $p \in \mathcal{P}_{v_{i,\ell}}(X)$, using $\langle w_{i,r}^{(t)}, x_p \rangle = \langle w_{i,r}^{(t)}, v_{i,\ell} \rangle z_p + \langle w_{i,r}^{(t)}, \xi_p \rangle \pm \widetilde{O}(\sigma_0 \gamma k)$ from Induction Hypothesis B.3d and using $\langle w_{i,r}^{(t)}, \xi_p \rangle \geq -\frac{1}{\mathsf{polylog}(k)}$ from Claim C.13a, and $\sum_{p \in \mathcal{P}_{v_{i,\ell}}(X)} z_p \geq 1$ from Def. 3.1, we also have

  - $F_i^{(t)}(X) \geq \Phi_{i,\ell}^{(t)} - 0.0001 \geq 0.9999 \Phi^{(t)}$.

In both cases, we have $1 - \mathbf{logit}_i\big(F^{(t)}, X\big) = e^{-\Omega(\log^5 k)}$ which is negligible. Therefore, we derived

that

whenever $\max_{i\in[k],\ell\in[2]}\{A_{i,\ell}^{(t)}\} \geq \mathsf{polylog}(k)$ $\max_{i\in[k],\ell\in[2]}\{A_{i,\ell}^{(t+1)}\} \leq \max_{i\in[k],\ell\in[2]}\{A_{i,\ell}^{(t)}\}+\widetilde{O}(\eta\cdot e^{-\Omega(\log^5 k)}+\eta\sigma_p P)$ .

This finishes the proof that all $A_{i,\ell}^{(t)} \leq \widetilde{O}(1)$. □

### C.4.2 Off-Diagonal Correlations are Small

Our previous subsection upper bounds the "diagonal" correlations $\langle w_{i,r}^{(t)}, v_{i,\ell}\rangle$, and in this subsection we bound the "off-diagonal" correlations $\langle w_{i,r}^{(t)}, v_{j,\ell}\rangle$ for $i \neq j$.

**Lemma C.22.** *Suppose Parameter C.1 holds and suppose Induction Hypothesis B.3 holds for all iterations $< t$. Then,*

$$\forall i \in [k] \ , \ \forall r \in [m] \ , \ \forall j \in [k] \setminus \{i\}: \quad |\langle w_{i,r}^{(t)}, v_{j,\ell}\rangle| \leq \widetilde{O}(\sigma_0)$$

*Proof of Lemma C.22.* Let us denote by $R_i^{(t)} \stackrel{\text{def}}{=} \max_{r\in[m], j\in[k]\setminus\{i\}} |\langle w_{i,r}^{(t)}, v_{j,\ell}\rangle|$.

By Claim C.7 and Claim C.8,

$$|\langle w_{i,r}^{(t+1)}, v_{j,\ell}\rangle| \leq |\langle w_{i,r}^{(t)}, v_{j,\ell}\rangle| + \eta \mathop{\mathbb{E}}_{(X,y)\sim\mathcal{Z}} \Big[ \mathbb{1}_{y=i}\Big(\mathcal{E}_{2,i,r}(X) + \mathcal{E}_1 + \mathcal{E}_3 + \mathcal{E}_{4,j,\ell}(X)\Big)\Big(1 - \mathbf{logit}_i\Big(F^{(t)}, X\Big)\Big)$$
$$+ \mathbb{1}_{y\neq i}\Big(\mathcal{E}_1 + \mathcal{E}_3 + \mathcal{E}_{4,j,\ell}(X)\Big)\mathbf{logit}_i\Big(F^{(t)}, X\Big)\Big]$$

**Stage 1.** In the first stage, namely when $t \leq T_{0,i}$, we have $\mathbf{logit}_i(F^{(t)}, X) \leq O(\frac{1}{k})$ (see Claim C.4), have $\mathcal{E}_{2,i,r}(X) \leq \widetilde{O}(\gamma(\Lambda_i^{(t)})^{q-1})$, and have $\mathcal{E}_{4,i,r}(X) \leq \widetilde{O}(\sigma_0)^{q-1}\mathbb{1}_{v_{j,\ell}\in\mathcal{V}(X)}$, so

$$|\langle w_{i,r}^{(t+1)}, v_{j,\ell}\rangle| \leq |\langle w_{i,r}^{(t)}, v_{j,\ell}\rangle| + \widetilde{O}\Big(\frac{\eta}{k}\Big)\Big(\gamma(\Lambda_i^{(t)})^{q-1} + (\sigma_0^{q-1})\gamma s + \widetilde{O}\left((\sigma_0\gamma k)^{q-1}\right)\gamma P + (\sigma_0)^{q-1}\frac{s}{k}\Big) \quad \text{(C.12)}$$

Recall in the first stage $\Lambda_i^{(t+1)} = \Lambda_i^{(t)} + \Theta\left(\frac{\eta}{k}\right)\widetilde{\mathsf{ReLU}}'(\Lambda_i^{(t)})$ (see Claim C.10) and therefore

$$\sum_{t\leq T_{0,i}} \eta(\Lambda_i^{(t)})^{q-1} \leq \widetilde{O}(k)$$

Hence, together with $T_{0,i} \leq T_0 = \Theta\big(\frac{k}{\eta\sigma_0^{q-2}}\big)$ (see Claim C.11), as long as

$$\gamma = \widetilde{O}(\sigma_0), \quad \gamma = \widetilde{O}(1/s), \quad {\color{blue}(\gamma k)^{q-1}\gamma P = \widetilde{O}(1)} \quad \text{(C.13)}$$

(Note that all of them are satisfied by Parameter C.1). Plugging them back to (C.12), we have we have

$$R_i^{(t)} \leq R_i^{(0)} + \widetilde{O}(\sigma_0) + \widetilde{O}\Big(\frac{\eta}{k}T_0\Big)\Big((\sigma_0^{q-1})\gamma s + \widetilde{O}\left((\sigma_0\gamma k)^{q-1}\right)\gamma P + (\sigma_0)^{q-1}\frac{s}{k}\Big) \leq \widetilde{O}(\sigma_0)$$

for every $t \leq T_{0,i}$.

**Stage 2.** In the second stage, namely when $t \in [T_{0,i}, T_0]$, we have the naive upper bounds $\mathcal{E}_{2,i,r}(X) \leq O(\gamma)$ and $\mathcal{E}_{4,i,r}(X) \leq \widetilde{O}(\sigma_0)^{q-1}\mathbb{1}_{v_{j,\ell}\in\mathcal{V}(X)}$, so

$$|\langle w_{i,r}^{(t+1)}, v_{j,\ell}\rangle| \leq |\langle w_{i,r}^{(t)}, v_{j,\ell}\rangle| + \eta \mathop{\mathbb{E}}_{(X,y)\sim\mathcal{Z}} \Big[ \mathbb{1}_{y=i}\Big(O(\gamma) + \mathcal{E}_1 + \mathcal{E}_3 + \widetilde{O}(\sigma_0)^{q-1}\mathbb{1}_{v_{j,\ell}\in\mathcal{V}(X)}\Big)\Big(1 - \mathbf{logit}_i\Big(F^{(t)}, X\Big)\Big)$$
$$+ \mathbb{1}_{y\neq i}\Big(\mathcal{E}_1 + \mathcal{E}_3 + \widetilde{O}(\sigma_0)^{q-1}\mathbb{1}_{v_{j,\ell}\in\mathcal{V}(X)}\Big)\mathbf{logit}_i\Big(F^{(t)}, X\Big)\Big]$$

Recall that for every $(X, y) \in \mathcal{Z}_s$ and $i \neq y$, it satisfies $\mathbf{logit}_i(F^{(t)}, X) = O\left(\frac{1}{k}\right)\left(1 - \mathbf{logit}_y(F^{(t)}, X)\right)$ (see Claim C.4). Therefore,

$$|\langle w_{i,r}^{(t+1)}, v_{j,\ell}\rangle| \leq |\langle w_{i,r}^{(t)}, v_{j,\ell}\rangle| + \eta \mathop{\mathbb{E}}_{(X,y) \sim \mathcal{Z}_m} \left[\mathbb{1}_{y=i}\left(O(\gamma) + \mathcal{E}_1 + \mathcal{E}_3 + \widetilde{O}(\sigma_0)^{q-1}\mathbb{1}_{v_{j,\ell} \in \mathcal{V}(X)}\right)\left(1 - \mathbf{logit}_y\left(F^{(t)}, X\right)\right)\right.$$
$$\left. + \mathbb{1}_{y \neq i}\left(\mathcal{E}_1 + \mathcal{E}_3 + \widetilde{O}(\sigma_0)^{q-1}\mathbb{1}_{v_{j,\ell} \in \mathcal{V}(X)}\right)\mathbf{logit}_i\left(F^{(t)}, X\right)\right]$$
$$+ O\left(\frac{\eta N_s}{kN}\right) \mathop{\mathbb{E}}_{(X,y) \sim \mathcal{Z}_s} \left[k\mathbb{1}_{y=i}\left(O(\gamma) + \mathcal{E}_1 + \mathcal{E}_3 + \widetilde{O}(\sigma_0)^{q-1}\mathbb{1}_{v_{j,\ell} \in \mathcal{V}(X)}\right)\left(1 - \mathbf{logit}_y\left(F^{(t)}, X\right)\right)\right.$$
$$\left. + \mathbb{1}_{y \neq i}\left(\mathcal{E}_1 + \mathcal{E}_3 + \widetilde{O}(\sigma_0)^{q-1}\mathbb{1}_{v_{j,\ell} \in \mathcal{V}(X)}\right)\left(1 - \mathbf{logit}_y\left(F^{(t)}, X\right)\right)\right]$$
(C.14)

Next, one can naively verify

$$\mathop{\mathbb{E}}_{(X,y) \sim \mathcal{Z}_s}\left[\mathbb{1}_{y=i}\mathbb{1}_{v_{j,\ell} \in \mathcal{V}(X)}\right] \leq \widetilde{O}\left(\frac{s}{k^2}\right) \quad \text{and} \quad \mathop{\mathbb{E}}_{(X,y) \sim \mathcal{Z}_s}\left[\mathbb{1}_{y \neq i}\mathbb{1}_{v_{j,\ell} \in \mathcal{V}(X)}\right] \leq \widetilde{O}\left(\frac{s}{k}\right)$$

Furthermore, by Claim C.17, we have

$$\forall t \in [T_{0,i}, T_0]: \quad \mathop{\mathbb{E}}_{(X,y) \sim \mathcal{Z}_m}\left[\mathbb{1}_{y \neq i}\mathbf{logit}_i\left(F^{(t)}, X\right)\right] \leq O\left(\frac{1}{k}\right)$$

$$\forall t \in [T_{0,i}, T_0]: \quad \mathop{\mathbb{E}}_{(X,y) \sim \mathcal{Z}_m}\left[\mathbb{1}_{y \neq i}\mathbb{1}_{v_{j,\ell} \in \mathcal{P}(X)}\mathbf{logit}_i\left(F^{(t)}, X\right)\right] \leq O\left(\frac{s}{k^2}\right)$$

$$\sum_{t=T_{0,i}}^{T_0} \mathop{\mathbb{E}}_{(X,y) \sim \mathcal{Z}_m}\left[\mathbb{1}_{y=i}\left(1 - \mathbf{logit}_y\left(F^{(t)}, X\right)\right)\right] \leq O\left(\frac{s}{k}T_0\Upsilon\right) + \widetilde{O}\left(\frac{1}{\eta}\right)$$

Putting these back to (C.14), we immediately conclude (using $N_s \ll N$) that

$$R_i^{(t)} \leq R_i^{(T_{0,i})} + \widetilde{O}\left(\eta T_0 \cdot \frac{s}{k^2}\sigma_0^{q-1}\right)$$
$$+ \eta\left(O\left(\frac{s}{k}T_0\Upsilon\right) + \widetilde{O}\left(\frac{1}{\eta}\right)\right)\widetilde{O}\left(\gamma + (\sigma_0^{q-1})\gamma s + \widetilde{O}\left((\sigma_0\gamma k)^{q-1}\right)\gamma P + (\sigma_0)^{q-1}\frac{s}{k}\right)$$

Hence, recalling that $T_0 = \Theta\left(\frac{k}{\eta\sigma_0^{q-2}}\right)$ (see Claim C.11), as long as (C.13) together with $s \leq k$ and[14]

$$\frac{s}{k}\frac{k\Upsilon\gamma}{\sigma_0^{q-2}} = \widetilde{O}(\sigma_0) \; , \quad s\Upsilon \leq O(1) \tag{C.15}$$

Then, we also have $R_i^{(t)} \leq \widetilde{O}(\sigma_0)$ for every $t \in [T_{0,i}, T_i]$.

**Stage 3.** At the third stage, namely when $t \geq T_0$, we can continue from (C.14) (and apply the naive bound $\mathbb{1}_{y=i} \leq 1$ and $\mathbb{1}_{v_{j,\ell} \in \mathcal{V}(X)} \leq 1$) to derive that

$$|\langle w_{i,r}^{(t+1)}, v_{j,\ell}\rangle| \leq |\langle w_{i,r}^{(t)}, v_{j,\ell}\rangle| + \eta \mathop{\mathbb{E}}_{(X,y) \sim \mathcal{Z}_m}\left[\left(O(\gamma) + \mathcal{E}_1 + \mathcal{E}_3 + \widetilde{O}(\sigma_0)^{q-1}\right)\left(1 - \mathbf{logit}_y\left(F^{(t)}, X\right)\right)\right]$$
$$+ O\left(\frac{\eta N_s}{kN}\right) \mathop{\mathbb{E}}_{(X,y) \sim \mathcal{Z}_s}\left[k\mathbb{1}_{y=i}\left(O(\gamma) + \mathcal{E}_1 + \mathcal{E}_3 + \widetilde{O}(\sigma_0)^{q-1}\right)\left(1 - \mathbf{logit}_y\left(F^{(t)}, X\right)\right)\right.$$
$$\left. + \mathbb{1}_{y \neq i}\left(\mathcal{E}_1 + \mathcal{E}_3 + \widetilde{O}(\sigma_0)^{q-1}\right)\left(1 - \mathbf{logit}_y\left(F^{(t)}, X\right)\right)\right]$$
(C.16)

---
[14] Note that we have required these parameter choices in Parameter C.1. This is the tightest place for our required upper bounds on $\gamma$ and on $s$.

If we denote by
$$S^{(t)} \stackrel{\text{def}}{=} \mathbb{E}_{(X,y)\sim \mathcal{Z}_m}\left[1 - \textbf{logit}_y\left(F^{(t)}, X\right)\right]$$
$$G_i^{(t)} \stackrel{\text{def}}{=} \mathbb{E}_{(X,y)\sim \mathcal{Z}_s}\left[\mathbb{1}_{i=y} \cdot \left(1 - \textbf{logit}_y\left(F^{(t)}, X\right)\right)\right]$$
then we can simplify (C.16) into
$$R_i^{(t+1)} \leq R_i^{(t)} + \eta \left(S^{(t)} + \frac{N_s}{N}G_i^{(t)} + \frac{N_s}{kN}\sum_{i'\in[k]}G_{i'}^{(t)}\right)\widetilde{O}\left(\gamma + \sigma_0^{q-1}\gamma s + (\sigma_0 \gamma k)^{q-1}\gamma P + (\sigma_0)^{q-1}\right) . \tag{C.17}$$

Recall
$$\text{Claim C.14} \implies \sum_{t \geq T_0} S^{(t)} \leq \widetilde{O}\left(\frac{k}{\eta}\right)$$
$$\text{Claim C.12} \implies \forall i' \in [k]\colon \sum_{t \geq T_0} G_i^{(t)} \leq \widetilde{O}\left(\frac{N}{\eta k \rho^{q-1}}\right)$$

Putting them back to (C.17), we know that as long as[15]
$$k\gamma = \widetilde{O}(\sigma_0), \quad \gamma = \widetilde{O}(\sigma_0^{q-1}), \quad \frac{N_s}{k\rho^{q-1}} \leq \widetilde{O}\left(\frac{1}{\sigma_0^{q-2}}\right) \tag{C.18}$$
it satisfies $R_i^{(t)} \leq \widetilde{O}(\sigma_0)$ for all $t \geq T_0$. $\square$

### C.4.3 View Lottery Winning

Recall $\Lambda_{i,\ell}^{(t)} \stackrel{\text{def}}{=} \max_{r \in [m]}[\langle w_{i,r}^{(t)}, v_{i,\ell}\rangle]^+$. Also recall
$$\mathcal{M} \stackrel{\text{def}}{=} \left\{(i, \ell^*) \in [k] \times [2] \,\bigg|\, \Lambda_{i,\ell^*}^{(0)} \geq \Lambda_{i,3-\ell^*}^{(0)} \left(\frac{S_{i,3-\ell}}{S_{i,\ell}}\right)^{\frac{1}{q-2}} + \frac{1}{\text{polylog}(k)}\right\}$$
in which $S_{i,\ell} = \mathbb{E}_{(X,y)\sim \mathcal{Z}_m}\left[\mathbb{1}_{y=i}\sum_{p \in P_{v_{i,\ell}}(X)} z_p^q\right]$.

Our next lemma shows that when $(i, \ell^* \in \mathcal{M}$, we have $\Lambda_{i,3-\ell^*}^{(t)} \leq \widetilde{O}(\sigma_0)$. (In other words, view $\ell$ wins the lottery and view $3 - \ell$ is negligible, on label $i$.)

**Lemma C.23.** *Suppose Parameter C.1 holds and suppose Induction Hypothesis B.3 holds for all iterations $< t$. Then,*
$$\forall i, \ell^* \in \mathcal{M}: \quad \Lambda_{i,3-\ell^*}^{(t)} = \max_{r \in [m]}[\langle w_{i,r}^{(t)}, v_{i,3-\ell^*}\rangle]^+ \leq \widetilde{O}(\sigma_0)$$

*Proof of Lemma C.23.* By Claim C.7 and Claim C.8,
$$\langle w_{i,r}^{(t+1)}, v_{i,\ell}\rangle = \langle w_{i,r}^{(t)}, v_{i,\ell}\rangle + \eta \mathop{\mathbb{E}}_{(X,y)\sim \mathcal{Z}}\left[\mathbb{1}_{y=i}\left(V_{i,r,\ell}(X) \pm O(\mathcal{E}_1 + \mathcal{E}_3)\right)\left(1 - \textbf{logit}_i(F^{(t)}, X)\right)\right.$$
$$\left. \pm O(1) \cdot \mathbb{1}_{y \neq i}\left(\mathcal{E}_1 + \mathcal{E}_3 + \mathbb{1}_{v_{i,\ell} \in \mathcal{P}(X)}V_{i,r,\ell}(X)\right)\textbf{logit}_i\left(F^{(t)}, X\right)\right] \tag{C.19}$$
where
$$V_{i,r,\ell}(X) = \sum_{p \in \mathcal{P}_{v_{i,\ell}}(X)} \widetilde{\textsf{ReLU}}'(\langle w_{i,r}, x_p\rangle)z_p$$

---
[15] Note that we have required these parameter choices in Parameter C.1. This is the tightest place for our required upper bounds on $\gamma$ and on $N_s$.

**Stage 1.** In the first stage, namely when $t \leq T_{0,i}$, we have $\Lambda_i^{(t)} \leq \Lambda_\varnothing^- \ll \widetilde{O}(\frac{1}{m_0})$ (see Claim C.11). This implies $\textbf{logit}_i\left(F^{(t)}, X\right) \leq O(\frac{1}{k})$ from Claim C.4. Also, we have $\textbf{Pr}[v_{i,\ell} \in \mathcal{P}(X) \mid y \neq i] = \frac{s}{k}$. Thus, we can simplify (C.19) as and thus

$$\langle w_{i,r}^{(t+1)}, v_{i,\ell}\rangle = \langle w_{i,r}^{(t)}, v_{i,\ell}\rangle + \eta \mathop{\mathbb{E}}_{(X,y)\sim\mathcal{Z}} \left[\mathbb{1}_{i=y} V_{i,r,\ell}(X)(1-O(\frac{1}{k})) \pm O(\frac{1}{k})\mathbb{1}_{i\neq y}\mathbb{1}_{v_{i,\ell}\in\mathcal{P}(X)}V_{i,r,\ell}(X) \pm O(\frac{\mathcal{E}_1 + \mathcal{E}_3}{k})\right]$$

Since $N_s \ll N$, we can ignore single-view data and only focus on $(X, y) \in \mathcal{Z}_m$. Using Induction Hypothesis B.3, we know for $(X, y) \in \mathcal{Z}_m$,

$$V_{i,r,\ell}(X) = \sum_{p\in\mathcal{P}_{v_{i,\ell}}(X)} \widetilde{\textsf{ReLU}}'\left(\langle w_{i,r}^{(t)}, v_{i,\ell}\rangle z_p \pm \widetilde{o}(\sigma_0)\right) z_p$$

and furthermore since we are in stage 1, it satisfies $\langle w_{i,r}^{(t)}, v_{i,\ell}\rangle z_p \leq O(\Lambda_\varnothing^-) \ll \varrho$ (see Claim C.11) so

$$V_{i,r,\ell}(X) = \frac{1}{\varrho^{q-1}}([\langle w_{i,r}^{(t)}, v_{i,\ell}\rangle]^+)^{q-1} \sum_{p\in\mathcal{P}_{v_{i,\ell}}(X)} z_p^q \pm \widetilde{O}(\sigma_0)$$

Therefore,

$$\langle w_{i,r}^{(t+1)}, v_{i,\ell}\rangle = \langle w_{i,r}^{(t)}, v_{i,\ell}\rangle + \frac{\eta}{\varrho^{q-1}}([\langle w_{i,r}^{(t)}, v_{i,\ell}\rangle]^+)^{q-1}\left((1-O(\frac{1}{k}))\mathop{\mathbb{E}}_{(X,y)\sim\mathcal{Z}_m}\left[\mathbb{1}_{i=y}\sum_{p\in\mathcal{P}_{v_{i,\ell}}(X)} z_p^q\right] \pm O(\frac{s}{k^2})\right)$$

$$\pm O(\frac{\mathcal{E}_1 + \mathcal{E}_3}{k}) \cdot \eta$$

$$= \langle w_{i,r}^{(t)}, v_{i,\ell}\rangle + \frac{\eta}{\varrho^{q-1}}([\langle w_{i,r}^{(t)}, v_{i,\ell}\rangle]^+)^{q-1}\left(1 - O\left(\frac{1}{\textsf{polylog}(k)}\right)\right)S_{i,\ell} \pm \widetilde{o}\left(\frac{\sigma_0}{k}\eta\right) \quad (\text{C.20})$$

This means, if we take $r^* = \arg\max_{r\in[m]}\left\{\langle w_{i,r}^{(0)}, v_{i,\ell^*}\rangle\right\}$ and an arbitrary $r \in [m]$, we can define

- $x_t = \langle w_{i,r^*}^{(t)}, v_{i,\ell^*}\rangle \cdot (S_{i,\ell^*}/\varrho^{q-1})^{\frac{1}{q-2}}$
- $y_t = \max\left\{\langle w_{i,r}^{(t)}, v_{i,3-\ell^*}\rangle, \sigma_0\right\} \cdot (S_{i,3-\ell^*}/\varrho^{q-1})^{\frac{1}{q-2}}$

Then by (C.20) we know

$$x_{t+1} \geq x_t + \eta C_t x_t^{q-1} \quad \text{and} \quad y_{t+1} \leq y_t + \eta S C_t y_t^{q-1}$$

for some $C_t = 1 - O(\frac{1}{\textsf{polylog}(k)}) \in [0.9, 1]$ (where the value in the big $O$ notion can vary per iteration) and constant $S = \frac{S_{i,3-\ell^*}}{S_{i,\ell^*}}\left(1 + \frac{1}{\textsf{polylog}(k)}\right)$ that does not depend on $t$.

Now, since $\Lambda_{i,\ell^*}^{(0)} \geq \Lambda_{i,3-\ell^*}^{(0)}\left(\frac{S_{i,3-\ell^*}}{S_{i,\ell^*}}\right)^{\frac{1}{q-2}} + \frac{1}{\textsf{polylog}(k)}$ implies $x_0 \geq y_0 S^{\frac{1}{q-2}}(1 + \frac{1}{\textsf{polylog}(k)})$ and we can apply Lemma C.19 to derive that

- when $\langle w_{i,r^*}^{(t)}, v_{i,\ell^*}\rangle$ reaches $\widetilde{\Omega}(1)$, which necessarily is an iteration $t \geq T_{0,i}$, we still have that

$$y_t \leq \widetilde{O}(y_0) \implies \langle w_{i,r}^{(t)}, v_{i,3-\ell^*}\rangle \leq \widetilde{O}(\sigma_0)$$

(This uses $|\langle w_{i,r}^{(0)}, v_{i,3-\ell^*}\rangle| \leq \widetilde{O}(\sigma_0)$.) Therefore, we finished the proof that for every $t \leq T_{0,i}$, $\Lambda_{i,3-\ell}^{(t)} \leq \widetilde{O}(\sigma_0)$.

**Stage 2.** In the second stage, namely when $t \in [T_{0,i}, T_i]$, let us denote $\ell = 3 - \ell^*$ for abbreviation. Suppose we want to prove $\Lambda_{i,3-\ell^*}^{(t+1)} \leq \widetilde{O}(\sigma_0)$. By Claim C.7 and Claim C.8 again (but this time only using the upper bound part),
$$\langle w_{i,r}^{(t+1)}, v_{i,\ell}\rangle = \langle w_{i,r}^{(t)}, v_{i,\ell}\rangle + \eta \mathop{\mathbb{E}}_{(X,y)\sim\mathcal{Z}} \Big[\mathbb{1}_{y=i}\big(V_{i,r,\ell}(X) \pm O(\mathcal{E}_1 + \mathcal{E}_3)\big)\big(1 - \mathbf{logit}_i(F^{(t)}, X)\big) \\
\pm \mathbb{1}_{y\neq i}\big(\mathcal{E}_1 + \mathcal{E}_3 + \mathbb{1}_{v_{i,\ell}\in\mathcal{P}(X)}V_{i,r,\ell}(X)\big)\mathbf{logit}_i\big(F^{(t)}, X\big)\Big] \tag{C.21}$$

where
$$V_{i,r,\ell}(X) = \sum_{p\in\mathcal{P}_{v_{i,\ell}}(X)} \widetilde{\mathsf{ReLU}}'(\langle w_{i,r}, x_p\rangle)z_p$$

- Using Induction Hypothesis B.3 and $\Lambda_{i,3-\ell^*}^{(t)} \leq \widetilde{O}(\sigma_0)$, we know for $(X,y) \in \mathcal{Z}_m$, and for $(X,y) \in \mathcal{Z}_s$ but $y \neq i$,
$$V_{i,r,\ell}(X) = \sum_{p\in\mathcal{P}_{v_{i,\ell}}(X)} \widetilde{\mathsf{ReLU}}'\Big(\langle w_{i,r}^{(t)}, v_{i,\ell}\rangle z_p \pm \widetilde{o}(\sigma_0)\Big) z_p \leq \widetilde{O}(\sigma_0^{q-1})$$

- Otherwise for $(X,y) \sim \mathcal{Z}_s$ and $y = i$, we can use $\ell = 3-\ell^*$ and Induction Hypothesis B.3e to derive
$$V_{i,r,\ell}(X) = \sum_{p\in\mathcal{P}_{v_{i,\ell}}(X)} \widetilde{\mathsf{ReLU}}'\Big(\langle w_{i,r}^{(t)}, x_p\rangle\Big) z_p \leq \widetilde{O}(\sigma_0^{q-1}) \tag{C.22}$$

- For $(X,y) \sim \mathcal{Z}_s$ and $y \neq i$, we have $\mathbf{logit}_i(F^{(t)}, X) \leq O(\frac{1}{k})(1 - \mathbf{logit}_y(F^{(t)}, X))$ (see Claim C.4)

Putting these back to (C.21), we have
$$|\langle w_{i,r}^{(t+1)}, v_{i,\ell}\rangle| \leq |\langle w_{i,r}^{(t)}, v_{i,\ell}\rangle| + O(\eta) \mathop{\mathbb{E}}_{(X,y)\sim\mathcal{Z}_m} \Big[\mathbb{1}_{y=i}\Big(\widetilde{O}(\sigma_0^{q-1}) + O(\mathcal{E}_1 + \mathcal{E}_3)\Big)\Big(1 - \mathbf{logit}_y(F^{(t)}, X)\Big) \\
+ \mathbb{1}_{y\neq i}\widetilde{O}(\sigma_0^{q-1})\mathbf{logit}_i(F^{(t)}, X)\Big] \\
+ O\Big(\frac{\eta N_s}{N}\Big) \mathop{\mathbb{E}}_{(X,y)\sim\mathcal{Z}_s} \Big[\mathbb{1}_{y=i}\widetilde{O}(\sigma_0^{q-1}) \cdot \Big(1 - \mathbf{logit}_y(F^{(t)}, X)\Big) \tag{C.23} \\
+ \mathbb{1}_{y\neq i}\frac{\mathcal{E}_1 + \mathcal{E}_3 + \widetilde{O}(\sigma_0^{q-1})}{k}\Big(1 - \mathbf{logit}_y(F^{(t)}, X)\Big)\Big]$$

Since $N_s \ll N$, we can ignore single-view data and only focus on $(X,y) \in \mathcal{Z}_m$. Applying Claim C.17, we can conclude that
$$|\langle w_{i,r}^{(t+1)}, v_{i,\ell}\rangle| \leq |\langle w_{i,r}^{(T_0,i)}, v_{i,\ell}\rangle| + \eta O\Big(\frac{s}{k}T_0\Upsilon + \frac{T_0}{k}\Big) \cdot \widetilde{O}(\sigma_0^{q-1}) + \widetilde{O}(\sigma_0)$$

Recall that $T_0 = \Theta\big(\frac{k}{\eta\sigma_0^{q-2}}\big)$ from Claim C.11, and recall we have $s\Upsilon \leq O(1)$ (from (C.15)), we finish the proof that $|\langle w_{i,r}^{(t+1)}, v_{i,\ell}\rangle| \leq \widetilde{O}(\sigma_0)$. This means, $\Lambda_{i,3-\ell^*}^{(t)} \leq \widetilde{O}(\sigma_0)$ for every $t \in [T_{0,i}, T_i]$.

**Stage 3.** At the third stage, namely when $t \geq T_0$, we can continue from (C.23) but this time we do not ignore single-view data, and apply the naive bound $\mathbf{logit}_i(F^{(t)}, X) \leq 1 - \mathbf{logit}_y(F^{(t)}, X)$ for $(X, y) \in \mathcal{Z}_m$ and $i \neq y$. Recall again we denote $\ell = 3 - \ell^*$ for notational simplicity. If we denote by
$$S^{(t)} \stackrel{\text{def}}{=} \mathop{\mathbb{E}}_{(X,y)\sim\mathcal{Z}_m}\big[1 - \mathbf{logit}_y(F^{(t)}, X)\big]$$
$$G_i^{(t)} \stackrel{\text{def}}{=} \mathop{\mathbb{E}}_{(X,y)\sim\mathcal{Z}_s}\big[\mathbb{1}_{i=y} \cdot \big(1 - \mathbf{logit}_y(F^{(t)}, X)\big)\big]$$

then we have

$$\Lambda_{i,3-\ell^*}^{(t+1)} \leq \Lambda_{i,3-\ell^*}^{(t)} + \widetilde{O}\left(\eta S^{(t)} \sigma_0^{q-1}\right) + O\left(\frac{\eta N_s}{N}\right) \cdot \left(G_i^{(t)} + \frac{\sum_{j \in [k]} G_j^{(t)}}{k}\right) \cdot \widetilde{O}(\sigma_0^{q-1}) \qquad (C.24)$$

Recall

$$\text{Claim C.14} \implies \sum_{t \geq T_0} S^{(t)} \leq \widetilde{O}\left(\frac{k}{\eta}\right)$$

$$\text{Claim C.12} \implies \forall i' \in [k]: \sum_{t \geq T_0} G_i^{(t)} \leq \widetilde{O}\left(\frac{N}{\eta k \rho^{q-1}}\right)$$

Putting them back to (C.24), we know that as long as $\frac{N_s}{k\rho^{q-1}} \leq \widetilde{O}\left(\frac{1}{\sigma_0^{q-2}}\right)$ (already satisfied in (C.18)) and additionally[16]

$$k\sigma_0^{q-1} = \widetilde{O}(\sigma_0)$$

we have $\Lambda_{i,3-\ell^*}^{(t+1)} \leq \widetilde{O}(\sigma_0)$. $\square$

### C.4.4 Neuron Lottery Winning: $\mathcal{M}_i^{(0)}$

In this subsection, we prove that the neurons outside $\mathcal{M}_i^{(0)}$ for each index $i \in [k]$ is negligible. (They did not win the lottery so only the neurons in $\mathcal{M}_i^{(0)}$ count.)

**Lemma C.24.** *Suppose Parameter C.1 holds and suppose Induction Hypothesis B.3 holds for all iterations $< t$. Then,*

$$\forall i \in [k], \ \forall \ell \in [2], \ \forall r \in [m] \setminus \mathcal{M}_i^{(0)}: \qquad \langle w_{i,r}^{(t)}, v_{i,\ell} \rangle \leq \widetilde{O}(\sigma_0)$$

*Proof of Lemma C.24.* The proof is nearly identical to that of Lemma C.23.

**Stage 1.** In the first stage, namely when $t \leq T_{0,i}$, the proof is nearly identical to stage 1 of the proof of Lemma C.23. At a high level, this time we instead compare two sequences $x_t = \langle w_{i,r^*}^{(t)}, v_{i,\ell} \rangle$ and $y_t = \langle w_{i,r}^{(t)}, v_{i,\ell} \rangle$ for $r^* = \arg\max_{r \in [m]} \langle w_{i,r}^{(0)}, v_{i,\ell} \rangle$ and $r \in \mathcal{M}_{i,\ell}^{(0)}$. We skip the details for the sake of cleanness.

**Stage 2.** In the second stage, namely when $t \in [T_{0,i}, T_0]$, we can also completely reuse the stage 2 of the proof of Lemma C.23. The only slight difference is that in order to derive (C.22), this time we instead use Induction Hypothesis B.3f.

**Stage 3.** In the third stage, namely when $t \geq T_0$, we can also completely reuse the stage 3 of the proof of Lemma C.23. The only slight difference is that in order to derive (C.22), this time we instead use Induction Hypothesis B.3f. $\square$

### C.4.5 Noise Correlation is Small

In this subsection, we prove that the neurons correlate negligibly with the random noise, that is $\langle w_{i,r}^{(t)}, \xi_p \rangle$ is small, except for those single-view data on the lottery winning views. (Recall single-view data are learned through *memorization*, so the learner network can correlate with their noise $\xi_p$ significantly.)

---

[16]This is the place in our proof that we require $\sigma_0^{q-2} \leq \frac{1}{k}$. For simplicity, we have chosen $\sigma_0^{q-2} \leq \frac{1}{k}$ in Parameter C.1.

**Lemma C.25.** *Suppose Parameter C.1 holds and suppose Induction Hypothesis B.3 holds for all iterations $< t$.[17] For every $\ell \in [2]$, for every $r \in [m]$, for every $(X,y) \in \mathcal{Z}_m$ and $i \in [k]$, or for every $(X,y) \in \mathcal{Z}_s$ and $i \in [k] \setminus \{y\}$:*

(a) *For every $p \in \mathcal{P}_{v_{i,\ell}}(X)$, we have: $\langle w_{i,r}^{(t)}, \xi_p \rangle \leq \widetilde{o}(\sigma_0)$.*

(b) *For every $p \in \mathcal{P}(X) \setminus \left(\mathcal{P}_{v_{i,1}}(X) \cup \mathcal{P}_{v_{i,2}}(X)\right)$, we have: $|\langle w_{i,r}^{(t)}, \xi_p \rangle| \leq \widetilde{O}(\sigma_0)$.*

(c) *For every $p \in [P] \setminus \mathcal{P}(X)$, we have: $|\langle w_{i,r}^{(t)}, \xi_p \rangle| \leq \widetilde{O}(\sigma_0 \gamma k)$.*

*In addition, for every $(X,y) \in \mathcal{Z}_s$, every $i \in [k]$, every $r \in [m]$, every $\ell \in [2]$,*

(d) *For every $p \in \mathcal{P}_{v_{i,\ell}}(X)$, if $(i, 3-\ell) \in \mathcal{M}$ we have: $|\langle w_{i,r}^{(t)}, \xi_p \rangle| \leq \widetilde{O}(\sigma_0)$.*

(e) *For every $p \in \mathcal{P}_{v_{i,\ell}}(X)$, if $r \in [m] \setminus \mathcal{M}_i^{(0)}$ we have: $|\langle w_{i,r}^{(t)}, \xi_p \rangle| \leq \widetilde{O}(\sigma_0)$.*

We prove Lemma C.25 after we establish the following claim.

**Claim C.26.** *For every $(X,y) \in \mathcal{Z}$, $i \in [k]$, $r \in [m]$, and $p \in [P]$, suppose it satisfies $|\langle w_{i,r}^{(t)}, x_p \rangle| \leq A$ for every $t < t_0$ where $t_0$ is any iteration $t_0 \leq T$. Then,*

- *If $(X,y) \in \mathcal{Z}_m$ then $\left|\langle w_{i,r}^{(t)}, \xi_p \rangle\right| \leq \widetilde{O}\left(\frac{kA^{q-1}}{N\sigma_0^{q-2}} + \frac{k^5 A^{q-1}}{s^2 N}\right) + \frac{\eta T}{\sqrt{d}}$*

- *If $(X,y) \in \mathcal{Z}_s$ then $\left|\langle w_{i,r}^{(t)}, \xi_p \rangle\right| \leq \widetilde{O}\left(\frac{kA^{q-1}}{N\sigma_0^{q-2}} + \frac{A^{q-1}}{\rho^{q-1}}\right) + \frac{\eta T}{\sqrt{d}}$*

We first prove Lemma C.25 using Claim C.26 (which is trivial), and then we prove Claim C.26.

*Proof of Lemma C.25.*

- In the case of Lemma C.25a, since we have $|\langle w_{i,r}^{(t')}, x_p \rangle| \leq \widetilde{O}(1)$ for every $t' < t$ according to our Induction Hypothesis B.3a and B.3g, by applying Claim C.26 we immediately have $|\langle w_{i,r}^{(t)}, \xi_p \rangle| \leq \widetilde{o}(\sigma_0)$ once we plug in $A = \widetilde{O}(1)$, $N \geq \widetilde{\omega}\left(\frac{k}{\sigma_0^{q-1}}\right)$, and $N \geq \widetilde{\omega}\left(\frac{k^5}{\sigma_0}\right)$.

- In the case of Lemma C.25b, since we have $|\langle w_{i,r}^{(t')}, x_p \rangle| \leq \widetilde{O}(\sigma_0)$ for every $t' < t$ according to our Induction Hypothesis B.3b, by applying Claim C.26 we immediately have $|\langle w_{i,r}^{(t)}, \xi_p \rangle| \leq \widetilde{O}(\sigma_0)$ once we plug in $A = \widetilde{O}(\sigma_0)$, $N \geq k^5$ and $(\sigma_0)^{q-2} \leq \rho^{q-1}$.

- In the case of Lemma C.25c, since we have $|\langle w_{i,r}^{(t')}, x_p \rangle| \leq \widetilde{O}(\sigma_0 \gamma k)$ for every $t' < t$ according to our Induction Hypothesis B.3c, by applying Claim C.26 we immediately have $|\langle w_{i,r}^{(t)}, \xi_p \rangle| \leq \widetilde{O}(\sigma_0 \gamma k)$ once we plug in $A = \widetilde{O}(\sigma_0 \gamma k)$, $N \geq k^5$ and $(\sigma_0 \gamma k)^{q-2} \leq \rho^{q-1}$.

- In the case of Lemma C.25d, since we have $|\langle w_{i,r}^{(t')}, x_p \rangle| \leq \widetilde{O}(\sigma_0)$ for every $t' < t$ according to our Induction Hypothesis B.3e, by applying Claim C.26 we immediately have $|\langle w_{i,r}^{(t)}, \xi_p \rangle| \leq \widetilde{O}(\sigma_0)$ once we plug in $A = \widetilde{O}(\sigma_0)$, $N \geq k^5$ and $(\sigma_0)^{q-2} \leq \rho^{q-1}$.

- In the case of Lemma C.25e, since we have $|\langle w_{i,r}^{(t')}, x_p \rangle| \leq \widetilde{O}(\sigma_0)$ for every $t' < t$ according to our Induction Hypothesis B.3f, by applying Claim C.26 we immediately have $|\langle w_{i,r}^{(t)}, \xi_p \rangle| \leq \widetilde{O}(\sigma_0)$ once we plug in $A = \widetilde{O}(\sigma_0)$, $N \geq k^5$ and $(\sigma_0)^{q-2} \leq \rho^{q-1}$.

□

---

[17]In particular, we need to ensure $(\sigma_0)^{q-2} \leq \rho^{q-1}$ and $N \geq \widetilde{\Omega}\left(\frac{k^5}{\sigma_0^{q-1}}\right)$ in the proof of this lemma.

*Proof of Claim C.26.* Recall from our earlier calculation (see (C.6)) that for every $(X, y) \in \mathcal{Z}$ and $p \in [P]$, if $y = i$ then

$$\langle w_{i,r}^{(t+1)}, \xi_p \rangle = \langle w_{i,r}^{(t)}, \xi_p \rangle + \widetilde{\Theta}\Big(\frac{\eta}{N}\Big)\widetilde{\mathsf{ReLU}}'(\langle w_{i,r}^{(t)}, x_p \rangle)\Big(1 - \mathbf{logit}_i(F^{(t)}, X)\Big) \pm \frac{\eta}{\sqrt{d}}$$

for similar reason, if $y \neq i$, then

$$\langle w_{i,r}^{(t+1)}, \xi_p \rangle = \langle w_{i,r}^{(t)}, \xi_p \rangle - \widetilde{\Theta}\Big(\frac{\eta}{N}\Big)\widetilde{\mathsf{ReLU}}'(\langle w_{i,r}^{(t)}, x_p \rangle)\mathbf{logit}_i(F^{(t)}, X) \pm \frac{\eta}{\sqrt{d}}$$

Note by our assumption, we have $\widetilde{\mathsf{ReLU}}'(\langle w_{i,r}^{(t)}, x_p \rangle) \leq \widetilde{O}(A^{q-1})$. For obvious reason, when $t = T_0 = \Theta\big(\frac{k}{\eta \sigma_0^{q-2}}\big)$ (see Claim C.11)

$$\left|\langle w_{i,r}^{(t)}, \xi_p \rangle\right| \leq \widetilde{O}\Big(\frac{\eta}{N}A^{q-1}T_0\Big) + \frac{\eta T_0}{\sqrt{d}} \leq \widetilde{O}\bigg(\frac{kA^{q-1}}{N\sigma_0^{q-2}}\bigg) + \frac{\eta T_0}{\sqrt{d}}$$

**Case 1: multi-view data.** We first consider $(X, y) \in \mathcal{Z}_m$. Using Claim C.16 and Claim C.14, we have

$$y = i \implies \sum_{t=T_0}^{T} 1 - \mathbf{logit}_y\left(F^{(t)}, X\right) \leq \widetilde{O}\left(\frac{k^4}{s^2}\right) \cdot \sum_{t=T_0}^{T} \mathop{\mathbb{E}}_{(X,y) \sim \mathcal{Z}_m}\left[1 - \mathbf{logit}_y\left(F^{(t)}, X\right)\right] \leq \widetilde{O}\left(\frac{k^5}{s^2 \eta}\right)$$

$$y \neq i \implies \sum_{t=T_0}^{T} \mathbf{logit}_i\left(F^{(t)}, X\right) \leq \sum_{t=T_0}^{T} 1 - \mathbf{logit}_y\left(F^{(t)}, X\right) \leq \widetilde{O}\left(\frac{k^5}{s^2 \eta}\right)$$

Combining this with the bound at $t = T_0$, we have

$$\left|\langle w_{i,r}^{(t)}, \xi_p \rangle\right| \leq \widetilde{O}\bigg(\frac{kA^{q-1}}{N\sigma_0^{q-2}} + \frac{k^5 A^{q-1}}{s^2 N}\bigg) + \frac{\eta T}{\sqrt{d}}$$

**Case 2: single-view data.** Let us now consider $(X, y) \in \mathcal{Z}_s$. Recall from Claim C.12b that

$$\sum_{t=T_0}^{T} \left(1 - \mathbf{logit}_y\left(F^{(t)}, X\right)\right) \leq \widetilde{O}\left(\frac{N}{\eta \rho^{q-1}}\right)$$

so using the same analysis, we have

$$\left|\langle w_{i,r}^{(t)}, \xi_p \rangle\right| \leq \widetilde{O}\bigg(\frac{kA^{q-1}}{N\sigma_0^{q-2}} + \frac{A^{q-1}}{\rho^{q-1}}\bigg) + \frac{\eta T}{\sqrt{d}} \qquad \square$$

### C.4.6 Diagonal Correlations are Nearly Non-Negative

**Lemma C.27.** *Suppose Parameter C.1 holds and suppose Induction Hypothesis B.3 holds for all iterations $< t$. Then,*

$$\forall i \in [k]\ , \forall r \in [m]\ , \forall \ell \in [2]: \qquad \langle w_{i,r}^{(t)}, v_{i,\ell} \rangle \geq -\widetilde{O}(\sigma_0)\ .$$

*Proof of Lemma C.27.* Let us consider any iteration $t$ so that $\langle w_{i,r}^{(t)}, v_{i,\ell} \rangle \leq -\widetilde{\Omega}(\sigma_0)$. We start from this iteration to see how negative the next iterations can be. Without loss of generality we consider the case when $\langle w_{i,r}^{(t')}, v_{i,\ell} \rangle \leq -\widetilde{\Omega}(\sigma_0)$ holds for every $t' \geq t$.

By Claim C.7 and Claim C.8,

$$\langle w_{i,r}^{(t+1)}, v_{i,\ell}\rangle \geq \langle w_{i,r}^{(t)}, v_{i,\ell}\rangle + \eta \mathop{\mathbb{E}}_{(X,y)\sim\mathcal{Z}}\left[\mathbb{1}_{y=i}\left(V_{i,r,\ell}(X) - \widetilde{O}(\sigma_p P)\right)\left(1 - \mathbf{logit}_i(F^{(t)}, X)\right)\right.$$
$$\left. - \mathbb{1}_{y\neq i}\left(\mathcal{E}_1 + \mathcal{E}_3 + \mathbb{1}_{v_{i,\ell}\in\mathcal{P}(X)}V_{i,r,\ell}(X)\right)\mathbf{logit}_i\left(F^{(t)}, X\right)\right]$$

Recall $V_{i,r,\ell}(X) = \sum_{p\in\mathcal{P}_{v_{i,\ell}}(X)} \widetilde{\mathsf{ReLU}}'(\langle w_{i,r}^{(t)}, x_p\rangle) z_p$. Since $V_{i,r,\ell}(X) \geq 0$ we can ignore it in the first occurence corresponding to $\mathbb{1}_{y=i}$. Also, applying Induction Hypothesis B.3, we know that as long as $(X,y) \in \mathcal{Z}$ and $y \neq i$, it satisfies

$$V_{i,r,\ell}(X) = \sum_{p\in\mathcal{P}_{v_{i,\ell}}(X)} \widetilde{\mathsf{ReLU}}'\left(\langle w_{i,r}^{(t)}, v_{i,\ell}\rangle z_p \pm \widetilde{o}(\sigma_0)\right) z_p = 0$$

because we have assumed $\langle w_{i,r}^{(t)}, v_{i,\ell}\rangle \leq -\widetilde{\Omega}(\sigma_0)$. Therefore,

$$\langle w_{i,r}^{(t+1)}, v_{i,\ell}\rangle \geq \langle w_{i,r}^{(t)}, v_{i,\ell}\rangle - \eta \mathop{\mathbb{E}}_{(X,y)\sim\mathcal{Z}}\left[\mathbb{1}_{y=i}\widetilde{O}(\sigma_p P)\left(1 - \mathbf{logit}_i(F^{(t)}, X)\right)\right.$$
$$\left. + \mathbb{1}_{y\neq i}\left(\mathcal{E}_1 + \mathcal{E}_3\right)\mathbf{logit}_i(F^{(t)}, X)\right]$$

We first consider every $t \leq T_0 \stackrel{\text{def}}{=} \Theta\left(\frac{k}{\eta\sigma_0^{q-2}}\right)$ (recall Claim C.11). Using Claim C.4 we have $\mathbf{logit}_i(F^{(t)}, X) = O(\frac{1}{k})$. This implies

$$\langle w_{i,r}^{(t)}, v_{i,\ell}\rangle \geq -\widetilde{O}(\sigma_0) - O\left(\frac{\eta T_0}{k}\right)(\mathcal{E}_1 + \mathcal{E}_3) = -\widetilde{O}(\sigma_0) - O\left(\frac{1}{\sigma_0^{q-2}}\right)(\mathcal{E}_1 + \mathcal{E}_3) \geq -\widetilde{O}(\sigma_0)$$

(Above, the last inequality uses our earlier parameter choices, see (C.12).)

As for $t \geq T_0$, we combining this with $\mathbf{logit}_i\left(F^{(t)}, X\right) \leq 1 - \mathbf{logit}_y\left(F^{(t)}, X\right)$ for $i \neq y$ and $(X,y) \in \mathcal{Z}_m$, and $\mathbf{logit}_i\left(F^{(t)}, X\right) \leq O(\frac{1}{k})\left(1 - \mathbf{logit}_y\left(F^{(t)}, X\right)\right)$ for $i \neq y$ and $(X,y) \in \mathcal{Z}_s$, we have

$$\langle w_{i,r}^{(t+1)}, v_{i,\ell}\rangle \geq \langle w_{i,r}^{(t)}, v_{i,\ell}\rangle - \eta \mathop{\mathbb{E}}_{(X,y)\sim\mathcal{Z}_m}\left[O(\mathcal{E}_1 + \mathcal{E}_3)\left(1 - \mathbf{logit}_y(F^{(t)}, X)\right)\right]$$
$$- O\left(\frac{\eta N_s}{N}\right)\mathop{\mathbb{E}}_{(X,y)\sim\mathcal{Z}_s}\left[\mathbb{1}_{y=i}\widetilde{O}(\sigma_p P)\left(1 - \mathbf{logit}_y(F^{(t)}, X)\right)\right.$$
$$\left. + \mathbb{1}_{y\neq i}\frac{\mathcal{E}_1 + \mathcal{E}_3}{k}\left(1 - \mathbf{logit}_y(F^{(t)}, X)\right)\right]$$

Finally, recall

Claim C.14 $\implies \sum_{t\geq T_0}\mathbb{E}_{(X,y)\sim\mathcal{Z}_m}\left[1 - \mathbf{logit}_y\left(F^{(t)}, X\right)\right] \leq \widetilde{O}\left(\frac{k}{\eta}\right)$

Claim C.12 $\implies \forall i' \in [k]: \sum_{t\geq T_0}\mathbb{E}_{(X,y)\sim\mathcal{Z}_s}\left[\mathbb{1}_{i=y}\cdot\left(1 - \mathbf{logit}_y\left(F^{(t)}, X\right)\right)\right] \leq \widetilde{O}\left(\frac{N}{\eta k\rho^{q-1}}\right)$

Therefore, we know for every $t \in [T_0, T]$:

$$\langle w_{i,r}^{(t)}, v_{i,\ell}\rangle \geq \langle w_{i,r}^{(T_0)}, v_{i,\ell}\rangle - \widetilde{O}\big(k\cdot(\mathcal{E}_1 + \mathcal{E}_3)\big) \stackrel{\text{①}}{\geq} \langle w_{i,r}^{(T_0)}, v_{i,\ell}\rangle - \widetilde{O}(\sigma_0) \geq -\widetilde{O}(\sigma_0)$$

(Above, the inequality ① uses our earlier parameter choices, see (C.18).) $\square$

### C.5 Putting All Together

We are now ready to restate Theorem C.2 and prove it.

**Theorem C.2.** *Under Parameter C.1, for any $m \in \left[\widetilde{\Omega}(1), \widetilde{o}(\frac{1}{\sigma_0})\right]$ and sufficiently small $\eta \leq \frac{1}{\mathsf{poly}(k)}$, our Induction Hypothesis B.3 holds for all iterations $t = 0, 1, \ldots, T$.*

*Proof of Theorem C.2.* At iteration $t$, we first calculate

$$\forall p \in P_{v_{j,\ell}}(X): \qquad \langle w_{i,r}^{(t)}, x_p \rangle = \langle w_{i,r}^{(t)}, v_{j,\ell} \rangle z_p + \sum_{v' \in \mathcal{V}} \alpha_{p,v'} \langle w_{i,r}^{(t)}, v' \rangle + \langle w_{i,r}^{(t)}, \xi_p \rangle \qquad (C.25)$$

$$\forall p \in [P] \setminus P(X): \qquad \langle w_{i,r}^{(t)}, x_p \rangle = \sum_{v' \in \mathcal{V}} \alpha_{p,v'} \langle w_{i,r}^{(t)}, v' \rangle + \langle w_{i,r}^{(t)}, \xi_p \rangle \qquad (C.26)$$

It is easy to verify Induction Hypothesis B.3 holds at iteration $t = 0$ (merely some simple high probability bounds on Gaussian random variables). Suppose Induction Hypothesis B.3 holds for all iterations $< t$. We have established several lemmas

$$\text{Lemma C.22} \implies \forall i \in [k] \, \forall r \in [m] \, \forall j \in [k] \setminus \{i\}: \qquad |\langle w_{i,r}^{(t)}, v_{j,\ell} \rangle| \leq \widetilde{O}(\sigma_0) \quad (C.27)$$

$$\text{Lemma C.21+Lemma C.27} \implies \forall i \in [k] \, \forall r \in [m] \, \forall \ell \in [2]: \qquad \langle w_{i,r}^{(t)}, v_{i,\ell} \rangle \in \big[-\widetilde{O}(\sigma_0), \widetilde{O}(1)\big] \quad (C.28)$$

$$\text{Lemma C.23+Lemma C.27} \implies \forall (i, \ell^*) \in \mathcal{M} \, \forall r \in [m]: \qquad |\langle w_{i,r}^{(t)}, v_{i,3-\ell^*} \rangle| \leq \widetilde{O}(\sigma_0) \quad (C.29)$$

$$\text{Lemma C.27} \implies \forall i \in [k] \,, \forall r \in [m] \,, \forall \ell \in [2]: \qquad \langle w_{i,r}^{(t)}, v_{i,\ell} \rangle \geq -\widetilde{O}(\sigma_0) \,. \quad (C.30)$$

$$\text{Lemma C.24+Lemma C.27} \implies \forall i \in [k], \forall \ell \in [2], \forall r \in [m] \setminus \mathcal{M}_i^{(0)}: \qquad |\langle w_{i,r}^{(t)}, v_{i,\ell} \rangle| \leq \widetilde{O}(\sigma_0) \quad (C.31)$$

- To prove B.3a, it suffices to plug (C.27), (C.28) into (C.25), use $\alpha_{p,v'} \in [0, \gamma]$, use $|\mathcal{V}| = 2k$, and use $|\langle w_{i,r}^{(t)}, \xi_p \rangle| \leq \widetilde{o}(\sigma_0)$ from Lemma C.25a.

- To prove B.3b, it suffices to plug (C.27), (C.28) into (C.25), use $\alpha_{p,v'} \in [0, \gamma]$, use $|\mathcal{V}| = 2k$, and use $|\langle w_{i,r}^{(t)}, \xi_p \rangle| \leq \widetilde{O}(\sigma_0)$ from Lemma C.25b.

- To prove B.3c, it suffices to plug (C.27), (C.28) into (C.26), use $\alpha_{p,v'} \in [0, \gamma]$, use $|\mathcal{V}| = 2k$, and use $|\langle w_{i,r}^{(t)}, \xi_p \rangle| \leq \widetilde{O}(\sigma_0 \gamma k)$ from Lemma C.25c.

- To prove B.3d, it suffices to plug (C.27), (C.28) into (C.25), use $\alpha_{p,v'} \in [0, \gamma]$, use $|\mathcal{V}| = 2k$.

- To prove B.3e, it suffices to plug (C.27), (C.29) into (C.25), use $\alpha_{p,v'} \in [0, \gamma]$, use $|\mathcal{V}| = 2k$, and use $|\langle w_{i,r}^{(t)}, \xi_p \rangle| \leq \widetilde{O}(\sigma_0)$ from Lemma C.25d.

- To prove B.3f, it suffices to plug (C.27), (C.31) into (C.25), use $\alpha_{p,v'} \in [0, \gamma]$, use $|\mathcal{V}| = 2k$, and use $|\langle w_{i,r}^{(t)}, \xi_p \rangle| \leq \widetilde{O}(\sigma_0)$ from Lemma C.25e.

- To prove B.3g, it suffices to note that (C.28) exactly implies $\Lambda_i^{(t)} \leq \widetilde{O}(1)$, and note that Claim C.10 (which says as long as $\Lambda_i^{(t)} \leq O(1/m_0)$, then it must grow by $\Lambda_i^{(t+1)} \geq \Lambda_i^{(t)} + \Theta\left(\frac{\eta}{k}\right) \widetilde{\mathsf{ReLU}}'(\Lambda_i^{(t)}))$ implies $\Lambda_i^{(t)} \geq \Omega(\Lambda_i^{(0)}) \geq \widetilde{\Omega}(\sigma_0)$.

- To prove B.3h, it suffices to invoke (C.30).

- To prove B.3i, it suffices to invoke (C.31).

□

# D  Single Model and Ensemble: Theorem Statements

We can now state the general version of the main theorem for single model, as below:

**Theorem 1** (single model, restated). *For sufficiently large $k > 0$, every $m \in \left[\mathsf{polylog}(k), \frac{1}{s\sigma_0^q \mathsf{polylog}(k)}\right]$, every $\eta \leq \frac{1}{\mathsf{poly}(k)}$, after $T = \frac{\mathsf{poly}(k)}{\eta}$ many iterations, when Parameter C.1 is satisfied, with probability at least $1 - e^{-\Omega(\log^2 k)}$:*

- *(training accuracy is perfect) for every $(X, y) \in \mathcal{Z}$:*

$$\forall i \neq y \colon F_y^{(T)}(X) \geq F_i^{(T)}(X) + \Omega(\log k).$$

- *(multi-view testing is good) for every $i, j \in [k]$ we have $\widetilde{O}(1) \geq \Phi_i^{(T)} \geq 0.4\Phi_j^{(T)} + \Omega(\log k)$, and thus*

$$\Pr_{(X,y) \in \mathcal{D}_m} \left[ F_y^{(T)}(X) \geq \max_{j \neq y} F_j^{(T)}(X) + \Omega(\log k) \right] \geq 1 - e^{-\Omega(\log^2 k)}$$

- *(single-view testing is bad) for every $(i, \ell) \in \mathcal{M}$ we have $\Phi_{i,3-\ell}^{(T)} \leq \widetilde{O}(\sigma_0 m) \ll \frac{1}{\mathsf{polylog}(k)}$, and since $|\mathcal{M}| \geq k(1 - o(1))$, we have[18]*

$$\Pr_{(X,y) \in \mathcal{D}_s} \left[ F_y^{(T)}(X) \geq \max_{j \neq y} F_j^{(T)}(X) - \frac{1}{\mathsf{polylog}(k)} \right] \leq \frac{1}{2}(1 + o(1))$$

We also state the general version of the main theorem for ensemble model, as below:

**Theorem 2** (ensemble accuracy, restated). *In the same setting as above, suppose $\{F^{[w]}\}_{w \in [K]}$ are $K$ independently randomly trained models with $m \in \left[\log^{\Omega(1)}(k), \log^{O(1)} k\right]$ for $T = \frac{\mathsf{poly}(k)}{\eta}$ iterations each. Let us define $G(X) = \frac{1}{K} \sum_w F^{[w]}(X)$.*

- *(training is perfect) same as the single model;*
- *(multi-view testing is good) same as the single model;*
- *(single-view testing is good) when $K \geq \mathsf{polylog}(k)$, ensemble model satisfies*

$$\Pr_{(X,y) \sim \mathcal{D}_s} \left[ G_y(X) \geq \max_{i \in [k] \setminus \{y\}} G_i(X) + \frac{1}{\mathsf{polylog}(k)} \right] \geq 1 - e^{-\Omega(\log^2 k)}$$

## D.1 Proof of Theorem 1

Since Theorem C.2 implies the induction hypothesis holds for every $t \leq T$, we have according to Claim C.14 and Claim C.12b that

$$\sum_{t=T_0}^{T} \mathbb{E}_{(X,y) \sim \mathcal{Z}_m} \left[ 1 - \mathbf{logit}_y \left( F^{(t)}, X \right) \right] \leq \widetilde{O}\left(\frac{k}{\eta}\right)$$

$$\sum_{t=T_0}^{T} \mathbb{E}_{(X,y) \sim \mathcal{Z}_s} \left( 1 - \mathbf{logit}_y \left( F^{(t)}, X \right) \right) \leq \widetilde{O}\left(\frac{N}{\eta \rho^{q-1}}\right) \quad (D.1)$$

Also recall that our training objective is

$$L(F^{(t)}) = \mathbb{E}_{(X,y) \sim \mathcal{Z}}[-\log \mathbf{logit}_y(F^{(t)}, X)]$$

Now, since for every data,

- if $\mathbf{logit}_y(F^{(t)}, X) \geq \frac{1}{2}$ then $-\log \mathbf{logit}_y(F^{(t)}, X) \leq O\left(1 - \mathbf{logit}_y(F^{(t)}, X)\right)$;
- if $\mathbf{logit}_y(F^{(t)}, X) \leq \frac{1}{2}$, this cannot happen for too many tuples $(X, y, t)$ thanks to (D.1), and when this happens we have a naive bound $-\log \mathbf{logit}_y(F^{(t)}, X) \in [0, \widetilde{O}(1)]$ using Claim C.4.

---

[18]Note that we have assumed for simplicity that there are 2 views with equal probability, and this is why the testing accuracy is close to $\frac{1}{2}$; in more general settings, as we stated in Section 4, this accuracy may be some other constant $\mu$.

Therefore, we can safely conclude using (D.1) that, when $T \geq \mathsf{poly}(k)/\eta$,

$$\frac{1}{T} \sum_{t=T_0}^{T} \mathbb{E}_{(X,y)\sim \mathcal{Z}}[-\log \mathbf{logit}_y(F^{(t)}, X)] \leq \frac{1}{\mathsf{poly}(k)}$$

On the other hand, since we are using full gradient descent and the objective function is $O(1)$-Lipscthiz continuous, it means the objective value is monotonically non-increasing. In other words, we have

$$\mathbb{E}_{(X,y)\sim \mathcal{Z}}\left(1 - \mathbf{logit}_y\left(F^{(T)}, X\right)\right) \leq \mathbb{E}_{(X,y)\sim \mathcal{Z}}[-\log \mathbf{logit}_y(F^{(T)}, X)] \leq \frac{1}{\mathsf{poly}(k)}$$

also for the last iteration $T$. This immediately implies that the training accuracy is perfect.

As for the multi-view test accuracy, we recall from Claim C.16 that $0.4\Phi_i^{(T)} - \Phi_j^{(T)} \leq -\Omega(\log k)$ for every $i \neq j$. This combined with the function approximation Claim C.9 shows that with high probability $F_y^{(T)}(X) \geq \max_{j\neq y} F_j^{(T)}(X) + \Omega(\log k)$ for every $(X, y) \in \mathcal{D}_m$.

As for the single-view test accuracy, whenever $(i, \ell) \in \mathcal{M}$, using Lemma C.23 we have $\Lambda_{i,3-\ell}^{(T)} \leq \widetilde{O}(\sigma_0)$ so $\Phi_{i,3-\ell}^{(T)} \leq \widetilde{O}(\sigma_0 m)$.

Now, for every single-view data $(X, y) \in \mathcal{D}_s$ with $y = i$, we know that with half probability $\widehat{\ell}(X) = 3 - \ell$. When this happens, according to Claim C.9, we have $F_y^{(T)}(X) \leq O(\rho) + \frac{1}{\mathsf{polylog}(k)}$ (using Def. 3.1 for the single-view distribution).

For every other $j \neq y$, whenever suppose $\ell' = \arg\max_{\ell' \in [2]}\{\Phi_{j,\ell'}^{(T)}\}$, we have $\Phi_{j,\ell'}^{(T)} \geq \Omega(\log k)$ by a few lines above. This means, as long as $v_{j,\ell'} \in \mathcal{V}(X)$ (which happens with probability $s/k$ for every $j$), invoking Claim C.9 again, that $F_j^{(T)}(X) \geq \widetilde{\Omega}(\Gamma)$. In other words, when this happens for some $j \in [k] \setminus \{i\}$ (which happens with probability at least $1 - e^{-\Omega(\log^2 k)}$, we have

$$F_y^{(T)}(X) \leq \max_{j\neq y} F_j^{(T)}(X) - \frac{1}{\mathsf{polylog}(k)} \tag{D.2}$$

To sum up, we have shown for every $(i, \ell) \in \mathcal{M}$, for every $(X, y) \in \mathcal{D}_s$ with $y = i$, we know that with probability at least $\frac{1}{2}(1 - o(1))$, inequality (D.2) holds. Since the size $|\mathcal{M}| \geq k(1 - o(1))$ (see Proposition B.2), we finish the proof. ∎

### D.2 Proof of Theorem 2

Recall from Proposition B.2 that for every $i \in [k]$, $\ell \in [2]$ and every model $F^{[w]}$, the probability for $(i, \ell)$ to be included in the set $\mathcal{M}^{[w]}$ (defined by model $F^{[w]}$) is at least $m^{-O(1)}$. When this happens, we also have $\Phi_{i,\ell}^{(T)} \geq \Omega(\log k)$ for this model (because $\Phi_i^{(T)} \geq \Omega(\log k)$ while $\Phi_{i,3-\ell}^{(T)} \ll 1$). Let us denote it as $\Phi_{i,\ell}^{[w]}$.

Now, for every $(X, y) \in \mathcal{D}_s$, with probability at least $1 - e^{-\Omega(\log^2 k)}$, letting $\ell = \widehat{\ell}(X)$, we have (see Claim C.9)

$$\text{for every } F^{[w]} \text{ with } \Phi_{y,\ell}^{[w]} \geq \Omega(\log k) \implies F_y^{[w]}(X) \geq \Phi_{y,\ell}^{[w]} - \frac{1}{\mathsf{polylog}(k)} \geq \Omega(\log k)$$

$$\text{for every } F^{[w]} \text{ with } i \neq y \implies F_i^{[w]}(X) \leq \Gamma(\Phi_{i,1} + \Phi_{i,2}) + \frac{1}{\mathsf{polylog}(k)} \leq O(\Gamma)$$

Therefore, once we have $K \geq m^{\Omega(1)}$ models in the ensemble, and suppose $\Gamma \leq \frac{1}{m^{\Omega(1)}}$, after taking average, we shall have $G_y(X) \geq G_i(X) + \frac{1}{\mathsf{polylog}(k)}$ for every $i \neq [k]$. ∎

# E  Knowledge Distillation: Theorem Statement

In this section we show how knowledge distillation (both for ensemble and for self-distillation) can improve the final generalization accuracy. For every $i$, let us define the *truncated scaled logit* as (for $\tau = \frac{1}{\log^2 k}$):

$$\mathbf{logit}_i^\tau(F, X) = \frac{e^{\min\{\tau^2 F_i(X), 1\}/\tau}}{\sum_{j \in [k]} e^{\min\{\tau^2 F_j(X), 1\}/\tau}}$$

This logit function should be reminiscent of the logit function with temperature used by the seminal knowledge distillation paper by Hinton, Vinyals and Dean [42]; we use the truncation function instead which is easier to analyze.

## E.1  Using Ensemble for Knowledge Distillation

Suppose $\{F^{[i]}\}_{i \in [K]}$ are $K = \widetilde{\Theta}(1)$ independently trained models of $F$ for $T = O\left(\frac{\mathsf{poly}(k)}{\eta}\right)$ iterations (i.e., the same setting as Theorem 1). Let us define their ensemble

$$G(X) = \frac{\Xi}{K} \sum_i F^{[i]}(X) \qquad \text{for some } \Xi = \widetilde{\Theta}(1) \tag{E.1}$$

Recall from (4.3) that we train a new network $F$ from random initialization, and at every iteration $t$, we update each weight $w_{i,r}$ by:

$$w_{i,r}^{(t+1)} = w_{i,r}^{(t)} - \eta \nabla_{w_{i,r}} L(F^{(t)}) - \eta' \mathop{\mathbb{E}}_{(X,y) \sim \mathcal{Z}} \left( \left(\mathbf{logit}_i^\tau(F^{(t)}, X) - \mathbf{logit}_i^\tau(G, X)\right)^- \nabla_{w_{i,r}} F_i^{(t)}(X) \right)$$

(4.3) restated

Let $F^{(t)}$ be the resulted network obtained by distilling $G$ using algorithm (4.3) at iteration $t$. We have the following theorem:

**Theorem 3** (ensemble distillation, restated). *For sufficiently large $k > 0$, for every $m \in \left[\log^{\Omega(1)}(k), \log^{O(1)} k\right]$, every $\eta \leq \frac{1}{\mathsf{poly}(k)}$, setting $\eta' = \eta \mathsf{poly}(k)$, after $T = \frac{\mathsf{poly}(k)}{\eta}$ many iterations, when Parameter F.2 is satisfied, with probability at least $1 - e^{-\Omega(\log^2 k)}$, for at least 90% of the iterations $t \leq T$:*

- *(training accuracy is perfect) for every $(X, y) \in \mathcal{Z}$:*

$$\forall i \neq y \colon F_y^{(t)}(X) \geq F_i^{(t)}(X) + \Omega(\log k).$$

- *(multi-view testing is good) for every $i, j \in [k]$ we have $\widetilde{O}(1) \geq \Phi_i^{(t)} \geq 0.4 \Phi_j^{(t)} + \Omega(\log k)$, and thus*

$$\mathop{\mathbf{Pr}}_{(X,y) \in \mathcal{D}_m} \left[ F_y^{(t)}(X) \geq \max_{j \neq y} F_j^{(t)}(X) + \Omega(\log k) \right] \geq 1 - e^{-\Omega(\log^2 k)}$$

- *(single-view testing is good) for every $i \in [k]$ and $\ell \in [2]$ we have $\Phi_{i,\ell}^{(t)} \geq \Omega(\log k)$ and thus*

$$\mathop{\mathbf{Pr}}_{(X,y) \in \mathcal{D}_s} \left[ F_y^{(T)}(X) \geq \max_{j \neq y} F_j^{(T)}(X) + \Omega(\log k) \right] \leq 1 - e^{-\Omega(\log^2 k)}$$

Our proof to Theorem 3 is in the next Section F.

## E.2  Self-Distillation: Using a Single Model to Distill Itself

Recall in the self-distillation case, we made an additional assumption for simplicity:

**Assumption 4.1** (balanced $\mathcal{D}_m$, restated)**.** *In Def. 3.1, for multi-view data $(X, y)$, we additionally assume that the marginal distributions of $\sum_{p \in \mathcal{P}_v(X)} z_p^q \in [1, O(1)]$ for $v \in \{v_{y,1}, v_{y,2}\}$.*

Let $G$ be a single model trained in the same way as Theorem 1. At the end of training, we scale it up by a small factor $G \leftarrow \log^4 k \cdot G$. Define a "lottery winning" set

$$\mathcal{M}_G \stackrel{\text{def}}{=} \left\{ (i, \ell^*) \in [k] \times [2] \,\Big|\, \Lambda_{i,\ell^*}^{(0)} \geq \Lambda_{i,3-\ell^*}^{(0)} \left( 1 + \tfrac{2}{\log^2(m)} \right) \right\} \tag{E.2}$$

which only depends on $\Lambda_{i,\ell^*}^{(0)}$ which in terms depends on $G$'s random initialization. (Note $\mathcal{M}_G$ is provably a subset of $\mathcal{M}$ defined in (B.2).)

As for $F$, we break its update into two stages.

1. (Learn.) In the first stage, in the same way as Theorem 1, we start from random initialization and update

$$w_{i,r}^{(t+1)} = w_{i,r}^{(t)} - \eta \nabla_{w_{i,r}} L(F^{(t)}) \text{ for } T = \frac{\mathsf{poly}(k)}{\eta} \text{ iterations}$$

We let $\mathcal{M}_F$ be the "lottery winning" set of network $F$ at the end of stage 1 defined in the same way as (E.2) (which now depends only on $F$'s random initialization).

2. (Distill.) In the second stage, for another $T' = \frac{\mathsf{poly}(k)}{\eta}$ iterations, we update

$$w_{i,r}^{(t+1)} = w_{i,r}^{(t)} - \eta \mathop{\mathbb{E}}_{(X,y) \sim \mathcal{Z}} \left( (\mathbf{logit}_i^\tau(F, X) - \mathbf{logit}_i^\tau(G, X))^- \nabla_{w_{i,r}} F_i^{(t)}(X) \right) \quad \text{(4.4) restated}$$

**Theorem 4** (self distillation, restated)**.** *Suppose the data satisfies Assumption 4.1. For sufficiently large $k > 0$, for every $m \in \left[ \log^{\Omega(1)}(k), k \right]$, every $\eta \leq \frac{1}{\mathsf{poly}(k)}$, setting $T = \frac{\mathsf{poly}(k)}{\eta}$ and $T' = \frac{\mathsf{poly}(k)}{\eta}$, when Parameter C.1 is satisfied, with probability at least $1 - e^{-\Omega(\log^2 k)}$:*

- *(training accuracy is perfect) for every $(X, y) \in \mathcal{Z}$:*

$$\forall i \neq y \colon F_y^{(T+T')}(X) \geq F_i^{(T+T')}(X) + \Omega(\log k).$$

- *(multi-view testing is good) for every $i, j \in [k]$ we have $\widetilde{O}(1) \geq \Phi_i^{(T+T')} \geq 0.4 \Phi_j^{(T+T')} + \Omega(\log k)$, and thus*

$$\mathop{\mathbf{Pr}}_{(X,y) \in \mathcal{D}_m} \left[ F_y^{(T+T')}(X) \geq \max_{j \neq y} F_j^{(T+T')}(X) + \Omega(\log k) \right] \geq 1 - e^{-\Omega(\log^2 k)}$$

- *(single-view testing is better) for every $(i, \ell) \in \mathcal{M}_F \cup \mathcal{M}_G$ we have $\Phi_{i,\ell}^{(T+T')} \geq \Omega\left(\frac{1}{\log k}\right)$, and since $|\mathcal{M}_F \cup \mathcal{M}_G| \geq 1.5k(1 - o(1))$, we have*

$$\mathop{\mathbf{Pr}}_{(X,y) \in \mathcal{D}_s} \left[ F_y^{(T+T')}(X) \geq \max_{j \neq y} F_j^{(T+T')}(X) + \Omega(\log k) \right] \geq \frac{3}{4}(1 - o(1))$$

The proof of Theorem 4 is quite easy once the reader is familiar with the proofs of Theorem 1 and Theorem 3. We include it at the end of the next Section F.

# F  Knowledge Distillation Proof for Ensemble

Our proof structure of Theorem 3 is the same as that for the single model case, but is a lot simpler thanks to our special choice of the truncated distillation function.

Specifically, we maintain the following set of simpler induction hypothesis.

**Induction Hypothesis F.1.** *For every $\ell \in [2]$, for every $r \in [m]$, for every $(X, y) \in \mathcal{Z}$ and $i \in [k]$,*

(a) For every $p \in \mathcal{P}_{v_{i,\ell}}(X)$, we have: $\langle w_{i,r}^{(t)}, x_p \rangle = \langle w_{i,r}^{(t)}, v_{i,\ell} \rangle z_p \pm \widetilde{o}(\sigma_0)$.

(b) For every $p \in \mathcal{P}(X) \setminus \left(\mathcal{P}_{v_{i,1}}(X) \cup \mathcal{P}_{v_{i,2}}(X)\right)$, we have: $|\langle w_{i,r}^{(t)}, x_p \rangle| \leq \widetilde{O}(\sigma_0)$.

(c) For every $p \in [P] \setminus \mathcal{P}(X)$, we have: $|\langle w_{i,r}^{(t)}, x_p \rangle| \leq \widetilde{O}(\sigma_0 \gamma k)$.

Moreover, we have for every $i \in [k]$, every $\ell \in [2]$,

(g) $\Phi_{i,\ell}^{(t)} \geq \Omega(\sigma_0)$ and $\Phi_{i,\ell}^{(t)} \leq \widetilde{O}(1)$.

(h) for every $r \in [m]$, it holds that $\langle w_{i,r}^{(t)}, v_{i,\ell} \rangle \geq -\widetilde{O}(\sigma_0)$.

(Recall $\Phi_{i,\ell}^{(t)} \stackrel{\text{def}}{=} \sum_{r \in [m]} [\langle w_{i,r}^{(t)}, v_{i,\ell} \rangle]^+$.)

**Parameter F.2.** The parameter range for our proofs in this section to hold is the same as Parameter C.1, except that

- $N \geq \eta T \cdot \text{poly}(k)$ and $\eta T \geq \text{poly}(k)$. (Instead of $\eta T \geq N \cdot \text{poly}(k)$.)

  Explanation: single models need a longer training time because they need to memorize single-view data; instead, here ensemble distillation can truly learn all the training data so the training time $T$ can be shorter.[19]

- $m = \text{polylog}(k)$.

  Explanation: we do not need the model to have too much over-parameterization.

- $\eta' = \eta \text{poly}(k)$.

> **Theorem F.3.** *Under Parameter F.2, for any $m = \text{polylog}(k)$ and sufficiently small $\eta \leq \frac{1}{\text{poly}(k)}$ and $\eta' = \eta \text{poly}(k)$, our Induction Hypothesis F.1 holds for all iterations $t = 0, 1, \ldots, T$.*

This entire section is denoted to proving Theorem F.3, and we shall explain in the end of this section how Theorem F.3 implies Theorem 3.

**Disclaimer.** To make this paper more concise, in the rest of this section we highlight the key technical claims/lemmas that we need to prove Induction Hypothesis F.1. Some of the proofs we give in this section are more "sketched" because we assume the readers are already familiar with our proof languages used in Section C.

### F.1 Gradient Calculations and Function Approximation

**Claim F.4.** *There exists some parameter $\Xi = \text{polylog}(k)$ in (E.1) so that for every $(X, y) \in \mathcal{Z}_m$,*

$$\forall i \in [k]: \quad \mathbf{logit}_i^\tau(G, X) = \begin{cases} \frac{1}{s(X)} - k^{-\Omega(\log k)}, & \text{if } v_{i,1} \text{ or } v_{i,2} \text{ is in } \mathcal{V}(X); \\ k^{-\Omega(\log k)}, & \text{if neither } v_{i,1} \text{ nor } v_{i,2} \text{ is in } \mathcal{V}(X); \end{cases}$$

*where $s(X)$ is the number of indices $i \in [k]$ such that $v_{i,1}$ or $v_{i,2}$ is in $\mathcal{V}(X)$. (Recall with high probability $s(X) = \Theta(s)$.) And, for every $(X, y) \in \mathcal{Z}_s$,*

$$\forall i \in [k]: \quad \mathbf{logit}_i^\tau(G, X) = \begin{cases} 1 - k^{-\Omega(\log k)}, & \text{if } i = y; \\ k^{-\Omega(\log k)}, & \text{if } i \neq y; \end{cases}$$

---

[19] Our result also holds for longer $T$, at the expense of adding an additional lemma. We choose to assume $T \geq N \cdot \text{poly}(k)$ for simplicity.

*Proof of Claim F.4.* Using the same analysis as the proof of Theorem 2, we know after ensemble, for every $i \in [k]$ and $\ell \in [2]$,

$$\frac{1}{K} \sum_{w \in [K]} \Phi_{i,\ell}^{[w]} \geq \frac{1}{\mathsf{polylog}(k)}$$

and therefore there exists some scale-up factor $\Xi = \mathsf{polylog}(k)$ for (E.1) so that for every $(X, y) \in \mathcal{Z}_m$, every $i \in [k]$,

- $G_i(X) \geq \log^4 k$ when either $v_{i,1}$ or $v_{i,2}$ is in $\mathcal{V}(X)$;
- $G_i(X) \leq 1$ when neither $v_{i,1}$ nor $v_{i,2}$ is in $\mathcal{V}(X)$.

and at the same time, for every $(X, y) \in \mathcal{Z}_s$, every $i \in [k]$,

- $G_i(X) \geq \log^4 k$ when $i = y$;
- $G_i(X) \leq 1$ when $i \neq y$.

Plugging this into the threshold logit function (4.2) finishes the proof. □

Note that our update rule (4.3) is not precisely the gradient of a function (due to our truncation to the negative part for simpler analysis). However, in the remainder of the proof, slightly abusing notation, let us denote by

$$\nabla_{w_{i,r}} \widetilde{L}(F; X, y) \stackrel{\text{def}}{=} \nabla_{w_{i,r}} L(F; X, y) - \frac{\eta'}{\eta} \left( \mathbf{logit}_i^\tau(F, X) - \mathbf{logit}_i^\tau(G, X) \right)^- \nabla_{w_{i,r}} F_i(X)$$

so that when

**Fact F.5.** *Given data point $(X, y) \in \mathcal{D}$, for every $i \in [k]$, $r \in [m]$, up to a negligible additive error $\frac{1}{k^{\Omega(\log k)}}$, we have*

$$-\nabla_{w_{i,r}} \widetilde{L}(F; X, y) = \left( \mathbb{1}_{y=i} - \mathbf{logit}_i(F, X) + \frac{\eta'}{\eta} \mathbb{1}_{v_{i,1}, v_{i,2} \in \mathcal{V}(X)} \left( \frac{1}{s(X)} - \mathbf{logit}_i^\tau(F, X) \right)^+ \right) \nabla_{w_{i,r}} F_i(X)$$

when $(X, y) \in \mathcal{Z}_m$

$$-\nabla_{w_{i,r}} \widetilde{L}(F; X, y) = \left( \mathbb{1}_{y=i} - \mathbf{logit}_i(F, X) + \frac{\eta'}{\eta} \mathbb{1}_{y=i} \left( 1 - \mathbf{logit}_i^\tau(F, X) \right)^+ \right) \nabla_{w_{i,r}} F_i(X)$$

when $(X, y) \in \mathcal{Z}_s$

*where recall $\nabla_{w_{i,r}} F_i(X) = \sum_{p \in [P]} \widetilde{\mathsf{ReLU}}'(\langle w_{i,r}, x_p \rangle) x_p$*

Because $\frac{1}{k^{\Omega(\log k)}}$ is negligible, for proof simplicity, we ignore it in the rest of the proof.

We also summarize a simple calculation that is analogous to Claim C.7 and Claim C.8.

**Claim F.6** (gradient, c.f. Claim C.7 and C.8). *For every $t \leq T$, for every $(X, y) \in \mathcal{Z}$, every $i \in [k]$, $r \in [m]$ and $\ell \in [2]$, we have:*

- *If $v_{i,1}, v_{i,2} \in \mathcal{V}(X)$ then $\left\langle \nabla_{w_{i,r}} F_i^{(t)}(X), v_{i,\ell} \right\rangle \geq \left( V_{i,r,\ell}(X) - \widetilde{O}(\sigma_p P) \right)$*
- *$\left\langle \nabla_{w_{i,r}} F_i^{(t)}(X), v_{i,\ell} \right\rangle \leq \left( \mathbb{1}_{v_{i,\ell} \in \mathcal{V}(X)} V_{i,r,\ell}(X) + \mathcal{E}_1 + \mathcal{E}_3 \right)$*
- *for every $j \in [k] \setminus \{i\}$, $\left| \left\langle -\nabla_{w_{i,r}} F_i^{(t)}(X), v_{j,\ell} \right\rangle \right| \leq (\mathcal{E}_{2,i,r}(X) + \mathcal{E}_1 + \mathcal{E}_3 + \mathcal{E}_{4,j,\ell}(X))$*

**Notation.** Throughout the remainder of the proof, let us use $v_{i,1}, v_{i,2} \in \mathcal{V}(X)$ to denote that at least one of $v_{i,1}, v_{i,2}$ is in $\mathcal{V}$. This simplifies our notations.

Recall
$$\Phi_{i,\ell}^{(t)} \stackrel{\text{def}}{=} \sum_{r \in [m]} [\langle w_{i,r}^{(t)}, v_{i,\ell} \rangle]^+ \quad \text{and} \quad \Phi_i^{(t)} \stackrel{\text{def}}{=} \sum_{\ell \in [2]} \Phi_{i,\ell}^{(t)}$$

and this time we have

**Claim F.7** (function approximation, c.f. Claim C.9). *Under the new Induction Hypothesis F.1, let us define $Z_{i,\ell}^{(t)}(X) := \mathbb{1}_{v_{i,\ell} \in \mathcal{V}(X)} \left( \sum_{p \in \mathcal{P}_{v_{i,\ell}}(X)} z_p \right)$, we have: for every $t$, every $i \in [k]$, every $(X, y) \in \mathcal{Z}$ (or for every new sample $(X, y) \sim \mathcal{D}$, with probability at least $1 - e^{-\Omega(\log^2 k)}$):*
$$F_i^{(t)}(X) = \sum_{\ell \in [2]} \left( \Phi_{i,\ell}^{(t)} \times Z_{i,\ell}^{(t)}(X) \right) \pm \widetilde{O}(\sigma_0 \cdot m) = \sum_{\ell \in [2]} \left( \Phi_{i,\ell}^{(t)} \times Z_{i,\ell}^{(t)}(X) \right) \pm O\left(\frac{1}{\mathsf{polylog}(k)}\right)$$

## F.2 Useful Claims as Consequences of the Induction Hypothesis

Recall we had five useful claims in Section C.2 for the proof of the single model case. This time, we only have three and they are also easier than their counterparts in Section C.2.

### F.2.1 Lambda Growth

**Claim F.8** (growth, c.f. Claim C.10). *Suppose Induction Hypothesis F.1 holds at iteration $t$, then for every $i \in [k]$, $\ell \in [2]$, suppose $\Phi_{i,\ell}^{(t)} \leq \frac{1}{\tau}$, then it satisfies*
$$\Phi_{i,\ell}^{(t+1)} \geq \Phi_{i,\ell}^{(t)} + \widetilde{\Omega}\left(\frac{\eta'}{k}\right) \widetilde{\mathsf{ReLU}}'(\Phi_{i,\ell}^{(t)})$$

*Proof of Claim C.10.* Using the same calculation as (C.5), but this time substituting the new gradient formula in Fact F.5, we have
$$\langle w_{i,r}^{(t+1)}, v_{i,\ell} \rangle \geq \langle w_{i,r}^{(t)}, v_{i,\ell} \rangle - O(\eta + \eta' \frac{N_s}{N})$$
$$+ \Omega(\eta') \mathop{\mathbb{E}}_{(X,y) \sim \mathcal{Z}_m} \left[ \mathbb{1}_{v_{i,1}, v_{i,2} \in \mathcal{V}(X)} \left( V_{i,r,\ell}(X) - \widetilde{O}(\sigma_p P) \right) \left( \frac{1}{s(X)} - \mathbf{logit}_i^\tau(F^{(t)}, X) \right)^+ \right]$$

Let us now consider $r \in [m]$ to be the $\arg\max_{r \in [m]} \{\langle w_{i,r}^{(t)}, v_{i,\ell} \rangle\}$, so we have $\langle w_{i,r}^{(t)}, v_{i,\ell} \rangle \geq \widetilde{\Omega}(\Phi_{i,\ell}^{(t)})$.

Following a similar analysis as before, we can derive that as long as $\mathbb{1}_{v_{i,1}, v_{i,2} \in \mathcal{V}(X)} = 1$, we have $V_{i,r,\ell}(X) \geq \Omega(1) \cdot \widetilde{\mathsf{ReLU}}'\left(\langle w_{i,r}^{(t)}, v_{i,\ell} \rangle\right) \geq \widetilde{\Omega}(\widetilde{\mathsf{ReLU}}'(\Phi_{i,\ell}^{(t)}))$.

Now, since $\Phi_{i,\ell}^{(t)} \leq \frac{1}{\tau}$, we know that as long as $v_{i,\ell} \in \mathcal{V}(X)$ and $v_{i,3-\ell} \notin \mathcal{V}(X)$ (which happens for $\Theta(\frac{s}{k})$ fraction of the multi-view training data), it satisfies (see Claim F.7):
$$F_i^{(t)}(X) \leq \Phi_{i,\ell}^{(t)} \times Z_{i,\ell}^{(t)}(X) - \widetilde{O}(\sigma_0 m) = O\left(\frac{1}{\tau}\right) - \widetilde{O}(\sigma_0 m) \leq O\left(\frac{1}{\tau}\right) .$$

When this happens, we know $\mathbf{logit}_i^\tau(F^{(t)}, X) \leq O(\frac{1}{k})$. This implies, after summing over $r \in [m]$,
$$\Phi_{i,\ell}^{(t+1)} \geq \Phi_{i,\ell}^{(t)} + \widetilde{\Omega}\left(\frac{\eta'}{k}\right) \widetilde{\mathsf{ReLU}}'(\Phi_{i,\ell}^{(t)})$$
□

Now we can define $T_0$ as follows.

**Claim F.9.** *Define iteration threshold $T_0 \stackrel{\text{def}}{=} \widetilde{\Theta}\left(\frac{k}{\eta' \sigma_0^{q-2}}\right)$, then*

- *for every $i \in [k], \ell \in [2]$ and $t \geq T_0$, it satisfies $\Phi_{i,\ell}^{(t)} \geq \frac{1}{2\tau}$*

### F.2.2 Single-View Error Till the End

**Claim F.10** (single-view after $T_0$). *Suppose Induction Hypothesis F.1 holds for all iterations $< t$ and $t \geq T_0$. For every single-view data $(X, y) \in \mathcal{Z}_s$ (or any $(X, y) \in \mathcal{D}_s$ but with probability $1 - e^{-\Omega(\log^2 k)}$), we have*

$$F_y^{(t)}(X) \geq \max_{i \in [k] \setminus \{y\}} F_i^{(t)}(X) + \Omega(\log k) \quad \text{and} \quad 1 - \mathbf{logit}_y(F^{(t)}, X) \leq \frac{1}{\mathsf{poly}(k)}$$

*Proof.* This is because for single-view data $(X, y) \in \mathcal{Z}_s$, it satisfies $Z_{i,\ell}^{(t)}(X) \leq \Gamma$ as long as $i \neq y$. As a result, applying Claim F.9, we must have $F_y^{(t)}(X) \geq \Omega(\frac{1}{\tau}) \geq \Omega(\log k)$ but $F_i^{(t)}(X) \leq O(1)$ for $i \neq y$ (using $\Gamma < \frac{1}{\mathsf{polylog}(k)}$). (Similar for $(X, y) \in \mathcal{D}_s$.) $\square$

### F.2.3 Multi-View Error Till the End

**Claim F.11** (multi till the end, c.f. Claim C.14). *Suppose Induction Hypothesis F.1 holds for every iteration $t < T$, then*

- $\eta \sum_{t=T_0}^{T} \mathbb{E}_{(X,y) \sim \mathcal{Z}_m} \left[ \mathbb{1}_{y=i} \left( 1 - \mathbf{logit}_y(F^{(t)}, X) \right) \right] \leq \widetilde{O}(1)$
- $\frac{\eta' N_s}{N} \sum_{t=T_0}^{T} \mathbb{E}_{(X,y) \sim \mathcal{Z}_s} \left[ \mathbb{1}_{y=i} \left( 1 - \mathbf{logit}_i^\tau(F^{(t)}, X) \right)^+ \right] \leq \widetilde{O}(1)$
- $\eta' \sum_{t=T_0}^{T} \mathbb{E}_{(X,y) \sim \mathcal{Z}_m} \left[ \mathbb{1}_{v_{i,1}, v_{i,2} \in \mathcal{V}(X)} \left( \frac{1}{s(X)} - \mathbf{logit}_i^\tau(F^{(t)}, X) \right)^+ \right] \leq \widetilde{O}(1)$

*Proof of Claim F.11.* By Fact F.5 and Claim F.6 again (similar to the calculation in the proof of Claim F.8), we have

$$\langle w_{i,r}^{(t+1)}, v_{i,\ell} \rangle \geq \langle w_{i,r}^{(t)}, v_{i,\ell} \rangle - \frac{1}{\mathsf{poly}(k)}$$
$$+ \Omega(\eta) \mathop{\mathbb{E}}_{(X,y) \sim \mathcal{Z}_m} \left[ \mathbb{1}_{y=i} \left( V_{i,r,\ell}(X) - \widetilde{O}(\sigma_p P) \right) \left( 1 - \mathbf{logit}_i(F^{(t)}, X) \right) \right]$$
$$+ \Omega(\frac{\eta' N_s}{N}) \mathop{\mathbb{E}}_{(X,y) \sim \mathcal{Z}_s} \left[ \mathbb{1}_{y=i} \left( V_{i,r,\ell}(X) - \widetilde{O}(\sigma_p P) \right) \left( 1 - \mathbf{logit}_i^\tau(F^{(t)}, X) \right)^+ \right]$$
$$+ \Omega(\eta') \mathop{\mathbb{E}}_{(X,y) \sim \mathcal{Z}_m} \left[ \mathbb{1}_{v_{i,1}, v_{i,2} \in \mathcal{V}(X)} \left( V_{i,r,\ell}(X) - \widetilde{O}(\sigma_p P) \right) \left( \frac{1}{s(X)} - \mathbf{logit}_i^\tau(F^{(t)}, X) \right)^+ \right]$$

In the above formula, we can ignore the single-view data for the $\left( 1 - \mathbf{logit}_i(F^{(t)}, X) \right)$ term because they are extremely small (see Claim F.10).

Now, if we take $r = \arg\max_{r \in [m]} \{ \langle w_{i,r}^{(t)}, v_{i,\ell} \rangle \}$, we must have (whenever $v_{i,1}, v_{i,2} \in \mathcal{V}(X)$) $V_{i,r,\ell}(X) \geq \Omega(1) \cdot \widetilde{\mathsf{ReLU}}' \left( \langle w_{i,r}^{(t)}, v_{i,\ell} \rangle \right) \geq \widetilde{\Omega}(1)$. Therefore, when summing up over all possible $r \in [m]$, we have

$$\Phi_{i,\ell}^{(t+1)} \geq \Phi_{i,\ell}^{(t)} - \frac{\eta}{\mathsf{poly}(k)}$$
$$+ \widetilde{\Omega}(\eta) \mathop{\mathbb{E}}_{(X,y) \sim \mathcal{Z}_m} \left[ \mathbb{1}_{y=i} \left( 1 - \mathbf{logit}_i(F^{(t)}, X) \right) \right]$$
$$+ \widetilde{\Omega}(\frac{\eta' N_s}{N}) \mathop{\mathbb{E}}_{(X,y) \sim \mathcal{Z}_s} \left[ \mathbb{1}_{y=i} \left( 1 - \mathbf{logit}_i^\tau(F^{(t)}, X) \right)^+ \right]$$
$$+ \widetilde{\Omega}(\eta') \mathop{\mathbb{E}}_{(X,y) \sim \mathcal{Z}_m} \left[ \mathbb{1}_{v_{i,1}, v_{i,2} \in \mathcal{V}(X)} \left( \frac{1}{s(X)} - \mathbf{logit}_i^\tau(F^{(t)}, X) \right)^+ \right]$$

After telescoping, and using $\Phi^{(t)} \leq \widetilde{O}(1)$, we finish the proof. $\square$

## F.3 Main Lemmas for Proving the Induction Hypothesis

In this subsection, we provide key technical lemmas that, when combined together, shall prove that Induction Hypothesis F.1 holds for every iteration (and thus prove Theorem F.3).[20]

### F.3.1 Correlation Growth

**Lemma F.12** (c.f. Lemma C.21). *Suppose Parameter F.2 holds and suppose Induction Hypothesis F.1 holds for all iterations $< t$. Then, letting $\Phi_{i,\ell}^{(t)} \stackrel{\text{def}}{=} \sum_{r \in [m]} [\langle w_{i,r}^{(t)}, v_{i,\ell} \rangle]^+$, we have*

$$\forall i \in [k], \forall \ell \in [2]: \qquad \Phi_{i,\ell}^{(t)} \leq \widetilde{O}(1)$$

*Proof of Lemma F.12.* Let us denote by $\Phi^{(t)} = \max_{i \in [k], \ell \in [2]} \Phi_{i,\ell}^{(t)}$. Suppose $t$ is some iteration so that $\Phi^{(t)} \geq \frac{10}{\tau^2}$ but $\Phi^{(t)} \leq \widetilde{O}(1)$. We wish to prove that if we continue from iteration $t$ for at most $T$ iterations, then $\Phi^{(t')} \leq \widetilde{O}(1)$ for every $t' \in [t, T]$.

Without loss of generality, we assume that $\Phi^{(t)} \geq \frac{10}{\tau^2}$ always holds from iteration $t$ onwards (because otherwise we can start with the next iteration $t'$ so that $\Phi^{(t)}$ goes above $\frac{10}{\tau^2}$.)

Let $(i, \ell) = \arg\max_{i \in [k], \ell \in [2]} \Phi_{i,\ell}^{(t)}$. Then, for every $(X, y) \in \mathcal{Z}_m$, (F.7) tells us that $F_i^{(t)}(X) \geq \frac{1}{\tau^2}$ for every $i$ such that $v_{i,1}, v_{i,2} \in \mathcal{V}(X)$. Therefore,

$$\mathbb{1}_{v_{i,1}, v_{i,2} \in \mathcal{V}(X)} \left( \frac{1}{s(X)} - \mathbf{logit}_i^\tau(F, X) \right)^+ \leq \frac{1}{k^{\Omega(\log k)}}$$

Also, for every $(X, y) \in \mathcal{Z}_s$, for similar reason we have

$$(1 - \mathbf{logit}_i^\tau(F, X))^+ \leq \frac{1}{k^{\Omega(\log k)}}$$

Therefore, at this iteration $t$, up to negligible $\frac{1}{k^{\Omega(\log k)}}$ terms, we have according to Claim F.7:

$$-\nabla_{w_{i,r}} \widetilde{L}(F; X, y) = (1 - \mathbf{logit}_i(F, X)) \nabla_{w_{i,r}} F_i(X) \qquad \text{when } i = y$$
$$-\nabla_{w_{i,r}} \widetilde{L}(F; X, y) = -\mathbf{logit}_i(F, X) \nabla_{w_{i,r}} F_i(X) \qquad \text{when } i \neq y$$

This is already identical to what we had in the single-model case (without distillation).

This time, we can calculate (using Claim F.6)

$$\langle w_{i,r}^{(t+1)}, v_{i,\ell} \rangle = \langle w_{i,r}^{(t)}, v_{i,\ell} \rangle + \eta \mathop{\mathbb{E}}_{(X,y) \sim \mathcal{Z}} \left[ \langle -\nabla_{w_{i,r}} \widetilde{L}(F^{(t)}; X, y), v_{i,\ell} \rangle \right]$$
$$\leq \langle w_{i,r}^{(t)}, v_{i,\ell} \rangle + \eta \mathop{\mathbb{E}}_{(X,y) \sim \mathcal{Z}} \left[ \mathbb{1}_{y=i} (V_{i,r,\ell}(X) + \mathcal{E}_1 + \mathcal{E}_3)(1 - \mathbf{logit}_i(F, X)) \right]$$
$$\leq \langle w_{i,r}^{(t)}, v_{i,\ell} \rangle + O(\eta) \mathop{\mathbb{E}}_{(X,y) \sim \mathcal{Z}} \left[ \mathbb{1}_{y=i} (1 - \mathbf{logit}_y(F, X)) \right]$$

- For every $(X, y) \in \mathcal{Z}_s$, since $\Phi_{i,\ell}^{(t)} \geq \frac{10}{\tau^2}$, it satisfies (similar to Claim F.10) that $F_y^{(t)}(X) \geq \Omega(\frac{1}{\tau^2}) \gg \Omega(\log^2 k)$ but $F_j^{(t)}(X) \leq O(1)$ for $j \neq y$ (using $\Gamma < \frac{1}{\mathsf{polylog}(k)}$). This implies

$$(1 - \mathbf{logit}_y(F, X)) \leq \frac{1}{k^{\Omega(\log k)}}$$

- For every $(X, y) \in \mathcal{Z}_m$ with $y = i$, recall from Claim F.7 that

$$F_j^{(t)}(X) = \sum_{\ell \in [2]} \left( \Phi_{j,\ell}^{(t)} \times \mathbb{1}_{v_{j,\ell} \in \mathcal{V}(X)} \left( \sum_{p \in \mathcal{P}_{v_{j,\ell}}(X)} z_p \right) \right) \pm O(\tfrac{1}{\mathsf{polylog}(k)})$$

---
[20]We only sketch the proofs to these technical lemmas, but ignore the last step of putting them together to prove Theorem F.3 because it is trivial (but anyways almost identical to that in Section C.5).

By our choice of the distribution, this implies

- $F_j^{(t)}(X) \leq 0.8001 \Phi^{(t)}$ for $j \neq i$, and
- $F_i^{(t)}(X) \geq 0.9999 \Phi^{(t)}$ because $(i, \ell)$ is the argmax of $\Phi_{i,\ell}^{(t)}$.

This again means
$$\left(1 - \mathbf{logit}_y(F, X)\right) \leq \frac{1}{k^{\Omega(\log k)}}$$

Together, and summing up over all $r \in [m]$, we have
$$\Phi^{(t+1)} \leq \Phi^{(t)} + \frac{\eta m}{k^{\Omega(\log k)}}$$

so if we continue this for $T$ iterations we still have $\Phi^{(T)} \leq \widetilde{O}(1)$. $\square$

### F.3.2 Off-Diagonal Correlations are Small

**Lemma F.13** (c.f. Lemma C.22). *Suppose Parameter F.2 holds and suppose Induction Hypothesis F.1 holds for all iterations $< t$. Then,*
$$\forall i \in [k] \ , \ \forall r \in [m] \ , \ \forall j \in [k] \setminus \{i\}: \qquad |\langle w_{i,r}^{(t)}, v_{j,\ell} \rangle| \leq \widetilde{O}(\sigma_0)$$

*Proof of Lemma F.13.* We separately treat $t \leq T_0$ and $t \geq T_0$. (This should be reminiscent of the three-stage proof in the single model case.)

**Consider $t \leq T_0$.** By Fact F.5 and Claim F.6
$$|\langle w_{i,r}^{(t+1)}, v_{j,\ell} \rangle| \leq |\langle w_{i,r}^{(t)}, v_{j,\ell} \rangle| + O(\eta + \frac{\eta' N_s}{N})$$
$$+ O(\eta') \mathop{\mathbb{E}}_{(X,y) \sim \mathcal{Z}_m} \left[ \left( \mathcal{E}_{2,i,r}(X) + \mathcal{E}_1 + \mathcal{E}_3 + \mathcal{E}_{4,j,\ell}(X) \right) \mathbb{1}_{v_{i,1}, v_{i,2} \in \mathcal{V}(X)} \left( \frac{1}{s(X)} - \mathbf{logit}_i^\tau(F, X) \right)^+ \right]$$

Using the property that $v_{i,1}, v_{i,2} \in \mathcal{V}(X)$ with probability $\Theta(\frac{s}{k})$ over a sample $(X, y) \in \mathcal{Z}_m$, and using the trivial bound $\left( \frac{1}{s(X)} - \mathbf{logit}_i^\tau(F, X) \right)^+ \leq \Theta(\frac{1}{s})$, we immediately have
$$|\langle w_{i,r}^{(t+1)}, v_{j,\ell} \rangle| \leq |\langle w_{i,r}^{(t)}, v_{j,\ell} \rangle| + O(\eta + \frac{\eta' N_s}{N})$$
$$+ O\left(\frac{\eta'}{k}\right) \left( \gamma + (\sigma_0^{q-1}) \gamma s + \widetilde{O}\left( (\sigma_0 \gamma k)^{q-1} \right) \gamma P + (\sigma_0)^{q-1} \frac{s}{k} \right)$$

Finally, using $T_0 = \widetilde{\Theta}\left(\frac{k}{\eta' \sigma_0^{q-2}}\right)$, $N_s \leq \frac{N}{\mathsf{poly}(k)}$, and using $\eta \leq \frac{\eta'}{\mathsf{poly}(k)}$, together with the same parameter choices as before, we conclude that $|\langle w_{i,r}^{(t+1)}, v_{j,\ell} \rangle| \leq \widetilde{O}(\sigma_0)$ for every $t \leq T_0$.

**Consider $t > T_0$.** By Fact F.5 and Claim F.6 again
$$|\langle w_{i,r}^{(t+1)}, v_{j,\ell} \rangle| \leq |\langle w_{i,r}^{(t)}, v_{j,\ell} \rangle| + O(\frac{\eta}{\mathsf{poly}(k)})$$
$$+ O(\eta) \mathop{\mathbb{E}}_{(X,y) \sim \mathcal{Z}_m} \left[ \left( \mathcal{E}_{2,i,r}(X) + \mathcal{E}_1 + \mathcal{E}_3 + \mathcal{E}_{4,j,\ell}(X) \right) \mathbb{1}_{y=i} \left(1 - \mathbf{logit}_i(F, X)\right) \right]$$
$$+ O(\eta) \mathop{\mathbb{E}}_{(X,y) \sim \mathcal{Z}_m} \left[ \left( \mathcal{E}_{2,i,r}(X) + \mathcal{E}_1 + \mathcal{E}_3 + \mathcal{E}_{4,j,\ell}(X) \right) \mathbb{1}_{y \neq i} \left( \mathbf{logit}_i(F, X) \right) \right]$$
$$+ O(\frac{\eta' N_s}{N}) \mathop{\mathbb{E}}_{(X,y) \sim \mathcal{Z}_s} \left[ \left( \mathcal{E}_{2,i,r}(X) + \mathcal{E}_1 + \mathcal{E}_3 + \mathcal{E}_{4,j,\ell}(X) \right) \mathbb{1}_{y=i} \left(1 - \mathbf{logit}_i^\tau(F, X)\right)^+ \right]$$

$$+ O(\eta') \mathop{\mathbb{E}}_{(X,y)\sim\mathcal{Z}_m} \left[ \left( \mathcal{E}_{2,i,r}(X) + \mathcal{E}_1 + \mathcal{E}_3 + \mathcal{E}_{4,j,\ell}(X) \right) \mathbb{1}_{v_{i,1}, v_{i,2} \in \mathcal{V}(X)} \left( \frac{1}{s(X)} - \mathbf{logit}_i^\tau(F, X) \right)^+ \right]$$

In the above formula, we can ignore the single-view data for the $\left(1 - \mathbf{logit}_y(F^{(t)}, X)\right)$ and $\mathbf{logit}_i(F^{(t)}, X)$ (for $i \neq y$) terms because they are extremely small (see Claim F.10).

Now, applying the naive upper bounds $\mathcal{E}_{2,i,r}(X) \leq \gamma$ and $\mathcal{E}_{4,j,\ell}(X) \leq \widetilde{O}(\sigma_0^{q-1})$, and telescoping for all $t \geq T_0$ and applying Claim F.11, we immediately have $|\langle w_{i,r}^{(t)}, v_{j,\ell}\rangle| \leq |\langle w_{i,r}^{(T_0)}, v_{j,\ell}\rangle| + \widetilde{O}(\sigma_0)$. □

### F.3.3 Noise Correlation is Small

**Lemma F.14** (c.f. Lemma C.25). *Suppose Parameter F.2 holds and suppose Induction Hypothesis F.1 holds for all iterations $< t$. For every $\ell \in [2]$, for every $r \in [m]$, for every $(X,y) \in \mathcal{Z}$ and $i \in [k]$:*

(a) *For every $p \in \mathcal{P}_{v_{i,\ell}}(X)$, we have: $\langle w_{i,r}^{(t)}, \xi_p\rangle \leq \widetilde{o}(\sigma_0)$.*

(b) *For every $p \in \mathcal{P}(X) \setminus \left( \mathcal{P}_{v_{i,1}}(X) \cup \mathcal{P}_{v_{i,2}}(X) \right)$, we have: $|\langle w_{i,r}^{(t)}, \xi_p\rangle| \leq \widetilde{O}(\sigma_0)$.*

(c) *For every $p \in [P] \setminus \mathcal{P}(X)$, we have: $|\langle w_{i,r}^{(t)}, \xi_p\rangle| \leq \widetilde{O}(\sigma_0 \gamma k)$.*

*Proof.* From a similar calculation (see (C.6)) we have for every $(X,y) \in \mathcal{Z}$ and $p \in [P]$,

$$\langle w_{i,r}^{(t+1)}, \xi_p\rangle = \langle w_{i,r}^{(t)}, \xi_p\rangle \pm \widetilde{\Theta}\left(\frac{\eta'}{N}\right) \widetilde{\mathsf{ReLU}}'(\langle w_{i,r}^{(t)}, x_p\rangle) \pm \frac{\eta}{\sqrt{d}}$$

After telescoping and using $N \gg T$ from Parameter F.2, we immediately finish the proof. (If one instead wishes to consider the case of $T \geq N$, she has to do a more careful calculation here. We skip it to keep this paper concise. □

### F.3.4 Diagonal Correlations are Nearly Non-Negative

**Lemma F.15** (c.f. Lemma C.27). *Suppose Parameter F.2 holds and suppose Induction Hypothesis F.1 holds for all iterations $< t$. Then,*

$$\forall i \in [k], \forall r \in [m], \forall \ell \in [2]: \quad \langle w_{i,r}^{(t)}, v_{i,\ell}\rangle \geq -\widetilde{O}(\sigma_0) \ .$$

The proof of Lemma F.15 is almost identical to Lemma C.27 so we skip here.

## F.4 Proof of Theorem 3

First of all, applying Claim F.11 and $T \geq \frac{\mathsf{poly}(k)}{\eta}$, we know there are at most 90% of the iterations $t \leq T_0$ satisfying

$$\mathop{\mathbb{E}}_{(X,y)\sim\mathcal{Z}_m} \left[ \left(1 - \mathbf{logit}_y(F^{(t)}, X)\right) \right] \leq \frac{1}{\mathsf{poly}(k)}$$

Applying Claim C.16, we immediately have the test accuracy result for multi-view data.

Applying Claim F.10 (which uses $\Phi_{i,\ell}^{(t)} \geq \Omega(\log k)$), we immediately have the test accuracy result for single-view data. ∎

## F.5 Proof of Theorem 4

We assume the readers are now familiar with the proofs of the single model Theorem 1 and of the ensemble distill Theorem 3. They easily imply Theorem 4 for reasons we explain below.

**Stage 1 of $F$.** Recall Theorem 1 (and Lemma C.21) imply that, at the end of the stage 1 for training a network $F$, the quantity $\Phi_i^{(T)} \in [\Omega(\log k), \widetilde{O}(1)]$ for every $i \in [k]$; thus, if $(i, \ell) \in \mathcal{M}$, we must have $\Phi_{i,\ell}^{(T)} \geq \Omega(\log k)$. At the end of stage 1, also recall for every $(i, \ell) \in \mathcal{M}$, for any single-view data $(X, y) \in \mathcal{D}_s$ with $y = i$ and $\widehat{\ell}(X) = \ell$, with high probability $F$ predicts correctly on $(X, y)$. Let us remind the readers from (B.2) that

$$\mathcal{M} \stackrel{\text{def}}{=} \left\{ (i, \ell^*) \in [k] \times [2] \,\Big|\, \Lambda_{i,\ell^*}^{(0)} \geq \Lambda_{i,3-\ell^*}^{(0)} \left(\frac{S_{i,3-\ell^*}}{S_{i,\ell^*}}\right)^{\frac{1}{q-2}} \left(1 + \frac{1}{\log^2(m)}\right) \right\}$$

where $S_{i,\ell} \stackrel{\text{def}}{=} \mathbb{E}_{(X,y)\sim \mathcal{Z}_m}\left[\mathbb{1}_{y=i} \sum_{p \in P_{v_{i,\ell}}(X)} z_p^q\right]$. Since in this self-distillation theorem, we have Assumption 4.1 which says the distribution of $\sum_{p \in \mathcal{P}_v(X)} z_p^q$ for $v \in \{v_{y,1}, v_{y,2}\}$ are the same over multi-view data, by standard concentration, we know with high probability $S_{i,1} = S_{i,2}\left(1 \pm \frac{1}{2 \log^2 k}\right)$ for every $i \in [k]$. This means, we can alternatively define

$$\mathcal{M}_F \stackrel{\text{def}}{=} \left\{ (i, \ell^*) \in [k] \times [2] \,\Big|\, \Lambda_{i,\ell^*}^{(0)} \geq \Lambda_{i,3-\ell^*}^{(0)} \left(1 + \frac{2}{\log^2(m)}\right) \right\}$$

(which is a subset of $\mathcal{M}$) and all the statements about $\mathcal{M}$ also apply to $\mathcal{M}_F$.

**Base Model $G$.** For a similar reason, if $(i, \ell) \in \mathcal{M}_G$ for the distill model $G$, then we have that the quantity $\Phi_{i,\ell}^{(T)} \geq \text{polylog}(k)$ for network $G$ (recall $G$ is scaled up by a $\text{polylog}(k)$ factor). Using a similar analysis to Claim F.4, we can derive that:

- for every $(X, y) \in \mathcal{Z}_m$, we have

$$\forall (i, \ell) \in \mathcal{M}_G: \quad \mathbf{logit}_i^\tau(G, X) \begin{cases} \geq \frac{1}{s(X)} - k^{-\Omega(\log k)}, & \text{if } v_{i,\ell} \text{ is in } \mathcal{V}(X); \\ = k^{-\Omega(\log k)}, & \text{if } v_{i,\ell} \text{ is not in } \mathcal{V}(X); \end{cases} \quad (\text{F.1})$$

$$\forall i \in [k]: \quad \mathbf{logit}_i^\tau(G, X) \begin{cases} \leq \frac{1}{s'(X)} + k^{-\Omega(\log k)}, & \text{if } v_{i,1} \text{ or } v_{i,2} \text{ is in } \mathcal{V}(X); \\ = k^{-\Omega(\log k)}, & \text{if neither } v_{i,1} \text{ or } v_{i,2} \text{ is not in } \mathcal{V}(X); \end{cases} \quad (\text{F.2})$$

where recall $s(X)$ is the number of indices $i \in [k]$ such that $v_{i,1}$ or $v_{i,2}$ is in $\mathcal{V}(X)$, and we newly define $s'(X)$ as the number of indices $i \in [k]$ such that $(i, \ell) \in \mathcal{M}_G$ and $v_{i,\ell} \in \mathcal{V}(X)$ for some $\ell \in [2]$. One can derive using concentration that with high probability $\frac{s}{2} \leq s'(X) \leq s(X) \leq 3s$ for all multi-view training data.[21]

**Stage 2 of $F$.** Similar to the proof of Theorem 3, we can ignore single-view data's contribution to the gradient updates (since they are negligible) and only focus on multi-view data. Using a similar gradient calculation to Fact F.5, we know that, up to some small error,

- the quantity $\Phi_{i,\ell} = \sum_{r \in [m]} \left[\langle w_{i,r}, v_{i,\ell}\rangle\right]^+$ never decreases during stage 2.
- the quantity $\Phi_i = \sum_{r \in [m], \ell \in [2]} \left[\langle w_{i,r}, v_{i,\ell}\rangle\right]^+$ no longer changes during stage 2, when it reaches $\Phi_i \geq \frac{2}{\tau^2}$.[22]

Recall after stage 1, we have $\Phi_i \geq \Omega(\log k)$ for network $F$; but since at the beginning of stage 2 we have scaled up $F$ by a factor of $\log^4 k$, this means $\Phi_i \geq \frac{2}{\tau^2}$ is already satisfied at the beginning of stage 2 (up to small error), so it does not change during stage 2. As a result, at the end of stage 2, network $F$ should give the same (nearly perfect) accuracy on multi-view data as claimed in Theorem 1.

---

[21] We remark here that $\mathcal{M}_G$ does not depend on the randomness of the training set, so the lower bound $\frac{s}{10} \leq s'(X)$ can be derived trivially using $|\mathcal{M}_G| \geq k(1 - o(1))$.

[22] Whenever $\Phi_i \geq \frac{2}{\tau^2}$ for network $F$, one can verify that $F_i(X) \geq \frac{1}{\tau^2}$ for every multi-view data $(X, y) \in \mathcal{Z}_m$ with $v_{i,1}, v_{i,2} \in \mathcal{V}(X)$. This means $\mathbf{logit}_i^\tau(F, X) \geq \frac{1}{s(X)} - k^{-\Omega(\log k)}$ for every multi-view data $(X, y)$ with $v_{i,1}, v_{i,2} \in \mathcal{V}(X)$. When this happens, combining with (F.2), we have $\left[\mathbf{logit}_i^\tau(G, X) - 10\mathbf{logit}_i^\tau(F, X)\right]^+ = 0$ so (up to small error) there is no gradient and $\langle w_{i,r}, v_{i,\ell}\rangle$ stays unchanged.

Furthermore, through a similar analysis to Claim F.8 and Claim F.9 (and combining with (F.1)), we know that for when $(i,\ell) \in \mathcal{M}_G$, the quantity $\Phi_{i,\ell}$ must increase to at least $\frac{1}{2\tau} \geq \Omega(\log^2 k)$. This allows us to conclude that, when $(i,\ell) \in \mathcal{M}_G$, at the end of stage 2, for those single-view data $(X,y) \in \mathcal{D}_s$ with $y = i$ and $\widehat{\ell}(X) = \ell$, with high probability $F_y(X) \geq \Omega(\log^2 k) \gg F_j(X)$ for $j \neq y$, so $F$ predicts correctly on $(X, y)$.

At the same time, for every $(i, \ell) \in \mathcal{M}_F$, we have $\Phi_{i,\ell} \geq \Omega(\log k)$ is already satisfied at the end of stage 1, so at the end of stage 2 it must also satisfy $\Phi_{i,\ell} \geq \Omega(\log^5 k)$ (the extra factors are due to scale-up). Thus, $F$ also predicts correctly on those single-view data $(X,y) \in \mathcal{D}_s$ with $y = i$ and $\widehat{\ell}(X) = \ell$.

Finally, using $|\mathcal{M}_F| \geq k(1 - o(1))$ and $|\mathcal{M}_G| \geq k(1 - o(1))$, together with the fact that they are totally independent random sets, we obtain $|\mathcal{M}_F \cup \mathcal{M}_G| \geq \frac{3}{2}k(1 - o(1))$. This means, learned model $F^{(T+T')}$ at the end of stage 2 through self-distillation has an accuracy of $\geq \frac{3}{4}(1 - o(1))$ over single-view data. ∎

## G  Simple Probability Lemmas

We first state a simple proposition that directly implies Fact B.1.

**Proposition G.1.** *Given $m$ i.i.d. standard Gaussian random variables $g_1, \ldots, g_m \sim \mathcal{N}(0,1)$, with probability at least $1 - \delta$, we have that except for at most $O(\log(1/\delta))$ indices $i \in [m]$, we have*

$$g_i \leq \max_{j \in [m]}\{g_j\} \cdot \left(1 - \Omega\left(\frac{1}{\log(m/\log(1/\delta))}\right)\right)$$

*Proof of Proposition G.1.* Recall for every $x > 0$

$$\left(\frac{1}{x} - \frac{1}{x^3}\right)\frac{e^{-x^2/2}}{\sqrt{2\pi}} \leq \Pr_g[g > x] \leq \frac{1}{x}\frac{e^{-x^2/2}}{\sqrt{2\pi}} \tag{G.1}$$

The probability for one of them to exceed

$$\Pr[\max_i g_i > x] = 1 - (1 - \Pr_g[g > x])^m$$

Let us choose $x^*$ so that $\Pr[\max_i g_i > x] = 1 - \delta/2$. By the asymptotic bound above, it is easy to derive that $x^* = \Theta(\sqrt{\log(m/\log(1/\delta))})$ and $\Pr[g > x^*] = \Theta(\frac{\log(1/\delta)}{m})$.

Now, consider $x = x^* - \frac{1}{x^*} = x^*(1 - \frac{1}{(x^*)^2})$. By the asymptotic bound above, it is not hard to see

$$\Pr[g > x] \leq O(1) \cdot \Pr[g > x^*] \leq O(\frac{\log(1/\delta)}{m})$$

By Chernoff bound, we know with probability at least $1 - \delta$, it satisfies that

$$\Pr\left[\sum_{i=1}^{m}\mathbb{1}_{g > x} \geq \Omega(\log(1/\delta))\right] \leq \delta/2$$

□

We next state a proposition that shall be used to prove Proposition B.2.

**Proposition G.2.** *Consider two sequences of i.i.d. Gaussian, $g_1, \ldots, g_m \sim \mathcal{N}(0,1)$ and $h_1, \ldots, h_m \sim \mathcal{N}(0, \sigma^2)$. Then,*

- when $\sigma > 1$, we have $\mathbf{Pr}\left[\max_{i\in[m]} g_i > \max_{i\in[m]} h_i\right] \geq \Omega(\frac{1}{\sigma})\frac{1}{m^{\sigma^2-1}}$.

- when $\sigma \leq 1$, given $\tau > 0$, we have $\mathbf{Pr}\left[\max_{i\in[m]} g_i = (1 \pm O(\tau))\max_{i\in[m]} h_i\right] \leq O(\tau \log m + \frac{1}{\mathsf{poly}(m)})$.

*Proof of Proposition G.2.* The case of $\sigma \leq 1$ is trivial and is simply by symmetry.

**The case of $\sigma > 1$.** For any threshold $x > 0$, we have
$$\mathbf{Pr}[\max_i g_i > x] = 1 - (1 - \Pr_{g\sim\mathcal{N}(0,1)}[g > x])^m$$
$$\mathbf{Pr}[\max_i h_i < x] = (1 - \Pr_{h\sim\mathcal{N}(0,\sigma^2)}[h > x])^m$$

Let $x^* > \sigma$ be a threshold satisfying $\frac{\sigma}{x^*}\frac{e^{-(x^*)^2/2\sigma^2}}{\sqrt{2\pi}} = \frac{1}{m}$. By the earlier Gaussian tail bound (G.1), it is easy to verify
$$\mathbf{Pr}[\max_i h_i < x^*] = (1 - \Pr_{h\sim\mathcal{N}(0,\sigma^2)}[h > x^*])^m \geq \Omega(1)$$

Using the earlier Gaussian tail bound (G.1), one can also verify that
$$\Pr_{g\sim\mathcal{N}(0,1)}[g > x^*] \geq \frac{1}{2x^*}\frac{e^{-(x^*)^2/2}}{\sqrt{2\pi}} \geq \frac{1}{2\sigma}\left(\frac{1}{m}\right)^{\sigma^2}$$

Therefore,
$$\mathbf{Pr}[\max_i g_i > x^*] = 1 - (1 - \Pr_{g\sim\mathcal{N}(0,1)}[g > x^*])^m \geq 1 - \left(1 - \Omega(m)\cdot\frac{1}{\sigma m^{\sigma^2}}\right)$$

Combining both, we have
$$\mathbf{Pr}[\max_i h_i < x^* < \max_i g_i] = \mathbf{Pr}[\max_i h_i < x^*]\cdot\mathbf{Pr}[\max_i g_i > x^*] \geq \frac{\Omega(1)}{\sigma m^{\sigma^2-1}}$$

**The case of $\sigma \leq 1$.** Let us first generate $h$ and then generate $g$. Since $\sigma \leq 1$, we have with probability at least $1 - \frac{1}{\mathsf{poly}(m)}$, it satisfies that $0 < \max_{i\in[m]} h_i \leq O(\sqrt{\log m})$. When this happens, denoting by $z = \max_{i\in[m]} h_i$, we can apply a known anti-concentration result for the maximum of Gaussian variables [21, Theorem 3]:
$$\mathbf{Pr}\left[\max_{i\in[m]} g_i \in [z(1-\tau), z(1+\tau)]\right] \leq O(z\tau)\cdot O(\mathbb{E}[\max_{i\in[m]} g_i]) \leq O(z\tau\sqrt{\log m})$$

This finishes the proof. □

Let us restate Proposition B.2 for the readers' convenience. Suppose we denote by $S_{i,\ell} \stackrel{\text{def}}{=} \mathbb{E}_{(X,y)\sim\mathcal{Z}_m}\left[\mathbb{1}_{y=i}\sum_{p\in P_{v_{i,\ell}}(X)} z_p^q\right]$. Then, define

$$\mathcal{M} \stackrel{\text{def}}{=} \left\{(i,\ell^*)\in[k]\times[2] \,\Big|\, \Lambda_{i,\ell^*}^{(0)} \geq \Lambda_{i,3-\ell^*}^{(0)}\left(\frac{S_{i,3-\ell^*}}{S_{i,\ell^*}}\right)^{\frac{1}{q-2}}\left(1 + \frac{1}{\log^2(m)}\right)\right\}$$

**Proposition B.2.** *Suppose $m \leq \mathsf{poly}(k)$. We have the following properties about $\mathcal{M}$.*

- *For every $i \in [k]$, at most one of $(i,1)$ or $(i,2)$ is in $\mathcal{M}$ (obvious).*
- *For every $i \in [k]$, suppose $S_{i,\ell} \geq S_{i,3-\ell}$, then*
  - $\mathbf{Pr}\left[(i,3-\ell)\in\mathcal{M}\right] \geq m^{-O(1)}$.
  - $\mathbf{Pr}\left[(i,\ell)\in\mathcal{M} \text{ or } (i,3-\ell)\in\mathcal{M}\right] \geq 1 - o(1)$

*Proof of Proposition B.2.* We only prove the second item since the first one is trivial. Suppose $S_{i,\ell} \geq S_{i,3-\ell}$.

By our assumption on the data distribution, it is easy to verify $S_{i,1}/S_{i,2} > 0$ is a constant for every $i \in [k]$. Therefore, $(i, 3-\ell) \in \mathcal{M}$ with probability at least $\frac{1}{m^{O(1)}}$ following the first item of Proposition G.2.

Finally, if neither $(i,\ell)$ or $(i, 3-\ell)$ is in $\mathcal{M}$, it necessarily implies

$$\Lambda_{i,\ell}^{(0)} = \Lambda_{i,3-\ell}^{(0)} \left(\frac{S_{i,3-\ell}}{S_{i,\ell}}\right)^{\frac{1}{q-2}} \left(1 \pm O\Big(\frac{1}{\log^2 m}\Big)\right)$$

but according to the second item of Proposition G.2, this happens with probability at most $\frac{1}{\log m}$.

□

# References

[1] Monther Alhamdoosh and Dianhui Wang. Fast decorrelated neural network ensembles with random weights. *Information Sciences*, 264:104–117, 2014.

[2] Zeyuan Allen-Zhu and Yuanzhi Li. What Can ResNet Learn Efficiently, Going Beyond Kernels? In *NeurIPS*, 2019. Full version available at http://arxiv.org/abs/1905.10337.

[3] Zeyuan Allen-Zhu and Yuanzhi Li. Can SGD Learn Recurrent Neural Networks with Provable Generalization? In *NeurIPS*, 2019. Full version available at http://arxiv.org/abs/1902.01028.

[4] Zeyuan Allen-Zhu and Yuanzhi Li. Backward feature correction: How deep learning performs deep learning. *arXiv preprint arXiv:2001.04413*, 2020.

[5] Zeyuan Allen-Zhu, Yuanzhi Li, and Zhao Song. On the convergence rate of training recurrent neural networks. In *NeurIPS*, 2019. Full version available at http://arxiv.org/abs/1810.12065.

[6] Zeyuan Allen-Zhu, Yuanzhi Li, and Zhao Song. A convergence theory for deep learning via over-parameterization. In *ICML*, 2019. Full version available at http://arxiv.org/abs/1811.03962.

[7] Jose M Alvarez, Yann LeCun, Theo Gevers, and Antonio M Lopez. Semantic road segmentation via multi-scale ensembles of learned features. In *European Conference on Computer Vision*, pages 586–595. Springer, 2012.

[8] Sanjeev Arora, Simon S Du, Wei Hu, Zhiyuan Li, Ruslan Salakhutdinov, and Ruosong Wang. On exact computation with an infinitely wide neural net. *arXiv preprint arXiv:1904.11955*, 2019.

[9] Sanjeev Arora, Simon S. Du, Wei Hu, Zhiyuan Li, and Ruosong Wang. Fine-grained analysis of optimization and generalization for overparameterized two-layer neural networks. *CoRR*, abs/1901.08584, 2019. URL http://arxiv.org/abs/1901.08584.

[10] Ainesh Bakshi, Rajesh Jayaram, and David P Woodruff. Learning two layer rectified neural networks in polynomial time. *arXiv preprint arXiv:1811.01885*, 2018.

[11] Verónica Bolón-Canedo and Amparo Alonso-Betanzos. Ensembles for feature selection: A review and future trends. *Information Fusion*, 52:1–12, 2019.

[12] Digvijay Boob and Guanghui Lan. Theoretical properties of the global optimizer of two layer neural network. *arXiv preprint arXiv:1710.11241*, 2017.

[13] Leo Breiman. Bagging predictors. *Machine learning*, 24(2):123–140, 1996.

[14] Gavin Brown, Jeremy Wyatt, Rachel Harris, and Xin Yao. Diversity creation methods: a survey and categorisation. *Information Fusion*, 6(1):5–20, 2005.

[15] Gavin Brown, Jeremy L Wyatt, and Peter Tiňo. Managing diversity in regression ensembles. *Journal of machine learning research*, 6(Sep):1621–1650, 2005.

[16] Alon Brutzkus and Amir Globerson. Globally optimal gradient descent for a convnet with gaussian inputs. *arXiv preprint arXiv:1702.07966*, 2017.

[17] Robert Bryll, Ricardo Gutierrez-Osuna, and Francis Quek. Attribute bagging: improving accuracy of classifier ensembles by using random feature subsets. *Pattern recognition*, 36(6):1291–1302, 2003.

[18] Jie Cai, Jiawei Luo, Shulin Wang, and Sheng Yang. Feature selection in machine learning: A new perspective. *Neurocomputing*, 300:70–79, 2018.

[19] Yuan Cao and Quanquan Gu. Generalization bounds of stochastic gradient descent for wide and deep neural networks. In *Advances in Neural Information Processing Systems*, pages 10835–10845, 2019.

[20] Yevgen Chebotar and Austin Waters. Distilling knowledge from ensembles of neural networks for speech recognition. In *Interspeech*, pages 3439–3443, 2016.

[21] Victor Chernozhukov, Denis Chetverikov, and Kengo Kato. Comparison and anti-concentration bounds for maxima of gaussian random vectors. *Probability Theory and Related Fields*, 162(1):47–70, 2015.

[22] Jia Cui, Brian Kingsbury, Bhuvana Ramabhadran, George Saon, Tom Sercu, Kartik Audhkhasi, Abhinav Sethy, Markus Nussbaum-Thom, and Andrew Rosenberg. Knowledge distillation across ensembles of multilingual models for low-resource languages. In *2017 IEEE International Conference on Acoustics, Speech and Signal Processing (ICASSP)*, pages 4825–4829. IEEE, 2017.

[23] Amit Daniely. Sgd learns the conjugate kernel class of the network. In *Advances in Neural Information Processing Systems*, pages 2422–2430, 2017.

[24] Amit Daniely, Roy Frostig, and Yoram Singer. Toward deeper understanding of neural networks: The power of initialization and a dual view on expressivity. In *Advances in Neural Information Processing Systems (NIPS)*, pages 2253–2261, 2016.

[25] Thomas G Dietterich. Ensemble methods in machine learning. In *International workshop on multiple classifier systems*, pages 1–15. Springer, 2000.

[26] Simon S Du, Jason D Lee, Haochuan Li, Liwei Wang, and Xiyu Zhai. Gradient descent finds global minima of deep neural networks. *arXiv preprint arXiv:1811.03804*, November 2018.

[27] Simon S Du, Xiyu Zhai, Barnabas Poczos, and Aarti Singh. Gradient descent provably optimizes over-parameterized neural networks. *arXiv preprint arXiv:1810.02054*, 2018.

[28] David A Freedman et al. Bootstrapping regression models. *The Annals of Statistics*, 9(6):1218–1228, 1981.

[29] Markus Freitag, Yaser Al-Onaizan, and Baskaran Sankaran. Ensemble distillation for neural machine translation. *arXiv preprint arXiv:1702.01802*, 2017.

[30] Yoav Freund and Robert E Schapire. A decision-theoretic generalization of on-line learning and an application to boosting. *Journal of computer and system sciences*, 55(1):119–139, 1997.

[31] Yoav Freund, Robert Schapire, and Naoki Abe. A short introduction to boosting. *Journal-Japanese Society For Artificial Intelligence*, 14(771-780):1612, 1999.

[32] Jerome Friedman, Trevor Hastie, Robert Tibshirani, et al. Additive logistic regression: a statistical view of boosting (with discussion and a rejoinder by the authors). *The annals of statistics*, 28(2):337–407, 2000.

[33] Jerome H Friedman. Greedy function approximation: a gradient boosting machine. *Annals of statistics*, pages 1189–1232, 2001.

[34] Takashi Fukuda, Masayuki Suzuki, Gakuto Kurata, Samuel Thomas, Jia Cui, and Bhuvana Ramabhadran. Efficient knowledge distillation from an ensemble of teachers. In *Interspeech*, pages 3697–3701, 2017.

[35] Tommaso Furlanello, Zachary C Lipton, Michael Tschannen, Laurent Itti, and Anima Anandkumar. Born again neural networks. *arXiv preprint arXiv:1805.04770*, 2018.

[36] Mikel Galar, Alberto Fernandez, Edurne Barrenechea, Humberto Bustince, and Francisco Herrera. A review on ensembles for the class imbalance problem: bagging-, boosting-, and hybrid-based approaches. *IEEE Transactions on Systems, Man, and Cybernetics, Part C (Applications and Reviews)*, 42(4):463–484, 2011.

[37] Rong Ge, Jason D Lee, and Tengyu Ma. Learning one-hidden-layer neural networks with landscape design. *arXiv preprint arXiv:1711.00501*, 2017.

[38] Rong Ge, Rohith Kuditipudi, Zhize Li, and Xiang Wang. Learning two-layer neural networks with symmetric inputs. *arXiv preprint arXiv:1810.06793*, 2018.

[39] Behrooz Ghorbani, Song Mei, Theodor Misiakiewicz, and Andrea Montanari. Linearized two-layers neural networks in high dimension. *arXiv preprint arXiv:1904.12191*, 2019.

[40] Boris Hanin and Mihai Nica. Finite depth and width corrections to the neural tangent kernel. *arXiv preprint arXiv:1909.05989*, 2019.

[41] Lars Kai Hansen and Peter Salamon. Neural network ensembles. *IEEE transactions on pattern analysis and machine intelligence*, 12(10):993–1001, 1990.

[42] Geoffrey Hinton, Oriol Vinyals, and Jeff Dean. Distilling the knowledge in a neural network. *arXiv preprint arXiv:1503.02531*, 2015.

[43] Tin Kam Ho. The random subspace method for constructing decision forests. *IEEE transactions on pattern analysis and machine intelligence*, 20(8):832–844, 1998.

[44] Arthur Jacot, Franck Gabriel, and Clément Hongler. Neural tangent kernel: Convergence and generalization in neural networks. In *Advances in neural information processing systems*, pages 8571–8580, 2018.

[45] Kenji Kawaguchi. Deep learning without poor local minima. In *Advances in Neural Information Processing Systems*, pages 586–594, 2016.

[46] Josef Kittler, Mohamad Hatef, Robert PW Duin, and Jiri Matas. On combining classifiers. *IEEE transactions on pattern analysis and machine intelligence*, 20(3):226–239, 1998.

[47] Ron Kohavi, George H John, et al. Wrappers for feature subset selection. *Artificial intelligence*, 97(1-2):273–324, 1997.

[48] J Zico Kolter and Marcus A Maloof. Dynamic weighted majority: An ensemble method for drifting concepts. *Journal of Machine Learning Research*, 8(Dec):2755–2790, 2007.

[49] Alex Krizhevsky. Learning multiple layers of features from tiny images. 2009.

[50] Anders Krogh and Jesper Vedelsby. Neural network ensembles, cross validation, and active learning. *Advances in neural information processing systems*, 7:231–238, 1994.

[51] Ludmila I Kuncheva. *Combining pattern classifiers: methods and algorithms*. John Wiley & Sons, 2014.

[52] Yuanzhi Li and Zehao Dou. When can wasserstein gans minimize wasserstein distance? *arXiv preprint arXiv:2003.04033*, 2020.

[53] Yuanzhi Li and Yingyu Liang. Provable alternating gradient descent for non-negative matrix factorization with strong correlations. In *Proceedings of the 34th International Conference on Machine Learning-Volume 70*, pages 2062–2070. JMLR. org, 2017.

[54] Yuanzhi Li and Yingyu Liang. Learning overparameterized neural networks via stochastic gradient descent on structured data. In *Advances in Neural Information Processing Systems*, 2018.

[55] Yuanzhi Li and Yang Yuan. Convergence analysis of two-layer neural networks with relu activation. In *Advances in Neural Information Processing Systems*, pages 597–607. http://arxiv.org/abs/1705.09886, 2017.

[56] Yuanzhi Li, Yingyu Liang, and Andrej Risteski. Recovery guarantee of non-negative matrix factorization via alternating updates. In *Advances in neural information processing systems*, pages 4987–4995, 2016.

[57] Yuanzhi Li, Tengyu Ma, and Hongyang Zhang. Algorithmic regularization in over-parameterized matrix sensing and neural networks with quadratic activations. In *COLT*, 2018.

[58] Yuanzhi Li, Colin Wei, and Tengyu Ma. Towards explaining the regularization effect of initial large learning rate in training neural networks. *arXiv preprint arXiv:1907.04595*, 2019.

[59] Yuanzhi Li, Tengyu Ma, and Hongyang R Zhang. Learning over-parametrized two-layer relu neural networks beyond ntk. *arXiv preprint arXiv:2007.04596*, 2020.

[60] Xiaodong Liu, Pengcheng He, Weizhu Chen, and Jianfeng Gao. Improving multi-task deep neural networks via knowledge distillation for natural language understanding. *arXiv preprint arXiv:1904.09482*, 2019.

[61] Pankaj Mehta, Marin Bukov, Ching-Hao Wang, Alexandre GR Day, Clint Richardson, Charles K Fisher, and David J Schwab. A high-bias, low-variance introduction to machine learning for physicists. *Physics reports*, 810:1–124, 2019.

[62] Hossein Mobahi, Mehrdad Farajtabar, and Peter L Bartlett. Self-distillation amplifies regularization in hilbert space. *arXiv preprint arXiv:2002.05715*, 2020.

[63] M Arthur Munson and Rich Caruana. On feature selection, bias-variance, and bagging. In *Joint European Conference on Machine Learning and Knowledge Discovery in Databases*, pages 144–159. Springer, 2009.

[64] Chris Olah, Alexander Mordvintsev, and Ludwig Schubert. Feature visualization. *Distill*, 2(11):e7, 2017. doi: 10.23915/distill.00007. https://distill.pub/2017/feature-visualization.

[65] Luiz S Oliveira, Robert Sabourin, Flávio Bortolozzi, and Ching Y Suen. Feature selection for ensembles: A hierarchical multi-objective genetic algorithm approach. In *Seventh International Conference on Document Analysis and Recognition, 2003. Proceedings.*, pages 676–680. Citeseer, 2003.

[66] David Opitz and Richard Maclin. Popular ensemble methods: An empirical study. *Journal of artificial intelligence research*, 11:169–198, 1999.

[67] David W Opitz. Feature selection for ensembles. In *AAAI*, pages 379–384, 1999.

[68] Samet Oymak and Mahdi Soltanolkotabi. Towards moderate overparameterization: global convergence guarantees for training shallow neural networks. *arXiv preprint arXiv:1902.04674*, 2019.

[69] Michael P Perrone and Leon N Cooper. When networks disagree: Ensemble methods for hybrid neural networks. Technical report, BROWN UNIV PROVIDENCE RI INST FOR BRAIN AND NEURAL SYSTEMS, 1992.

[70] Robi Polikar. Ensemble based systems in decision making. *IEEE Circuits and systems magazine*, 6(3): 21–45, 2006.

[71] Juan José Rodriguez, Ludmila I Kuncheva, and Carlos J Alonso. Rotation forest: A new classifier ensemble method. *IEEE transactions on pattern analysis and machine intelligence*, 28(10):1619–1630, 2006.

[72] Lior Rokach. Ensemble-based classifiers. *Artificial intelligence review*, 33(1-2):1–39, 2010.

[73] Lior Rokach. *Pattern classification using ensemble methods*, volume 75. World Scientific, 2010.

[74] Lior Rokach and Oded Z Maimon. *Data mining with decision trees: theory and applications*, volume 69. World scientific, 2008.

[75] Robert E Schapire, Yoav Freund, Peter Bartlett, Wee Sun Lee, et al. Boosting the margin: A new explanation for the effectiveness of voting methods. *The annals of statistics*, 26(5):1651–1686, 1998.

[76] Vaishaal Shankar, Alex Fang, Wenshuo Guo, Sara Fridovich-Keil, Ludwig Schmidt, Jonathan Ragan-Kelley, and Benjamin Recht. Neural kernels without tangents. *arXiv preprint arXiv:2003.02237*, 2020.

[77] Mahdi Soltanolkotabi, Adel Javanmard, and Jason D Lee. Theoretical insights into the optimization landscape of over-parameterized shallow neural networks. *arXiv preprint arXiv:1707.04926*, 2017.

[78] Daniel Soudry and Yair Carmon. No bad local minima: Data independent training error guarantees for multilayer neural networks. *arXiv preprint arXiv:1605.08361*, 2016.

[79] Yuandong Tian. An analytical formula of population gradient for two-layered relu network and its applications in convergence and critical point analysis. *arXiv preprint arXiv:1703.00560*, 2017.

[80] Alexey Tsymbal, Mykola Pechenizkiy, and Pádraig Cunningham. Diversity in search strategies for ensemble feature selection. *Information fusion*, 6(1):83–98, 2005.

[81] Giorgio Valentini. An experimental bias-variance analysis of svm ensembles based on resampling techniques. *IEEE Transactions on Systems, Man, and Cybernetics, Part B (Cybernetics)*, 35(6):1252–1271, 2005.

[82] Giorgio Valentini and Thomas G Dietterich. Bias-variance analysis of support vector machines for the development of svm-based ensemble methods. *Journal of Machine Learning Research*, 5(Jul):725–775, 2004.
68

[83] Santosh Vempala and John Wilmes. Polynomial convergence of gradient descent for training one-hidden-layer neural networks. *arXiv preprint arXiv:1805.02677*, 2018.

[84] Bo Xie, Yingyu Liang, and Le Song. Diversity leads to generalization in neural networks. *arXiv preprint Arxiv:1611.03131*, 2016.

[85] Greg Yang. Scaling limits of wide neural networks with weight sharing: Gaussian process behavior, gradient independence, and neural tangent kernel derivation. *arXiv preprint arXiv:1902.04760*, 2019.

[86] Gilad Yehudai and Ohad Shamir. On the power and limitations of random features for understanding neural networks. *arXiv preprint arXiv:1904.00687*, 2019.

[87] Sergey Zagoruyko and Nikos Komodakis. Wide residual networks. *arXiv preprint arXiv:1605.07146*, 2016.

[88] Linfeng Zhang, Jiebo Song, Anni Gao, Jingwei Chen, Chenglong Bao, and Kaisheng Ma. Be your own teacher: Improve the performance of convolutional neural networks via self distillation. In *ICCV*, pages 3713–3722, 2019.

[89] Kai Zhong, Zhao Song, Prateek Jain, Peter L Bartlett, and Inderjit S Dhillon. Recovery guarantees for one-hidden-layer neural networks. *arXiv preprint arXiv:1706.03175*, 2017.

[90] Zhi-Hua Zhou, Jianxin Wu, and Wei Tang. Ensembling neural networks: many could be better than all. *Artificial intelligence*, 137(1-2):239–263, 2002.

[91] Xiatian Zhu, Shaogang Gong, et al. Knowledge distillation by on-the-fly native ensemble. In *Advances in neural information processing systems*, pages 7517–7527, 2018.

[92] Difan Zou, Yuan Cao, Dongruo Zhou, and Quanquan Gu. Stochastic gradient descent optimizes over-parameterized deep relu networks. *arXiv preprint arXiv:1811.08888*, 2018.